%% file: arxiv_main.tex
\documentclass[journal]{IEEEtran}

\input{sections/preamble_arxiv}

\begin{document}

\title{Reproducible Dynamic Parameter Identification\\ for a Low-Cost Robot Arm:\\ A Positive-Definiteness Audit for Model Acceptance}

\author{Junji~Oaki,~\IEEEmembership{Member,~IEEE,}
        Koki~Yamane,~\IEEEmembership{Student~Member,~IEEE,}
        Koki~Inami,~\IEEEmembership{Student~Member,~IEEE,}
        and~Sho~Sakaino,~\IEEEmembership{Member,~IEEE}%
\thanks{This work was supported by JSPS KAKENHI Grant Numbers 24K00905 and 24KJ0503, JST PRESTO Grant Number JPMJPR24T3, JST ALCA-Next Japan Grant Number JPMJAN24F1, and JST-Research \& Development Program for Next Generation Edge AI Semiconductors Japan Grant Number JPMJES2514. This study was based on the results obtained from the JPNP20004 project subsidized by the New Energy and Industrial Technology Development Organization (NEDO).}%
\thanks{J. Oaki and S. Sakaino are with the Faculty of Engineering, Information and Systems, University of Tsukuba, 1-1-1 Tennodai, Tsukuba, Ibaraki 305-8573, Japan (e-mail: oaki.junji.fu@u.tsukuba.ac.jp).}%
\thanks{K. Yamane and K. Inami are with the Degree Programs in Intelligent and Mechanical Interaction Systems, University of Tsukuba, 1-1-1 Tennodai, Tsukuba, Ibaraki 305-8573, Japan.}%
\thanks{This work has been submitted to the IEEE for possible publication. Copyright may be transferred without notice, after which this version may no longer be accessible.}%
\thanks{This is version 3 of arXiv:2605.15949. Versions 1 and 2 reported an earlier stage of the same study under the title ``A Reproducible and Physically Feasible Dynamic Parameter Identification Framework for a Low-Cost Robot Arm'' and are superseded by this one. Four defects of the earlier versions were found and corrected, three of which changed reported numbers: the closed-loop simulator did not reproduce the velocity feedforward that the robot applies, so all 160 identification conditions were recomputed; the pose set behind the feasibility audit was too sparse, so that the accepted model's own verdict depended on the random seed; the three feasibility stages used different pose sets without saying so; and the trajectory specification described the excitation inaccurately. The representative model is now named by a structural property of the excitation design rather than by statistical centrality, every comparison is reported against the measured run-to-run variation of the platform, and the material that was supplementary in the journal submission is included here as Appendices~\ref{app:traj} to~\ref{app:validation}.}}

\maketitle

\input{sections/abstract}


\IEEEpeerreviewmaketitle

\input{sections/01_introduction}
\input{sections/02_related_work}
\input{sections/03_robot_model}

\input{sections/04_identification_procedure}

\input{sections/05_experimental_conditions}
\input{sections/06_results}
\input{sections/07_discussion}
\input{sections/08_conclusion}

\appendices
\input{sections/appendix_a_model}

\input{sections/appendices}
\input{sections/appendix_d_implementation}
\input{sections/appendix_e_validation}

\input{sections/acknowledgment}

\bibliographystyle{IEEEtran}
\bibliography{refs}

\input{sections/author}

\end{document}

%% file: sections/preamble_arxiv.tex
\usepackage{cite}
\usepackage{amsmath}
\usepackage{newtxtext}
\usepackage{newtxmath}
\usepackage{algorithm}
\usepackage{algorithmic}
\usepackage{graphicx}
\usepackage{textcomp}
\usepackage{bm}
\usepackage{booktabs,tabularx,array,makecell,multirow}
\usepackage[caption=false,font=footnotesize]{subfig}
\usepackage{url}
\usepackage{verbatim}
\usepackage{needspace}   

\newcolumntype{Y}{>{\raggedright\arraybackslash}X}

\newcommand{\PARstart}{\IEEEPARstart}

\newcommand{\btheta}{\bm{\theta}}
\newcommand{\btau}{\bm{\tau}}
\newcommand{\bphi}{\bm{\phi}}

\newcommand{\bM}{\bm{M}}
\newcommand{\bC}{\bm{C}}
\newcommand{\bg}{\bm{g}}
\newcommand{\bW}{\bm{W}}

\newcommand{\bI}{\bm{I}}
\newcommand{\bA}{\bm{A}}

\newcommand{\bD}{\bm{D}}

\newcommand{\bq}{\bm{q}}
\newcommand{\be}{\bm{e}}
\newcommand{\bh}{\bm{h}}
\newcommand{\bw}{\bm{w}}

\newcommand{\epsPD}{\varepsilon_{\mathrm{PD}}}

\DeclareMathOperator*{\argmin}{arg\,min}

\DeclareMathOperator{\diag}{diag}


%% file: sections/abstract.tex
\begin{abstract}
Dynamic parameter identification of low-cost robot arms is challenging because limited sensing and drivetrain nonidealities can yield models that predict measured torques well but are physically unsuitable for model-based control. This paper presents a reproducible dynamic parameter identification pipeline for CRANE-X7, a low-cost seven-degree-of-freedom arm driven by modular smart actuators. A 39-parameter OpenSYMORO base-parameter model is identified without CAD inertial data using fully specified single-joint and adjacent-pair excitation, ordinary least squares, a conditional semidefinite-programming projection, and closed-loop input error refinement. Model acceptance is decided by a separate positive-definiteness audit of the identified inertia matrix over 221{,}875 sampled configurations. Experiments cover 40 identification trajectories at four sampling intervals and three held-out validation trajectories. The reduced model improves held-out prediction over the full 65-parameter model on all seven identification trajectories used for the model comparison. Fixed-configuration analyses show the distinct role of the feasibility audit: models with similar torque predictions can produce unstable acceleration-resolved dynamics and reverse the direction of the inertia inversion. For the selected identification trajectory, two executions separated by 26 days yield a 1.25\% relative spread in held-out root-mean-square error. The accepted model passes audits with five random seeds, and sensitivity analyses quantify the effects of the sampling interval and of torque-constant uncertainty. The complete trajectory specification and numerical record support independent implementation and comparison. These results establish predictive performance, inertia-matrix feasibility, and repeatability as complementary criteria for evaluating dynamic models of low-cost arms.
\end{abstract}

%% file: sections/01_introduction.tex
\section{Introduction}

\PARstart{L}{ow-cost} robot manipulators are becoming increasingly important as research platforms for teleoperation, data collection, and imitation learning. 
Open low-cost manipulation platforms such as ALOHA and ALOHA~2 have accelerated large-scale teleoperation and imitation-learning research, partly by combining accessible hardware with reproducible data-collection procedures \cite{ZhaoEtAl2023ALOHA,AldacoEtAl2024ALOHA2}. 
Many such platforms rely on compact modular smart actuators such as ROBOTIS DYNAMIXEL X-series devices, which support current-based operation and daisy-chain communication but also introduce limited sensing resolution, communication delay, and drivetrain nonidealities.
The CRANE-X7 platform used in this study is a representative low-cost arm based on DYNAMIXEL X-series actuators~\cite{CRANEX7Website, robotis_dynamixel_x}, and recent studies have demonstrated its practical use in bilateral-control-based manipulation, 
motion modification, and sensorless teleoperation~\cite{Yamane2024CrossStructureHand,Inami2025MotionReTouch,Yamane2025FastBilateral}.
More broadly, robot arms are also used as positioning devices in automated visual inspection systems, where arm motion can account for a substantial fraction of the total cycle time~\cite{MaoEtAl2022RimInspection}.

Dynamic parameter identification on low-cost robot arms remains challenging. Compared with industrial manipulators, the sensing resolution is limited, communication and realization delays are not negligible, and drivetrain nonidealities such as backlash-related effects are more pronounced. As a result, straightforward identification with a high-dimensional rigid-body model is often numerically fragile and may produce parameter sets that fit measured torque data well but remain physically implausible over a broader pose range.

Classical studies established that robot dynamics can be expressed linearly with respect to inertial parameters and that only a reduced set of base parameters is structurally identifiable \cite{Mayeda1984NewIdentification,MayedaYoshidaOsukaBaseParam,KhalilGautierIDIM}. Subsequent work addressed physical feasibility and positive-definiteness recovery through constrained formulations such as linear matrix inequalities and semidefinite programming \cite{YoshidaKhalilPhysicalConsistency,SousaCortesao2014LMI,WensingPhysicalConsistency}. Closed-loop and output-error/input-error methods further improved predictive performance in robot identification \cite{CLOEReference,CLIEReference,BIRDy2021}, while symbolic tools such as OpenSYMORO improved the reproducibility of model and regressor generation \cite{OpenSYMORO}. However, a compact framework that combines these ideas for a low-cost robot arm, while explicitly screening the final model for broad-range physical acceptability, is still lacking.

This paper presents a reproducible dynamic parameter identification pipeline for CRANE-X7 and tests its result for physical feasibility before accepting it. The excitation is hand-designed from single-joint and adjacent-two-joint primitives under practical joint-range limits, in the line of the classical test motions of Mayeda \textit{et al.} \cite{Mayeda1984NewIdentification,KhalilDombre2002}, and is specified completely. The model is a 39-parameter OpenSYMORO base-parameter model without CAD data, the products of inertia and the Coulomb friction terms removed. Identification runs through inverse dynamics identification model (IDIM) regression, ordinary least squares (OLS), a conditional semidefinite programming (SDP) projection, and closed-loop input error (CLIE) refinement; the candidate is named by the excitation design, and the accepted model is the one that passes an all-pose positive-definiteness audit of the inertia matrix on configurations the identification never visited.

The contributions are threefold.
\begin{itemize}
\item A reproducible dynamic identification pipeline for a low-cost arm, combining fully specified excitation, a no-CAD reduced model, and a complete numerical record of 160 identification conditions.
\item An evaluation of inertia-matrix feasibility as a model-acceptance criterion separate from prediction accuracy, including its consequences for acceleration-resolved control and reaction-force observation.
\item A quantitative assessment of practical reproducibility and sensitivity through repeated trajectory executions, pose-sampling audits, and perturbations of the sampling interval and of the torque constants.
\end{itemize}
For low-cost arms, an explicit inertia-matrix feasibility test complements prediction-based model evaluation.
Appendix~\ref{app:model} specifies the model and its base-parameter grouping and gives the complete numerical record of the 160 identification conditions; Appendices~\ref{app:traj} to~\ref{app:validation} carry the trajectory specification, the routing tables, the implementation settings, and the validation details. The numerical specification of every trajectory and the solver check accompany this preprint as ancillary files.

%% file: sections/02_related_work.tex
\section{Related Work}

\subsection{Base Parameters and Excitation}

Classical robot dynamic identification established inverse-dynamics regressors, least-squares estimation, and base-parameter representations \cite{Mayeda1984NewIdentification,MayedaYoshidaOsukaBaseParam,KhalilGautierIDIM,AtkesonAnHollerbach1986IJRR}. Two consequences of that literature are used throughout this paper: only linear combinations of the standard inertial parameters are identifiable, and the specific combinations follow from the kinematic structure, so which quantities survive ungrouped is a property of the arm rather than of the estimator. Symbolic tools such as OpenSYMORO~\cite{OpenSYMORO} make the grouping explicit and reproducible for a given robot.

Excitation design has been studied using condition numbers, optimized Fourier or polynomial trajectories, and information-theoretic criteria that select the next motion by its expected information gain \cite{SweversEtAl1996MSSP,SweversEtAl1997TRA,PresseGautier1993ICRA,Park2006Robotica,WuWangYou2010RCIM,SujanDubowsky2003TMECH}. These designs are powerful when accurately executable, but they assume the commanded waveform is realized, which is where a low-cost platform departs: delay, finite encoder resolution, and drivetrain imperfections all act between the command and the motion. The present work therefore uses interpretable single-joint and adjacent-joint primitives, following the classical observation that motions of at most two joints suffice when the products of inertia are not carried and consecutive axes are perpendicular or parallel~\cite{Mayeda1984NewIdentification,KhalilDombre2002}.

\subsection{Physical Feasibility: Conditions, Enforcement, and Testing}

Three separable questions arise once feasibility is taken seriously, and the literature answers them in different places.

\emph{Which condition.} Positive definiteness of the inertia matrix can be verified directly from base parameters~\cite{YoshidaKhalilPhysicalConsistency}. Full physical consistency is stronger and is characterized by positive definiteness of the \(4\times4\) pseudo-inertia of each link~\cite{WensingPhysicalConsistency}, which additionally forces the triangle inequality on the principal moments and is what link-level simulators require.

\emph{How to enforce it.} Linear matrix inequalities and semidefinite programming impose the condition during estimation \cite{SousaCortesao2014LMI,WensingPhysicalConsistency,MataEtAl2005AdvancedRobotics,JinGans2015RCIM,SousaCortesao2019TMECH}. Geometric~\cite{LeeWensingPark2020Geometric}, log-Cholesky~\cite{RuckerWensing2022SmoothInertia}, and Lie-theoretic~\cite{FuEtAl2021TMECH} parameterizations make consistency a property of the parameterization rather than of the solver; all operate on standard parameters, which a reduced model does not have.

\emph{How to test it.} Verification has received less attention than enforcement, although a model that was constrained under one condition still has to be checked under the one that matters for its intended use. Rapid feasibility tests for base parameters have been proposed, including a learned surrogate for the LMI test~\cite{HuangChengHuang2023FeasibilityTest}. The all-pose audit used here belongs to this third category: it asks whether the assembled \(\bM(\btheta)\) is positive definite everywhere in a stated pose domain, which is the property the uses in Sec.~VII-C depend on, and Sec.~VI shows that the answer depends strongly on how densely the domain is sampled.

\subsection{Closed-Loop Refinement and Data-Driven Alternatives}

Closed-loop and output-error or input-error methods, including CLOE, CLIE, and DIDIM, account for realized closed-loop behavior rather than for the inverse-dynamics residual alone \cite{CLOEReference,CLIEReference,BIRDy2021,SweversVerdonckDeSchutter2007CSM}. Physically consistent DIDIM avoids simulating from infeasible intermediate estimates and shows that seeding the non-identifiable subspace from CAD leaves the base parameters unbiased, since the regroupings absorb the error~\cite{JanotWensing2021CEP}; a constrained instrumental-variable method has since argued the opposite, imposing sign constraints on base parameters alone and dispensing with the prior~\cite{ArdianiJanotBenoussaad2025CEP}. This study starts from data-driven OLS estimates in the OpenSYMORO base-parameter space without CAD inertial parameters, and uses CLIE for the refinement stage.

A different response to unmodelled dynamics is to learn them. Hybrid schemes fit a neural network to the friction torque left over after a Stribeck model~\cite{HuangKeZhangOta2023TRO}, and fully data-driven estimators such as Gaussian process regression with energy-based kernels can predict inverse dynamics without an explicit parameter set~\cite{GiacomuzzoEtAl2024GPLIP}. These are effective where the residual is a function of the measured state; Sec.~VII-A reports why that route was not taken here.

\subsection{Platforms}

Studies on the Franka Emika Panda, KUKA LBR iiwa, Universal Robot UR5, and the da Vinci Research Kit demonstrate the need for feasible identified models in predictive control and simulation \cite{GazEtAl2019Panda,XuEtAl2020KUKAiiwa,HuangKeZhangOta2023TRO,WangEtAl2019DVRK}. These arms provide joint torque sensing or high-resolution encoders, and the identification problem is correspondingly better conditioned; the same holds for parallel mechanisms, where an identified reduced parameter set has been carried through to model-based control~\cite{DiazRodriguezEtAl2013TMECH}. In parallel, CRANE-X7, ALOHA, and ALOHA~2 have made low-cost hardware the standard vehicle for imitation learning and teleoperation data collection \cite{CRANEX7Website,Yamane2024CrossStructureHand,Inami2025MotionReTouch,Yamane2025FastBilateral,ZhaoEtAl2023ALOHA,AldacoEtAl2024ALOHA2}, without a matching body of work on obtaining dynamic models for them. The present work combines the strands above into a no-CAD pipeline for that setting.

%% file: sections/03_robot_model.tex
\section{Robot System and Dynamic Modeling}

\subsection{CRANE-X7 Platform}

\begin{figure}[t]
\centering
\includegraphics[width=\linewidth]{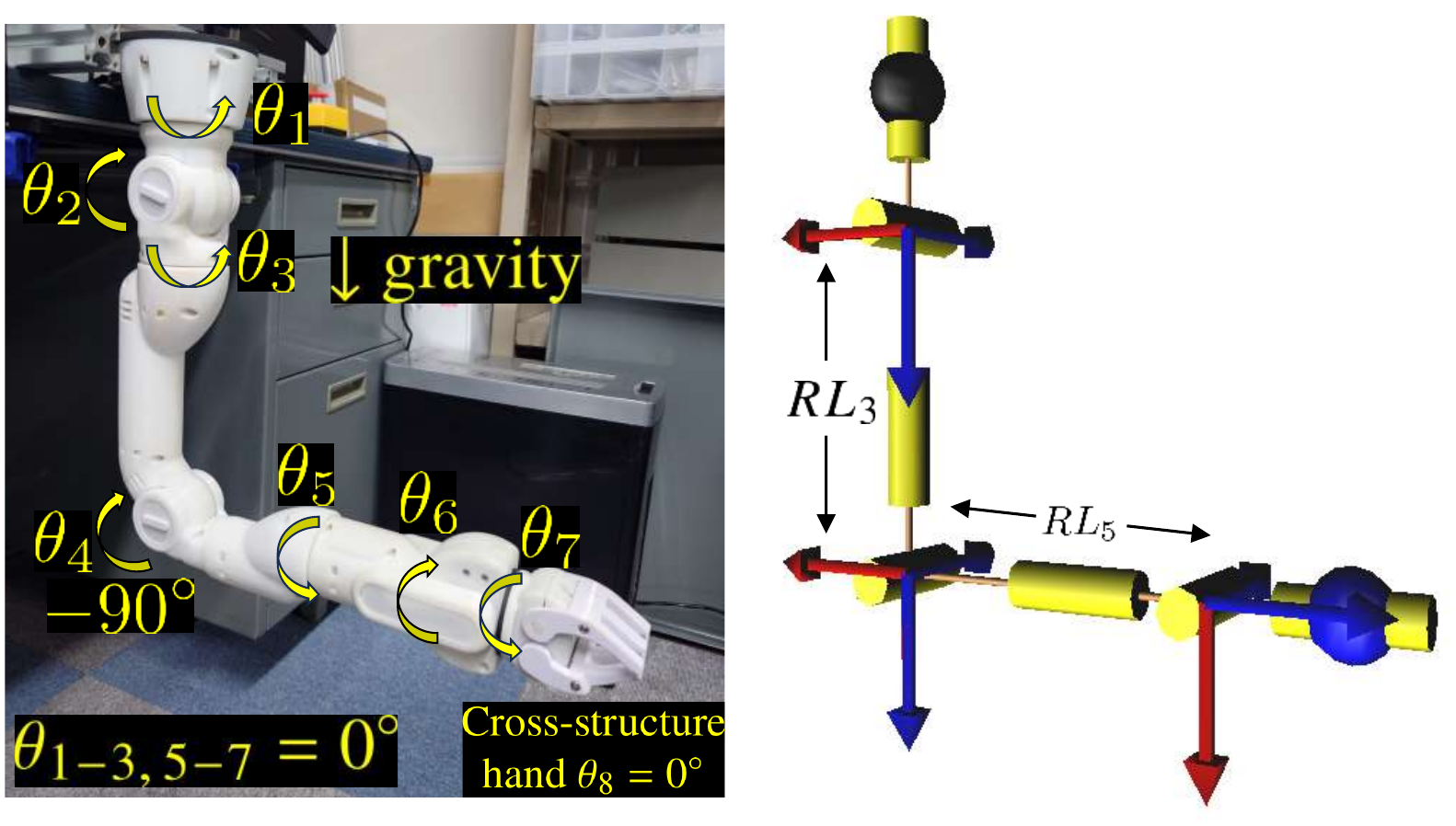}
\caption{CRANE-X7 with the replaced cross-structure hand \cite{Yamane2024CrossStructureHand} and its frame assignment under the modified Denavit-Hartenberg (DH) parameterization. The corresponding DH parameters are listed in Table~\ref{tab:dh}.}
\label{fig:platform}
\end{figure}

The experimental platform used in this study was CRANE-X7, a low-cost 7-degree-of-freedom serial robot arm equipped with a one-degree-of-freedom hand axis.
Figure~\ref{fig:platform} shows the CRANE-X7 platform and the modified DH parameter setting adopted for symbolic model generation.
The posture shown in Fig.~\ref{fig:platform} is defined by
\(\theta_1=\theta_2=\theta_3=\theta_5=\theta_6=\theta_7=0^\circ\) and
\(\theta_4=-90^\circ\), and was used as the nominal posture for the joint-level realization model introduced below.

Despite its flexibility as a research platform, CRANE-X7 presents several challenges for dynamic parameter identification: 12-bit link-side encoders, communication delay and jitter, and realized joint behavior that deviates from the ideal torque source assumed in industrial robot identification. All eight axes were driven by ROBOTIS DYNAMIXEL X-series actuators connected through an RS485 daisy chain~\cite{robotis_dynamixel_x}. Axes are numbered 1 to 8 outward from the base, the eighth being the hand; the figures label them J1 to J8. Axes~1 and 3--8 used XM430-W350 actuators, whereas Axis~2 used an XM540-W270 actuator. The current-to-torque conversion constants used in this study were \(2.3179~\mathrm{N\cdot m/A}\) for the XM430-W350 actuators and \(3.1317~\mathrm{N\cdot m/A}\) for the XM540-W270 actuator. The torque records used for identification were obtained from the actuator current values using these conversion constants. Each constant is the catalogue torque constant of the device, \(1.783\) and \(2.409~\mathrm{N\cdot m/A}\) respectively, multiplied by an empirical correction factor of \(1.3\) that is supplied with the platform's control library; the current itself is the commanded value in units of \(2.69~\mathrm{mA}\). No part of that chain is calibrated against a torque measurement in the experiments reported here, which makes it the least well characterized input to the identification. Sec.~VII-D quantifies what an error in it would do.
The robot was operated in current-command mode from a Linux PC, which enabled torque-level control and model-based motion generation. As a result, high-gain feedback alone is insufficient for repeatable identification motions, motivating a model-assisted motion-realization layer.

\input{tables/tab_dh_parameters}

Table~\ref{tab:dh} gives the modified DH parameters corresponding to the frame assignment in Fig.~\ref{fig:platform}.
They are listed here rather than embedded in the figure so that the kinematic model can be reproduced directly: together with the reduced parameter set of Sec.~III-C, they are sufficient to regenerate the symbolic regressor and the direct dynamic model with OpenSYMORO.

Two offsets of the manufacturer's published URDF~\cite{RTCraneX7Description}, a \(0.105~\mathrm{m}\) rise along the Joint-1 axis and a \(0.019~\mathrm{m}\) offset along \(z_7\), are absent from Table~\ref{tab:dh} and do not enter the dynamic model as kinematic quantities; Appendix~A states where each goes.

\subsection{Joint-Level Modeling and Identification}

\begin{figure}[t]
    \centering
    \includegraphics[width=\linewidth]{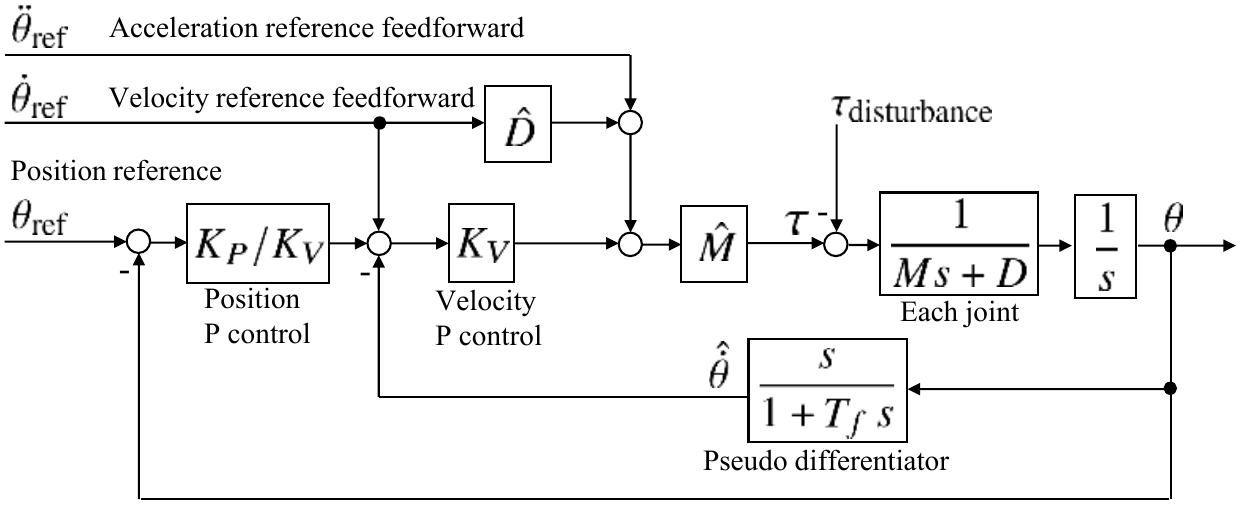}
\caption{Fixed FF+FB controller with pseudo differentiation that executes the trajectories; \(\hat{\bM}\) and \(\hat{\bD}\) are the diagonal realization constants of Table~\ref{tab:realization}, identified as in Fig.~\ref{fig:joint_ident}.}
\label{fig:joint_modeling}
\end{figure}

A simplified joint-level model of each axis serves only to execute the excitation trajectories stably and reproducibly; Fig.~\ref{fig:joint_modeling} shows the FF+FB controller built on it.

The joint dynamics are modeled as
\begin{equation}
G_j(s) = \frac{K_{p,j}}{1 + T_{p1,j} s} \exp(-T_{d,j} s),
\label{eq:joint_fopdt}
\end{equation}
where $K_{p,j}$, $T_{p1,j}$, and $T_{d,j}$ denote the low-frequency gain, time constant, and effective dead time of Joint $j$, respectively.

\begin{figure}[t]
\centering
\includegraphics[width=\columnwidth]{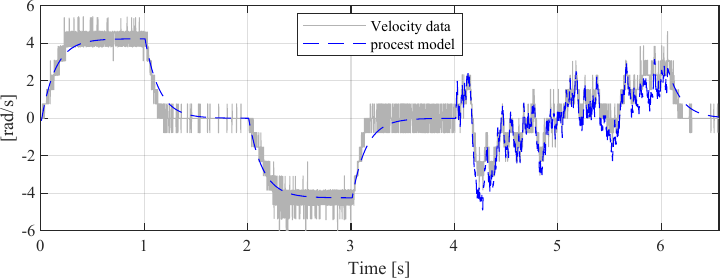}\\[1mm]
{\footnotesize(a)}\\[2mm]
\includegraphics[width=\columnwidth]{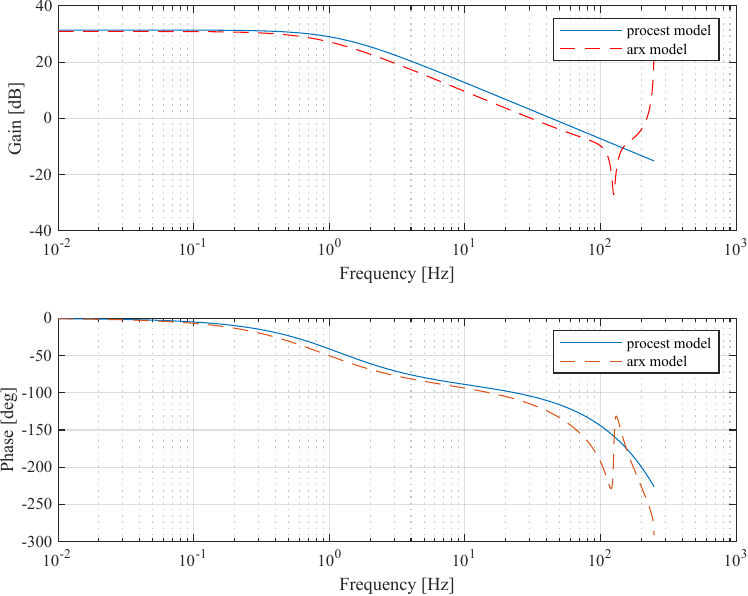}\\[1mm]
{\footnotesize(b)}
\caption{Joint-level identification of Joint~7 for the model \eqref{eq:joint_fopdt}: (a) measured velocity and the response of the fitted first-order-plus-dead-time model on the square-wave and PRBS record; (b) frequency response of that model against a fourth-order ARX model from the same record.}
\label{fig:joint_ident}
\end{figure}

The parameters were identified by fitting \eqref{eq:joint_fopdt} in MATLAB with \texttt{procest} to an open-loop record of the joint containing a square wave followed by a pseudo-random binary sequence (PRBS), the output velocity being obtained by encoder differencing; the square wave fixes the behavior down to DC, and the PRBS portion, about \(2~\mathrm{s}\) long, is informative from roughly \(0.5~\mathrm{Hz}\) upward. Fig.~\ref{fig:joint_ident} shows the fit for Joint~7, (a) in the time domain and (b) against the frequency response of a fourth-order autoregressive model with exogenous input (ARX) identified from the same record and used only for this comparison, which resolves what the step response alone does not: the dead time and the backlash-induced resonance near \(100~\mathrm{Hz}\), which is not modeled at the robot level. With the input torque converted from current by the catalogue constant \(1.783~\mathrm{N\cdot m/A}\) of Sec.~III-A and the output velocity in \(\mathrm{rad/s}\), the Joint-7 parameters are \(K_{p,7} = 37.1~\mathrm{(rad/s)/(N\cdot m)}\), the \(31.4~\mathrm{dB}\) of Fig.~\ref{fig:joint_ident}(b), \(T_{p1,7} = 0.136~\mathrm{s}\), and \(T_{d,7} = 1.53~\mathrm{ms}\), the dead time being consistent with the \(0.9~\mathrm{ms}\) of encoder reading and command dispatch plus the \(0.5~\mathrm{ms}\) sample-and-hold of the \(1~\mathrm{ms}\) cycle; similar models were obtained for the other joints. These are working values for the realization layer rather than actuator constants: the low-frequency gain in particular varies from run to run and is left to the feedback loop of \eqref{eq:fffb_command} to absorb.

Using this realization-oriented model, the structured excitation trajectories are executed by a fixed feedforward-plus-feedback controller with pseudo differentiation.
The feedback part is proportional-derivative; we write FF+FB rather than FF+PD because PD is reserved in this paper for positive definiteness of the inertia matrix.
The pseudo-differentiated velocity is \(\hat{\dot{\btheta}}=s\,\btheta/(1+T_f s)\) with \(T_f=0.002~\mathrm{s}\), and the realized joint command is
\begin{equation}
\begin{split}
\bm{u}_{\mathrm{FF+FB}} = \hat{\bM}\bigg[\,
& K_V \Big(
\dot{\btheta}_{\mathrm{ref}}
+ K_P K_V^{-1} (\btheta_{\mathrm{ref}} - \btheta)
- \hat{\dot{\btheta}}
\Big) \\
& + \ddot{\btheta}_{\mathrm{ref}}
+ \hat{\bD}\, \dot{\btheta}_{\mathrm{ref}}
\bigg].
\end{split}
\label{eq:fffb_command}
\end{equation}
Here \(K_P\) and \(K_V\) are scalar position and velocity gains common to every axis, and \(\hat{\bM}=\diag(\hat M_j)\) and \(\hat{\bD}=\diag(\hat D_j)\) are fixed diagonal realization matrices whose entries are the per-axis constants of Table~\ref{tab:realization}, obtained from the joint-level modeling.
They serve only for motion realization and are not the robot-level dynamic parameters identified in Sec.~IV; in particular \(\hat{\bM}\) is a constant diagonal matrix and is not an estimate of the configuration-dependent inertia matrix \(\bM(\btheta)\) of \eqref{eq:rigid_dynamics}.

The velocity-feedforward term sits \emph{inside} the \(\hat{\bM}\) bracket, so its torque-level contribution on axis \(j\) is \(\hat M_j\hat D_j\dot\theta_{j,\mathrm{ref}}\), one to two orders of magnitude below \(\hat D_j\dot\theta_{j,\mathrm{ref}}\); the CLIE simulation of Sec.~IV-C must use this form to reproduce the measured closed loop.

The position and velocity gains are common to every axis, and \(\hat M_j\) enters the velocity gain as well, so the closed-loop bandwidth is nominally uniform across the arm at \(\sqrt{K_P}=31.6~\mathrm{rad/s}=5.03~\mathrm{Hz}\).

\subsection{Reduced Rigid-Body Dynamic Model}

The robot-level parameter identification is performed using the following reduced rigid-body model:
\begin{equation}
\btau = \bM(\btheta)\ddot{\btheta} + \bC(\btheta,\dot{\btheta})\dot{\btheta} + \bg(\btheta) + \bm{f}_v(\dot{\btheta}),
\label{eq:rigid_dynamics}
\end{equation}
where \(\btau\) is the joint torque vector, \(\bM(\btheta)\) is the inertia matrix, \(\bC(\btheta,\dot{\btheta})\dot{\btheta}\) represents Coriolis and centrifugal effects, \(\bg(\btheta)\) is the gravity term, and \(\bm{f}_v(\dot{\btheta})\) denotes viscous friction.
Here \(\bM(\btheta)\in\mathbb{R}^{7\times7}\) is the arm's inertia matrix; the hand axis is decoupled from it and enters only through \(IA_8\) and \(FV_8\) (Appendix~A), so the feasibility conditions of Sec.~IV concern the arm alone while all eight torque equations enter the regression.

The symbolic model and regressor are generated with OpenSYMORO~\cite{OpenSYMORO}, selected for its integrated symbolic identification workflow: from one kinematic description it returns the base-parameter grouping in closed form, so that the relations of Appendix~A can be inspected and regenerated independently of our implementation, together with robot-specific expressions for the inverse-dynamics regressor and the direct dynamic model, which keeps the OLS regressor and the CLIE simulation of Sec.~IV consistent with each other. Pinocchio~\cite{CarpentierEtAl2019Pinocchio} also provides rigid-body dynamics and torque-regressor algorithms; OpenSYMORO was preferred here for the convenience of that single symbolic workflow.
The full grouped model contains 65 parameters; the reduced model used here contains 39, obtained by removing the 18 products of inertia and the 8 Coulomb friction coefficients. Appendix~A gives the base-parameter grouping and the exact composition of both sets.

\input{tables/tab1_param_categories}

Table~\ref{tab:reduced_param_structure} summarizes the reduced parameter structure.
The 25 link-side terms consist of 13 inertia-related base terms and 12 first-moment base terms.
The complete parameter names and final numerical values are listed in Table~\ref{tab:appc}.
Terms without the prefix $M$ are inertias in kg$\cdot$m$^2$, $M$-prefixed terms are first moments in kg$\cdot$m, and the viscous coefficients are in N$\cdot$m$\cdot$s/rad.

\subsection{Why the products of inertia are removed}

The links of CRANE-X7 are bracket-and-plate assemblies close to mirror-symmetric about the plane containing the joint axis, so their products of inertia are expected to be small, and the manufacturer's design table distributed with the URDF~\cite{CRANEX7Website,RTCraneX7Description} puts them between \(0.1\) and \(8\%\) of the largest diagonal entry of the same link; that is a check on the geometry, not an input, since no CAD inertial data enters the identification.
The expectation is borne out where it matters, in the fit itself: fitting the full 65-parameter model to the identification data, the 18 products of inertia together account for \(1.70\%\) of the RMS of the commanded torque, against \(96.12\%\) for the 39 retained terms.
Smallness alone, however, is not a reason to remove them. What makes their recovery doubtful is the third figure in that comparison: the residual that the full 65-parameter rigid-body model leaves is \(27.41\%\) of the torque RMS, sixteen times the contribution being estimated, and it is structured rather than white.
All three shares are RMS values of non-orthogonal contributions taken against the same total, so they are not proportions of it and need not sum to \(100\%\). As Sec.~VII-A discusses, that residual appears to be dominated by encoder-quantization ripple above the closed-loop bandwidth, which the model structurally cannot represent. Whether a \(1.7\%\) contribution can be recovered against such a floor is an empirical question, and Table~\ref{tab:reduction} answers it for this data set.

\input{tables/tab_reduction_evidence}

Table~\ref{tab:reduction} shows the consequence. Four nested models were fitted on each of seven identification trajectories spanning all five families, and each was used to predict the three held-out validation trajectories.
Adding the products of inertia improves the in-sample residual by at most \(0.2\%\), degrades the regressor condition number by between 2 and 93\%, and makes held-out prediction worse on \emph{every} trajectory tested, by \(0.5\) to \(5.9\%\).
The full 65-parameter model is likewise worse than the reduced model on all seven. The three validation records serve every held-out evaluation in this paper, this model-class comparison included; they were never used to select among the 40 candidates or to tune a setting.
The reduction is therefore justified by generalization rather than by an assertion that the removed terms vanish, and the symmetry argument enters only as the reason they are small enough to be lost in the noise floor.

The same table shows that the two removals are not equivalent.
The eight Coulomb friction coefficients cost almost nothing in conditioning and \emph{improve} held-out prediction on six of the seven trajectories, by up to \(0.9\%\).
Their removal cannot be justified on the grounds above, and we retain only viscous friction for a different reason: a more flexible friction model absorbs quantization, timestamp fluctuation, communication delay and drivetrain imperfection into coefficients that then no longer describe friction. That is a robustness choice, and Sec.~VII-D quantifies what it costs.

%% file: tables/tab_dh_parameters.tex
\begin{table}[t]
\centering
\footnotesize
\setlength{\tabcolsep}{5pt}
\renewcommand{\arraystretch}{1.05}
\caption{Modified Denavit--Hartenberg parameters of CRANE-X7 used for symbolic model generation, in the OpenSYMORO convention. \(RL_3 = RL_5 = 0.25~\mathrm{m}\); all other link offsets are zero.}
\label{tab:dh}
\begin{tabular*}{\columnwidth}{@{\extracolsep{\fill}}c c c c c c@{}}
\toprule
\(i\) & \(\gamma_i\) & \(\alpha_i\) & \(d_i\) & \(\theta_i\) & \(r_i\) \\
\midrule
1 & 0 & 0          & 0 & \(\theta_1\) & 0 \\
2 & 0 & \(-\pi/2\) & 0 & \(\theta_2\) & 0 \\
3 & 0 & \(\pi/2\)  & 0 & \(\theta_3\) & \(RL_3\) \\
4 & 0 & \(-\pi/2\) & 0 & \(\theta_4\) & 0 \\
5 & 0 & \(\pi/2\)  & 0 & \(\theta_5\) & \(RL_5\) \\
6 & 0 & \(-\pi/2\) & 0 & \(\theta_6\) & 0 \\
7 & 0 & \(\pi/2\)  & 0 & \(\theta_7\) & 0 \\
\bottomrule
\end{tabular*}
\end{table}

%% file: tables/tab1_param_categories.tex
\begin{table}[t]
\centering
\caption{Reduced 39-parameter structure used for CRANE-X7 dynamic identification.}
\label{tab:reduced_param_structure}
\footnotesize
\setlength{\tabcolsep}{3.5pt}
\renewcommand{\arraystretch}{1.05}
\begin{tabular*}{\columnwidth}{@{\extracolsep{\fill}}l l c r}
\toprule
Group & Parameter type & Unit & Count \\
\midrule
Link-side & Inertia base terms      & $\mathrm{kg\cdot m^2}$       & 13 \\
Link-side & First-moment base terms & $\mathrm{kg\cdot m}$         & 12 \\
Motor-side & Rotor inertia terms    & $\mathrm{kg\cdot m^2}$       & 6  \\
Joint-side & Viscous friction terms & $\mathrm{N\cdot m\cdot s/rad}$   & 8  \\
\midrule
\multicolumn{3}{r}{Reduced model total} & 39 \\
\multicolumn{3}{r}{Original symbolic model} & 65 \\
\bottomrule
\end{tabular*}
\end{table}

%% file: tables/tab_reduction_evidence.tex
\begin{table}[t]
\centering
\scriptsize
\setlength{\tabcolsep}{3pt}
\renewcommand{\arraystretch}{1.0}
\caption{Evidence for the reduced parameterization. Each row fits the four nested models on one identification trajectory at the 40~ms setting and predicts the three held-out validation trajectories (pooled RMSE, N$\cdot$m). PoI: the 18 products of inertia; FS: the 8 Coulomb friction coefficients.}
\label{tab:reduction}
\begin{tabular*}{\columnwidth}{@{\extracolsep{\fill}}l c c c c c c@{}}
\toprule
\multirow{2}{*}{Fit trajectory} & \multicolumn{2}{c}{\(\kappa(\bW_{\mathrm{all}})\)} & \multicolumn{4}{c}{Held-out RMSE} \\
\cmidrule(lr){2-3}\cmidrule(l){4-7}
 & 39 & \(+\)PoI & 39 & \(+\)PoI & \(+\)FS & 65 \\
\midrule
AG02 & 33.3 & 45.1 & \textbf{0.16890} & 0.16983 & 0.16848 & 0.16944 \\
PA04 & 39.0 & 65.0 & \textbf{0.18046} & 0.18391 & 0.17973 & 0.18323 \\
PB08 & 62.4 & 77.2 & \textbf{0.17897} & 0.18096 & 0.17752 & 0.17938 \\
PC08 & 42.7 & 45.8 & \textbf{0.18144} & 0.19210 & 0.17978 & 0.18982 \\
AP03 & 42.7 & 82.5 & \textbf{0.19792} & 0.20340 & 0.19803 & 0.20340 \\
PB04 & 49.9 & 64.9 & \textbf{0.19628} & 0.19859 & 0.19617 & 0.19839 \\
PC02 & 45.6 & 46.7 & \textbf{0.17280} & 0.17373 & 0.17200 & 0.17292 \\
\midrule
\multicolumn{7}{@{}l@{}}{\(+\)PoI worse than 39: \textbf{7 of 7}\quad 65 worse than 39: \textbf{7 of 7}} \\
\multicolumn{7}{@{}l@{}}{\(+\)FS better than 39: 6 of 7} \\
\bottomrule
\end{tabular*}
\end{table}

%% file: sections/04_identification_procedure.tex
\section{Dynamic Parameter Identification Framework}

The joint-level models of Sec.~III only realize the excitation; identification is performed at the robot level with the reduced model \eqref{eq:rigid_dynamics}, as a pipeline of IDIM-based linear regression, ordinary least squares (OLS), conditional semidefinite-programming (SDP) projection, and closed-loop input error (CLIE) refinement, after which the candidate named by the excitation design is checked against the parameter cloud of all candidates, which principal component analysis (PCA) characterizes, and screened by the all-pose PD criterion.

\subsection{IDIM-Based Linear Regression}

Starting from the reduced rigid-body model in \eqref{eq:rigid_dynamics}, the inverse dynamics are rewritten in a form that is linear in the unknown reduced parameter vector \(\bphi\): \begin{equation} \btau = \bW(\btheta,\dot{\btheta},\ddot{\btheta})\,\bphi, \label{eq:idim_linear} \end{equation} where \(\bW(\btheta,\dot{\btheta},\ddot{\btheta})\) is the inverse dynamic identification matrix generated from the symbolic rigid-body model, and \(\bphi\) is the reduced base-parameter vector. 

In the present study, the symbolic regressor is constructed using OpenSYMORO, specifically from its dynamic identification model output \texttt{dim.txt}, so that model generation remains reproducible and consistent with the reduced parameterization described in Sec.~III. 

For each identification trajectory, the measured joint motion and torque data are first preprocessed by anti-aliasing, time alignment, and resampling, and are then substituted into \eqref{eq:idim_linear}. By stacking all samples from a given identification run, the regression model is written as \begin{equation} \btau_{\mathrm{all}} = \bW_{\mathrm{all}}\,\bphi + \bm{\varepsilon}, \label{eq:stacked_regression} \end{equation} where \(\btau_{\mathrm{all}}\) is the stacked torque vector, \(\bW_{\mathrm{all}}\) is the stacked regressor, and \(\bm{\varepsilon}\) is the residual vector.
\subsection{OLS Estimation and Conditional SDP-Based Feasible Projection}

The initial reduced-parameter estimate is obtained by ordinary least squares: 
\begin{equation} \hat{\bphi}_{\mathrm{OLS}} = \argmin_{\bphi} \left\| \btau_{\mathrm{all}} - \bW_{\mathrm{all}}\bphi \right\|_2^2. \label{eq:ols} 
\end{equation} 
Equivalently, the closed-form solution is 
\begin{equation} \hat{\bphi}_{\mathrm{OLS}} = \left( \bW_{\mathrm{all}}^{\mathsf T}\bW_{\mathrm{all}} \right)^{-1} \bW_{\mathrm{all}}^{\mathsf T}\btau_{\mathrm{all}}. \label{eq:ols_closed} 
\end{equation} 
After obtaining \(\hat{\bphi}_{\mathrm{OLS}}\), its inertia matrix is tested for positive definiteness at the configurations the identification trajectory visits. This routing test is deliberately cheap, and Sec.~VI-E measures how much weaker it is than the acceptance audit over \(\Theta\). If the OLS solution passes, it is passed directly to the CLIE stage. Otherwise, an SDP-based feasible projection is applied before CLIE, written as
\begin{equation} 
\begin{aligned} \hat{\bphi}_{\mathrm{SDP}} =&\ \argmin_{\bphi} \left\| \bphi-\hat{\bphi}_{\mathrm{OLS}} \right\|_2^2 \\ \text{s.t.}\quad & \bM(\btheta;\bphi)\succeq \varepsilon_{\mathrm{PD}}\bI, \quad \forall \btheta\in\Theta_{\mathrm c} ,
 \end{aligned}
 \label{eq:sdp_projection}
 \end{equation}
 where \(\bI\) is the identity matrix, \(\varepsilon_{\mathrm{PD}}\) is the positive-definiteness margin, and \(\Theta_{\mathrm c}\subset\Theta\) is a finite constraint set of at most \(200\) configurations drawn from the identification trajectory (Appendix~\ref{app:impl}-E). Positive definiteness over the audit domain \(\Theta\) of Sec.~V-A is decided afterwards by the audit, not inferred from the constraints.
 
 In the present study, \(\varepsilon_{\mathrm{PD}}\) was selected from the discrete set \(\{0.001, 0.002, 0.005, 0.01\}\), and the smallest value whose projected solution passed the positive-definiteness test at every configuration of the record, not only at the constrained ones, was adopted for each trajectory and sampling condition. 
This pre-CLIE margin is a conservative regularization of a rough OLS solution on the sampled configurations, not the margin required of the accepted model.
Running CLIE from each of the four projections of AG02 at \(40~\mathrm{ms}\) changes the final objective by \(0.16\%\) and the iteration count between \(23\) and \(86\) (Appendix~\ref{app:impl}-E). 

In the present implementation, the constrained optimization in \eqref{eq:sdp_projection} was solved in MATLAB using CVX (Appendix~\ref{app:impl}-B). 
 \subsection{CLIE Refinement and Final PD-Feasible Acceptance}

The OLS and SDP stages do not minimize the discrepancy between measured and simulated closed-loop torque; the closed-loop input error (CLIE) stage does.
Let \(\btau_m(t_k)\) denote the measured joint torque vector at time \(t_k\), and let \(\btau_s(t_k;\bphi)\) denote the simulated torque vector generated by the closed-loop forward-dynamics model parameterized by \(\bphi\). Then, the CLIE estimate is defined by 
\begin{equation} 
\hat{\bphi}_{\mathrm{CLIE}} = \argmin_{\bphi} \sum_{k=1}^{N} \left\| \btau_m(t_k) - \btau_s(t_k;\bphi) \right\|_2^2. 
\label{eq:clie} 
\end{equation} 

In implementation the residual is band-limited, decimated to \(40~\mathrm{ms}\) at every identification interval, and masked at torque-clamped samples before it enters the sum (Appendix~\ref{app:impl}-F).

\begin{figure}[t]
\centering
\includegraphics[width=\linewidth]{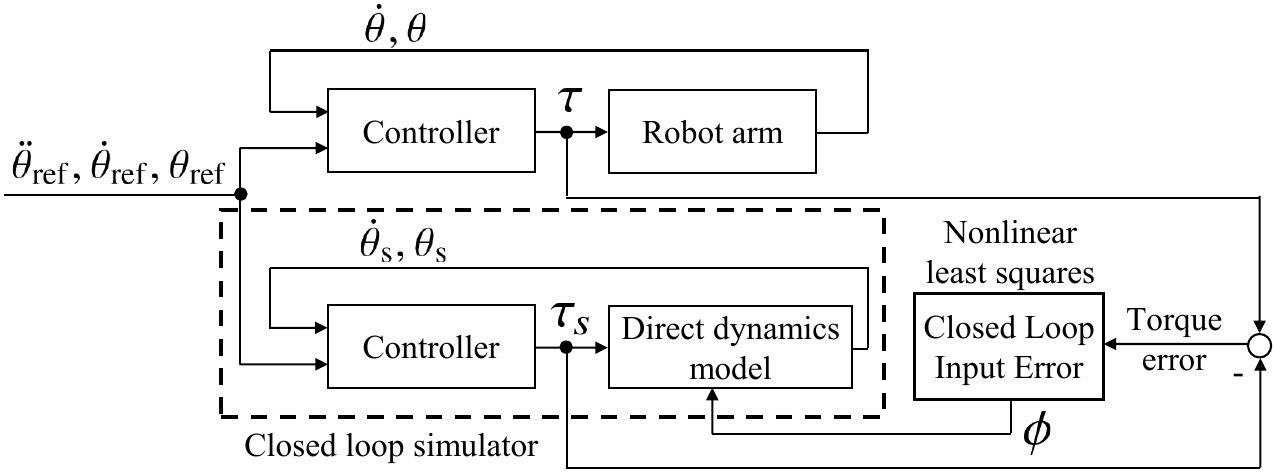}
\caption{
Closed-loop input error (CLIE) refinement.
A candidate parameter vector is evaluated through a closed-loop forward-dynamics simulation using the same fixed FF+FB controller as in the experiment.
The simulated controller input torque is compared with the measured torque, and the parameter vector is updated by nonlinear least squares.
In the diagram, \(\btau\) and \(\btau_s\) are the \(\btau_m(t_k)\) and \(\btau_s(t_k;\bphi)\) of \eqref{eq:clie}.
}
\label{fig:clie_block}
\end{figure}

Figure~\ref{fig:clie_block} illustrates the stage: the direct dynamic model, the OpenSYMORO output \texttt{ddm.txt} of the same reduced model, is embedded in the same fixed FF+FB controller that realized the experimental motions, and the parameter vector is updated by nonlinear least squares on the closed-loop torque-input error, so that CLIE refines the OLS/SDP estimate by reproducing the measured closed-loop realization rather than by fitting the inverse-dynamics regressor alone.

The nonlinear optimization in \eqref{eq:clie} was implemented in MATLAB using \texttt{lsqnonlin} and initialized from the OLS/SDP estimate: 
CLIE started from an arbitrary or non-PD parameter vector can make the forward-dynamics simulation diverge.

In the present experiments, 93 of the 160 OLS estimates were non-PD before the conditional SDP step, as summarized in Table~\ref{tab:routing_summary}.
The solver ran the trust-region-reflective algorithm with a function tolerance of \(10^{-6}\) and every other \texttt{lsqnonlin} setting at its default, including the iteration limit of \(400\); the search range, the \(0.25~\mathrm{ms}\) trapezoidal integration of the simulation, and the parameter bounds are recorded in Appendix~\ref{app:impl}-F. 

Cost is reported as an iteration count rather than as elapsed time: the \(160\) refinements ran concurrently on one workstation, and sixteen of them, solved twice after worker restarts, reproduced their iterates exactly while their elapsed times differed by up to \(61\%\) (Appendix~\ref{app:impl}-F). Each iteration costs \(40\) function evaluations, one per parameter plus the reference point. Half of the \(160\) conditions finish within \(21\) iterations and nine-tenths within \(183\); the route dominates, with medians of \(6\) for O-C and \(43\) for O-S-C, and the sampling interval does not. The complete set cost \(307~\mathrm{h}\) of solver time, an offline cost incurred once (Appendix~\ref{app:impl}-I). Nine refinements, all on the O-S-C route, stopped at the iteration limit with their objectives still decreasing but slowly, by \(0.011\%\) over the final eleven iterations for AP02 at \(10~\mathrm{ms}\), and were left as they are; they are marked with a dagger in Table~\ref{tab:appc}, and the accepted model, which converged on the function tolerance at every interval, is not among them.

Nevertheless, CLIE does not by itself guarantee practical physical feasibility over the entire joint domain. For this reason, final model acceptance is based on an all-pose positive-definiteness check of the inertia matrix: 
\begin{equation} 
\bM(\btheta;\bphi)\succ \bm{0} \qquad \forall \btheta\in\Theta, \label{eq:pd_check} 
\end{equation} 
or, equivalently in numerical implementation, 
\begin{equation} 
\lambda_{\min}\!\left(\bM(\btheta;\bphi)\right) > 0 \qquad \forall \btheta\in\Theta. \label{eq:pd_check_lambda} 
\end{equation} 

In implementation \eqref{eq:pd_check_lambda} is tested on a finite audit set \(\Theta_{\mathrm a}\subset\Theta\) of \(221{,}875\) configurations (Sec.~VI-E); ``PD'' in what follows means that no violation was found on \(\Theta_{\mathrm a}\), and \eqref{eq:pd_check}--\eqref{eq:pd_check_lambda} state the condition aimed at, not a certificate of it.
Positive definiteness is weaker than full physical consistency, a mass distribution that realizes the parameters (Sec.~VII-E); it is the property the two intended uses of the model, acceleration-resolved control and the reaction-force observer, rest on, since both invert \(\bM(\btheta)\), and where \eqref{eq:pd_check_lambda} fails a negative eigenvalue reverses the inertial response along its eigendirection. Sec.~VII-C quantifies that failure and shows that torque-prediction accuracy does not detect it.
If the CLIE solution of the named candidate still violates this criterion, a post-CLIE SDP rescue step is applied:
\begin{equation} 
\begin{aligned} 
\hat{\bphi}_{\mathrm{rescue}} =&\ \argmin_{\bphi} \left\| \bphi-\hat{\bphi}_{\mathrm{CLIE}} \right\|_2^2 \\ \text{s.t.}\quad & \bM(\btheta;\bphi)\succeq \varepsilon_{\mathrm{PD}}\bI, \quad \forall \btheta\in\Theta_{\mathrm r} .
\end{aligned} 
\label{eq:sdp_rescue} 
\end{equation} 

The rescue moves the CLIE solution as little as feasibility requires; applied to a refined solution already near the PD-feasible boundary, its margin is a minimal feasibility constraint rather than the conservative screening margin of the pre-CLIE stage.

The margin and the criterion \eqref{eq:pd_check_lambda} are enforced on different pose sets. The rescue holds \(\bM(\btheta;\bphi)\succeq\varepsilon_{\mathrm{PD}}\bI\) on \(\Theta_{\mathrm r}\), a finite set drawn from \(\Theta\) and enlarged by constraint generation (Appendix~\ref{app:impl}-E), and returns a solution whose smallest eigenvalue on that set equals \(\varepsilon_{\mathrm{PD}}=10^{-3}\) to solver precision; the audit then tests \(\lambda_{\min}>0\) on an independently drawn, much denser set, on which the accepted model returns \(5.1\) to \(9.1\times10^{-4}\) depending on the draw, below the constraint margin but strictly positive, with no violating pose. That gap is why the margin is kept nonzero: it is the room the constraint set leaves for configurations it did not contain. With \(\varepsilon_{\mathrm{PD}}=0\) the constrained eigenvalues sit at zero and nothing is left for the audit to find; the model labelled ``submitted'' in Table~\ref{tab:auditsampling}, the model accepted in the earlier version of this study (arXiv:2605.15949v1, v2), was rescued that way and fails the denser audit of Sec.~VI-E.

The complete trajectory-wise routing of OLS, conditional SDP projection, CLIE refinement, representative selection, and post-CLIE rescue is summarized in Algorithm~\ref{alg:pipeline}. 

\input{algorithms/algo1_pipeline}

%% file: algorithms/algo1_pipeline.tex
\begin{algorithm}[t]
\caption{Trajectory-wise parameter identification and acceptance. The selection in lines \ref{alg:selfirst}--\ref{alg:sellast} is structural and deterministic, and uses no property of the estimates; centrality and conditioning enter in lines \ref{alg:chkfirst}--\ref{alg:chklast} as acceptance checks against the 160-condition study, not as the selector.}
\label{alg:pipeline}
\begin{algorithmic}[1]
\FOR{each trajectory and sampling condition}
  \STATE Construct \(\bW_{\mathrm{all}}\) from the OpenSYMORO model
  \STATE Compute \(\hat{\bphi}_{\mathrm{OLS}}\) by least squares
  \STATE Test PD of \(\hat{\bphi}_{\mathrm{OLS}}\) at the configurations the
  \STATE \quad trajectory visits \emph{(not on \(\Theta\); see Sec.~VI-E)}
  \IF{\(\hat{\bphi}_{\mathrm{OLS}}\) is PD}
    \STATE Set the CLIE initial value to \(\hat{\bphi}_{\mathrm{OLS}}\)
  \ELSE
    \FOR{\(\varepsilon_{\mathrm{PD}} \in \{0.001,0.002,0.005,0.01\}\) in ascending order}
      \STATE Solve the SDP projection
      \IF{the projected solution is PD}
        \STATE Set the CLIE initial value to \(\hat{\bphi}_{\mathrm{SDP}}\)
        \STATE \textbf{break}
      \ENDIF
    \ENDFOR
  \ENDIF
  \STATE Run CLIE from the selected initial value
\ENDFOR
\STATE \textbf{Select the representative candidate:} \label{alg:selfirst}
\STATE \quad trajectory \(\leftarrow\) the unique candidate whose block sequence
\STATE \quad\quad contains every block of the design (Table~S3)
\STATE \quad interval \(\leftarrow\) the coarsest \(T_s\) at which every acceleration
\STATE \quad\quad ramp carries \(\ge 3\) samples \label{alg:sellast}
\STATE \textbf{Accept only if} the candidate is within the central \label{alg:chkfirst}
\STATE \quad quintile of the PCA cloud and \(\kappa(\bW_{\mathrm{all}})\) is within
\STATE \quad the lowest decile of the 160 conditions \label{alg:chklast}
\IF{the selected solution is not PD on \(\Theta\)}
  \STATE Apply the post-CLIE SDP rescue of \eqref{eq:sdp_rescue}
\ENDIF
\STATE Re-audit on an independent, denser pose set
\end{algorithmic}
\end{algorithm}

%% file: sections/05_experimental_conditions.tex
\section{Experimental Conditions}

\subsection{Candidate Trajectory Families}

The candidate identification trajectories are hand-designed joint-space command sequences built from single-joint and adjacent-two-joint primitives under practical joint-range limits, in the line of the classical test motions of Mayeda \textit{et al.}~\cite{Mayeda1984NewIdentification}, rather than numerically optimized waveforms.

\begin{figure}[t]
\centering
\includegraphics[width=0.98\columnwidth]{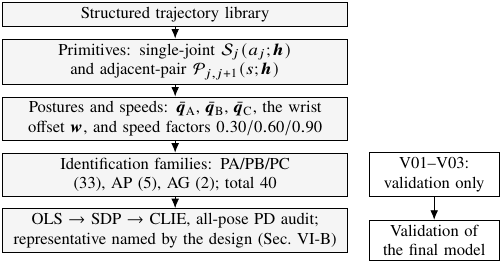}
\caption{PA, PB, PC, AP, and AG form 40 identification candidates; V01--V03 are reserved for validation.}
\label{fig:traj_families}
\end{figure}

Every trajectory is a concatenation of two kinds of block: a single-joint block \(\mathcal S(\bar{\bq})\) of ten two-sided one-axis sweeps at a nominal posture \(\bar{\bq}\), and one or more of three adjacent-pair blocks \(\Pi_\alpha,\Pi_\beta,\Pi_\gamma\), each applying the six adjacent pairs \((1,2),\dots,(6,7)\) at its own set of home postures. The PA, PB, and PC families differ only in the nominal posture supplied to \(\mathcal S\), namely Postures A, B, and C, in increasing order of severity for the inertia-matrix positive-definiteness (PD) audit; the pair blocks are common to the three families. Two sweeps of \(\mathcal S\) place Joint~6 at \(-89^\circ\) or Joint~5 at \(-90^\circ\) as a wrist-posture variant. Table~\ref{tab:traj_families} and Fig.~\ref{fig:traj_families} summarize the 40 identification candidates, and Appendix~\ref{app:traj} specifies the blocks, the posture vectors, the speed factors, and the block sequence of each trajectory completely. Fig.~\ref{fig:g02_commands} in Appendix~\ref{app:traj}-F shows the commanded angles of all eight axes for AG02, the trajectory whose solution is finally accepted; the single-joint blocks occupy the first two thirds of the record and the three pair blocks the last third, which is where several axes move together.
The all-pose PD audit is evaluated over the domain \(\Theta\) given by \(\pm180^\circ\) for Joints~1, 3, 5, and 7, \(\pm90^\circ\) for Joints~2 and 6, and \([-180^\circ,9^\circ]\) for Joint~4. This box is wider than the manufacturer's joint limits on five joints and \(2^\circ\) narrower on Joint~2; the accepted model has no violating configuration over the manufacturer's box either, with a minimum eigenvalue of \(8.67\times10^{-4}\) there and \(3.36\times10^{-3}\) over the omitted two-degree strip. Sec.~VI-E specifies how \(\Theta\) is sampled, and Table~\ref{tab:appc} is the numerical record.

\input{tables/tab2_trajectory_families_v2}

\subsection{Execution Record and Preprocessing Conditions}

All measurements in this study come from a single CRANE-X7 unit, under the same controller and logging conditions. The identification data set uses one recording per trajectory, 31 of the 40 recorded on a single day and the rest over the following three weeks. Five identification trajectories were also executed more than once from a byte-identical command file, at separations from 19 minutes to 26 days (Table~\ref{tab:repeatability}), and each of V01--V03 three times over one to two weeks; those repetitions are what makes run-to-run variation measurable. Sec.~VII-D measures it: refitting each execution changes held-out RMSE by \(0.45\) to \(3.55\%\), \(1.25\%\) for AG02, and that is the scale against which every comparison below is read.

The measured data were processed by time-axis correction, anti-alias filtering, and resampling to effective sampling intervals of \(10\), \(20\), \(40\), and \(80~\mathrm{ms}\).
For each interval, only the target-grid samples were retained.
Off-grid samples were not reused as phase-shifted training records because doing so added highly correlated samples
and tended to worsen regressor conditioning.
The interval used for the accepted model is fixed in Sec.~VII-D from the sampling of the acceleration ramps, and Sec.~VI-B reports the checks applied to it.

\subsection{Validation Conditions}

The held-out validation trajectories V01--V03 were kept separate from the 40 identification candidates. They combine simultaneous seven-arm-joint plus hand-axis motions and four-arm-joint-group plus hand-axis motions. Only V03 is shown qualitatively, in Fig.~\ref{fig:validation_traj} of Appendix~\ref{app:traj}-G; quantitative results are summarized in Table~\ref{tab:validation_metrics}.

%% file: tables/tab2_trajectory_families_v2.tex
\begin{table}[t]
\centering
\caption{Structured trajectory families used in the experiments. Appendix~\ref{app:traj} specifies the single-joint block, the three adjacent-pair blocks, and the block sequence of every trajectory.}
\label{tab:traj_families}
\footnotesize
\setlength{\tabcolsep}{2.4pt}
\begin{tabular}{c c p{0.50\columnwidth} c}
\toprule
Family & Labels & Description & Count \\
\midrule
PA & PA01--PA11 & \(\mathcal S(\bar{\bq}_{\mathrm A})\) followed by one or all of the pair blocks, at three speed levels. & 11 \\
PB & PB01--PB11 & The same eleven combinations with \(\mathcal S(\bar{\bq}_{\mathrm B})\). & 11 \\
PC & PC01--PC11 & The same eleven combinations with \(\mathcal S(\bar{\bq}_{\mathrm C})\). & 11 \\
AP & AP01--AP05 & Pair blocks used standalone: \(\Pi_\alpha\), \(\Pi_\gamma\), \(\Pi_\beta\) at the high level, and \(\Pi_\gamma\) repeated at the two lower levels. No primitive moves more than two arm joints. AP03 moves no more than two at any instant; the others contain one transit that moves a third (Appendix~\ref{app:traj}-E). & 5 \\
AG & AG01--AG02 & Aggregated: the three pair blocks (AG01), and the three single-joint blocks followed by the three pair blocks (AG02). & 2 \\
V & V01--V03 & Held-out multi-axis validation trajectories. & 3 \\
\bottomrule
\end{tabular}
\end{table}

%% file: sections/06_results.tex
\section{Results}

\subsection{PCA-Based Evolution of the Parameter Clouds}

\begin{figure*}[t]
\centering
\includegraphics[width=\linewidth]{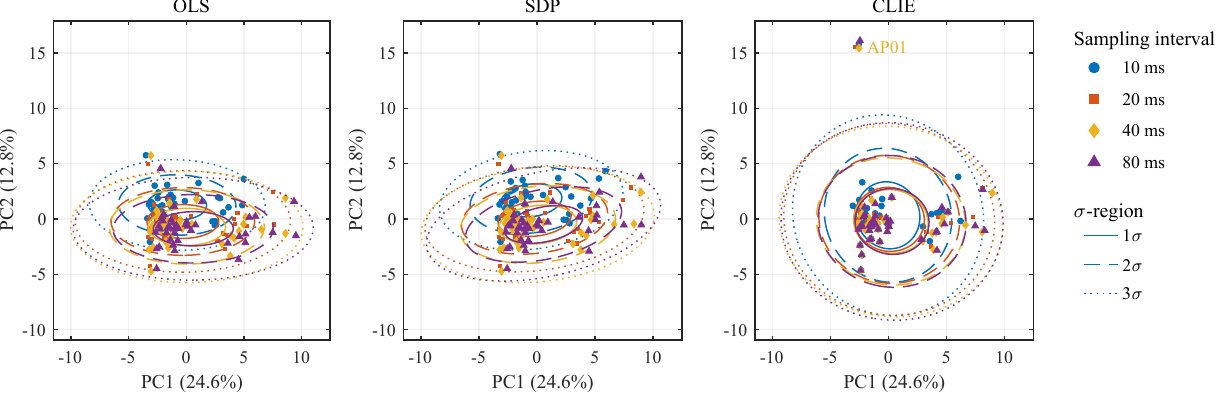}
\caption{PCA-based visualization of parameter clouds from OLS, SDP, and CLIE. Each point is one of the \(160\) conditions, projected onto the first two principal components of the pooled, standardized set of all \(480\) solutions (\(24.6\%\) and \(12.8\%\) of its variance); the ellipses are the \(1\sigma\), \(2\sigma\), and \(3\sigma\) covariance contours of each sampling interval's points in this plane. Total variance, as quoted in the text, is the sum over the 39 standardized coordinates of the squared deviations from each cloud's own mean, and the median distance is measured from that mean.}
\label{fig:pca_clouds}
\end{figure*}

Figure~\ref{fig:pca_clouds} projects the 39-dimensional parameter vectors onto a common PCA basis; the figures quoted here are taken in the full 39 dimensions, because the plotted plane carries only \(37.4\%\) of the variance and a cloud can appear to shrink in it while growing overall. The two stages act in opposite directions. SDP \emph{expands} the cloud: each infeasible candidate is moved to the boundary of its own constraint set, \(\Theta_{\mathrm c}\) and \(\varepsilon_{\mathrm{PD}}\) differing between candidates, and in this data set those moves spread the cloud, which a projection onto one common convex set, being non-expansive, could not; the total variance rises from \(5520\) to \(6293\) and the median distance of the points from the cloud's own centre from \(4.761\) to \(5.062\). CLIE then contracts it: the median distance falls to \(4.723\), below the OLS cloud it started from, and although the total variance stays at \(6435\), that figure is set by the single candidate AP01, without which CLIE is the smallest cloud on both measures, at a total variance of \(4806\) against \(5157\) for OLS and \(5946\) for SDP. AP01, weakly excited, remains separated from the main cloud after CLIE and dominates the visual impression of the figure; the record behind it is Table~\ref{tab:appc}.

\subsection{Representative Candidate Selection, Routing, and PD Rescue}

Table~\ref{tab:routing_summary} summarizes the routing test of Sec.~IV-B over the 160 conditions, O-C and O-S-C denoting OLS\(\rightarrow\)CLIE and OLS\(\rightarrow\)SDP\(\rightarrow\)CLIE, and Table~\ref{tab:route_examples} in Appendix~\ref{app:routing} compares AP01 with AG02 by regressor condition number \(\kappa(\bW_{\mathrm{all}})=\sigma_{\max}(\bW_{\mathrm{all}})/\sigma_{\min}(\bW_{\mathrm{all}})\), route, and SDP margin.

The accepted representative model was obtained from AG02 at \(40~\mathrm{ms}\). Centrality alone cannot select it: AG02 at \(40~\mathrm{ms}\) places 21st under a Euclidean metric on the standardized 39 dimensions and between 6th and 86th under Mahalanobis metrics with 2 to 39 principal components, and four different candidates are the most central under four of the metrics. The rule that names it is structural, needs no statistics, and was formalized after the choice had been made (Appendix~\ref{app:impl}-H). Among the 40 candidates, AG02 is the unique trajectory whose block sequence contains every block of the design: the single-joint block at all three nominal postures, and all three adjacent-pair blocks (Table~\ref{tab:trajcomp}). It is the maximal element of the excitation set, and the only candidate that excites every posture family. The sampling interval follows from the excitation alone: \(40~\mathrm{ms}\) is the coarsest interval tested at which every acceleration ramp still carries at least three samples (Sec.~VII-D). Neither condition uses held-out prediction error, so the two fix the candidate without reference to any estimate; the sub-\(1\%\) held-out improvement beyond \(40~\mathrm{ms}\) is a descriptive check reported in Appendix~\ref{app:validation}-B, not a selection criterion.

Centrality and conditioning then enter as checks rather than as selectors, and both are passed: the regressor condition number of AG02 at \(40~\mathrm{ms}\) is \(33.3\), sixth lowest of the 160 conditions and the lowest among trajectories that span all three postures, and the candidate sits in the central fifth of the cloud under four of the six centrality metrics. Its passage through OLS, the pre-CLIE SDP projection, CLIE, and the post-CLIE rescue is summarized in Table~\ref{tab:g02_transition}, with the PD verdict after each stage. The CLIE solution fails the all-pose PD criterion, so a post-CLIE SDP rescue was applied; Table~\ref{tab:g02_rescue_delta} in Appendix~\ref{app:routing} shows that the correction is localized: four entries carry it, none moves by more than \(8.3\times10^{-4}~\mathrm{kg\cdot m^2}\), three move by less than one standard error of their OLS estimate, and the fourth, \(ZZR_6\), is itself within \(1.2\) standard errors of zero in that fit.

\input{tables/tab3_routing_summary}
\input{tables/tab5_g02_transition}

\subsection{Validation on Held-Out Trajectories}

After representative selection and PD rescue, the final model was evaluated on V01--V03, which were not used for identification or representative selection. Table~\ref{tab:validation_metrics} shows that the rescue does not cost accuracy. It slightly improves both the root-mean-square error (RMSE) and the mean normalized error (MNE) of the CLIE solution it was applied to, on all three records, by \(0.1\) to \(0.8\%\).
RMSE is taken over all eight axes and all samples of a record. MNE is the per-axis RMSE divided by the peak-to-peak measured torque of that axis on the same record, averaged over the eight axes; it is dimensionless, and unlike RMSE it weights the light distal axes equally with the heavily loaded proximal ones, which is why the two metrics order V01--V03 differently in Table~\ref{tab:validation_metrics}.

The held-out set is deliberately unlike the excitation set. All three validation trajectories move the seven arm joints simultaneously, which no identification trajectory does: across the 40 command files no segment commands more than three arm joints at once, and in the records measured in Table~\ref{tab:valseverity} at most five are simultaneously above its threshold. Their amplitude is limited to \(\pm45^\circ\) by torque margin and by collision with the desk mounting, and the short travel leaves little constant-velocity phase, so \(\bM(\btheta)\ddot\btheta\) carries a larger share of the predicted-torque RMS than on the accepted trajectory, \(76.9\%\) on V01 and \(53.8\%\) on V02 against \(32.9\%\) on AG02 (Table~\ref{tab:valseverity}), though not larger than on every identification trajectory: AP01 is higher, and V03 is close to AG02. Together with the simultaneous seven-joint motion this makes validation a complementary test of the inertia matrix rather than a rerun of the excitation at lower speed, and since the Coriolis term adds no parameter beyond those in \(\bM(\btheta)\) and only the two shoulder first moments and the viscous coefficients lie outside it, most of the model is covered. Appendix~\ref{app:traj}-G quantifies all of this and states the one regime, the highest joint speeds, that V01--V03 do not cover.

\begin{figure}[t]
\centering
\includegraphics[width=0.85\columnwidth]{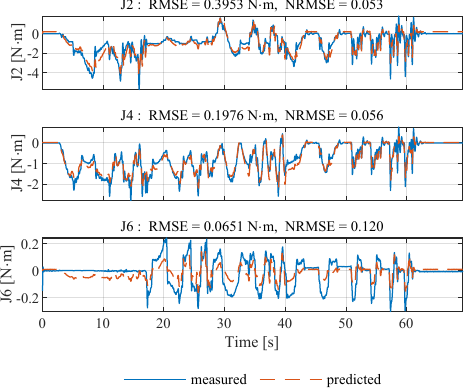}
\caption{Measured and predicted torque of the accepted model on the held-out trajectory V03 for J2, J4, and J6, the pitch axes that carry the gravity load. RMSE is per axis on this record; NRMSE is that RMSE over the axis's peak-to-peak measured torque, the per-axis quantity that MNE averages. Table~\ref{tab:perjoint} in Appendix~\ref{app:validation} gives all eight axes.}
\label{fig:validation_torque}
\end{figure}

Fig.~\ref{fig:validation_torque} shows measured and predicted torques on V03 for Axes~2, 4, and 6, the pitch axes that carry the gravity load and span the loading range, and Table~\ref{tab:perjoint} in Appendix~\ref{app:validation} reports the error on all eight axes. The absolute error follows the torque, from about \(0.40~\mathrm{N\cdot m}\) on Axis~2 to \(0.02\) on the hand; the relative error runs the other way, from \(3.5\%\) of the peak-to-peak range on Axis~1 to \(13.2\%\) on Axis~7, so part of the residual does not scale with the signal, as one expects of encoder quantization, current resolution, and unmodelled drivetrain effects, and the light distal axes pay for it; the figure is representative of the model, not of its worst case, which Table~\ref{tab:perjoint} supplies.

\input{tables/tab7_validation_metrics}

\subsection{What Each Stage Is Worth on Held-Out Data}

Table~\ref{tab:ladder} scores the four stages identically, same fit trajectory, preprocessing, 39 parameters, held-out records, and metric, to show what the pipeline buys over a simpler alternative.

\input{tables/tab_baseline_ladder}

On held-out torque prediction the full pipeline differs little from ordinary least squares on these records. For AG02 the accepted model reaches \(0.16755~\mathrm{N\cdot m}\) against \(0.16890\) for OLS alone, an improvement of \(0.8\%\); across nine fit trajectories CLIE improves on eight, by between \(0.0\) and \(1.3\%\). These improvements are of the same size as the run-to-run variation of Sec.~VII-D, which is the scale for reading them rather than a significance threshold, so no accuracy advantage is attributed to the later stages.

Two entries carry more information than the rest. The rescue step, which exists to restore feasibility and could reasonably be expected to cost accuracy, instead produces the best value in the table; feasibility is obtained here for nothing. And AP01 is the sole trajectory on which CLIE makes prediction markedly worse, by \(20.4\%\). AP01 is the weakly excited pairs-only trajectory that also sits apart from the cloud in Fig.~\ref{fig:pca_clouds}, so the three observations agree.

A second scale makes the same point from the other side: evaluating the single accepted model on the three executions of each validation trajectory gives RMSE standard deviations of \(4.2\), \(1.2\), and \(2.1\%\) on V01, V02, and V03, with nothing changed but the recording, larger than the differences between the rungs of Table~\ref{tab:ladder}.

For AG02, the full pipeline reduces pooled held-out RMSE by \(0.8\%\) relative to OLS while producing a model that passes the independent PD audit; Sec.~VII-C shows why that feasibility check matters once the inertia matrix is inverted.

\subsection{The All-Pose Audit: Sampling and What It Can Miss}

The term \emph{all-pose} describes the intent, not the implementation: positive definiteness is checked on a finite set of configurations, whose domain, size, and concentration decide what the check is worth.

The audit set contains \(221{,}875\) configurations over the domain \(\Theta\) of Sec.~V-A, assembled in three parts. A structured grid of \(1\times5\times5\times7\times5\times5\times5\) points, \(21{,}875\) in total, covers the box uniformly and is deterministic; Joint~1 is held at its centre because \(\bM(\btheta)\) does not depend on it. A further \(150{,}000\) points are drawn uniformly over the same box. The remaining \(50{,}000\) are drawn from the smaller region where violations were found to concentrate: Joint~2 above \(30^\circ\), Joint~4 between \(-160^\circ\) and \(-75^\circ\), and Joint~6 above \(55^\circ\), the folded-arm posture. That region is \(2.9\%\) of the domain by volume, so it ends up sampled about eleven times more densely than the rest.

The three stages do not share a pose set. The SDP projection and the rescue constrain \(\bM(\btheta;\bphi)\succeq\varepsilon_{\mathrm{PD}}\bI\) on the much smaller sets \(\Theta_{\mathrm c}\) and \(\Theta_{\mathrm r}\) of \eqref{eq:sdp_projection} and \eqref{eq:sdp_rescue}, whereas the audit is a denser draw used only to accept or reject, its random part drawn afresh after the model is fixed (seeds 1 to 5) and sharing with \(\Theta_{\mathrm r}\) only 64 grid configurations that coincide, up to \(\theta_1\), with constrained corners (Appendix~\ref{app:impl}-E). The routing of Table~\ref{tab:routing_summary} is a third set again, the configurations each identification trajectory actually visits.

How much that last choice matters is measurable. Re-evaluating all 160 OLS estimates on both criteria, \(69\) are positive definite at the configurations their own trajectory visits but only \(30\) are positive definite over \(\Theta\). Thirty-nine therefore pass the test the routing applies and fail the one that decides whether the model is usable. The reverse did not occur; for the conditions themselves it cannot, since the trajectory configurations lie inside \(\Theta\), although with finite samples on both sides it could. The re-evaluation uses the stored three-digit estimates and disagrees with the run-time routing on two marginal conditions (Appendix~\ref{app:routing}). Checking feasibility only where the data were collected is thus not a conservative approximation of checking it over the workspace but a different and much weaker test, which is why the acceptance set is kept separate and wider.

The consequence is that both the margin and, when the set is too small, the verdict itself depend on the draw. Repeating the audit of the accepted model five times with different seeds and the identical construction returns the same verdict every time, PD with no violating configuration, but a minimum eigenvalue ranging from \(5.09\times10^{-4}\) to \(9.08\times10^{-4}\), a factor of \(1.8\). A reported margin is therefore a property of a sampling and means little without the construction and the sample count that produced it. Differences of a few tens of per cent between pose sets sit inside that spread.
The verdict is sturdier than the margin, but only above a certain sample size. Table~\ref{tab:auditsampling} evaluates both models under both audits: on the sparse audit of the earlier version the submitted model was called PD by three seeds out of five, whereas under the audit described here it is non-PD on all five, at \(248\) to \(279\) of the \(221{,}875\) configurations; the accepted model passes both audits on all five seeds.

Violations are reported as counts rather than as fractions: the oversampled block finds them at fourteen times the rate of the uniform draw for the submitted model and five times for the raw OLS estimate, as it is built to do, so a fraction over the pooled set would overstate how much of the domain is infeasible. Where a measure rather than a detection is needed, in Table~\ref{tab:pdnecessity}, it is taken from the uniform draw alone.

\input{tables/tab_audit_sampling}

The violating configurations lie in the shoulder-up, elbow-folded region that a human operator drives through routinely, which is why the audit oversamples it; Sec.~VII-C shows what they cost once the inertia matrix is inverted. An audit is only as good as the region it samples densely, and where violations concentrate is itself something to be measured rather than assumed.

%% file: tables/tab3_routing_summary.tex
\begin{table}[t]
\centering
\footnotesize
\renewcommand{\arraystretch}{0.95}
\caption{Routing summary before CLIE refinement.}
\label{tab:routing_summary}
\begin{tabular*}{\columnwidth}{@{\extracolsep{\fill}}c c c c@{}}
\toprule
\(T_s\) [ms] & PD-OLS \(\rightarrow\) CLIE & NPD-OLS \(\rightarrow\) SDP \(\rightarrow\) CLIE & Total \\
\midrule
10 & 6  & 34 & 40 \\
20 & 14 & 26 & 40 \\
40 & 22 & 18 & 40 \\
80 & 25 & 15 & 40 \\
\bottomrule
\end{tabular*}
\end{table}

%% file: tables/tab5_g02_transition.tex
\begin{table}[t]
\centering
\footnotesize
\setlength{\tabcolsep}{3pt}
\renewcommand{\arraystretch}{0.95}
\caption{Stage-wise evolution of the representative candidate AG02.}
\label{tab:g02_transition}
\begin{tabularx}{\columnwidth}{@{}>{\raggedright\arraybackslash}p{0.18\columnwidth}
    c@{\hspace{5mm}}c@{\hspace{5mm}}c@{\hspace{5mm}}Y@{}}
\toprule
Stage & PD & Route & \(\varepsilon_{\mathrm{PD}}\) & Remark \\
\midrule
OLS       & No  & --    & --    & Initial estimate \\
SDP       & Yes & O-S-C & 0.001 & Applied: OLS was not PD \\
CLIE      & No  & --    & --    & Refinement loses PD again \\
PD rescue & Yes & --    & 0.001 & Final accepted model \\
\bottomrule
\end{tabularx}
\vspace{0.6mm}
{\footnotesize\raggedright The earlier version solved this projection with \(\varepsilon_{\mathrm{PD}}=0\), positive semidefiniteness only, as its table reported; the present rescue uses \(10^{-3}\), and Sec.~IV-C explains why the margin matters.\par}
\end{table}

%% file: tables/tab7_validation_metrics.tex
\begin{table}[t]
\centering
\scriptsize
\renewcommand{\arraystretch}{0.92}
\caption{Validation metrics on the held-out trajectories V01--V03 at the 40~ms setting, for the accepted model and for the CLIE solution it was rescued from. RMSE and MNE are defined in Sec.~VI-C; the last column of Table~\ref{tab:perjoint} is the same MNE in per cent. Pooling the three records rather than averaging them gives the rescue entry of Table~\ref{tab:ladder}, \(0.16755~\mathrm{N\cdot m}\).}
\label{tab:validation_metrics}
\begin{tabular*}{\columnwidth}{@{\extracolsep{\fill}}l c c c c@{}}
\toprule
& \multicolumn{2}{c}{RMSE [N$\cdot$m]} & \multicolumn{2}{c}{MNE [--]} \\
\cmidrule(lr){2-3}\cmidrule(l){4-5}
Trajectory & CLIE & Rescue & CLIE & Rescue \\
\midrule
V01  & 0.16710 & 0.16570 & 0.07749 & 0.07702 \\
V02  & 0.17083 & 0.16978 & 0.08844 & 0.08815 \\
V03  & 0.16740 & 0.16697 & 0.06763 & 0.06753 \\
\midrule
Mean & 0.16844 & 0.16748 & 0.07785 & 0.07757 \\
\bottomrule
\end{tabular*}
\end{table}

%% file: tables/tab_baseline_ladder.tex
\begin{table}[t]
\centering
\footnotesize
\setlength{\tabcolsep}{3pt}
\renewcommand{\arraystretch}{1.05}
\caption{Each stage of the pipeline scored the same way: pooled RMSE on the held-out set V01--V03 at \(40~\mathrm{ms}\), in N$\cdot$m, from the same trajectory, the same preprocessing and the same 39 parameters. A dash means the stage does not occur on that route. The rescue exists only for the accepted model.}
\label{tab:ladder}
\begin{tabular*}{\columnwidth}{@{\extracolsep{\fill}}l c c c c c c@{}}
\toprule
Fit trajectory & Route & OLS & SDP & CLIE & Rescue & \(\Delta\) \\
\midrule
AG02 & O-S-C & 0.16890 & 0.16781 & 0.16830 & \textbf{0.16755} & \(-0.8\%\) \\
AG01 & O-C   & 0.17308 & ---     & 0.17240 & --- & \(-0.4\%\) \\
PA03 & O-C   & 0.17390 & ---     & 0.17285 & --- & \(-0.6\%\) \\
PA07 & O-S-C & 0.17619 & 0.17839 & 0.17393 & --- & \(-1.3\%\) \\
PB01 & O-C   & 0.17286 & ---     & 0.17281 & --- & \(-0.0\%\) \\
PB08 & O-C   & 0.17897 & ---     & 0.17870 & --- & \(-0.2\%\) \\
PC02 & O-C   & 0.17280 & ---     & 0.17273 & --- & \(-0.0\%\) \\
PC08 & O-C   & 0.18144 & ---     & 0.18127 & --- & \(-0.1\%\) \\
AP01 & O-C   & 0.17478 & ---     & 0.21035 & --- & \(+20.4\%\) \\
\bottomrule
\end{tabular*}
\vspace{0.6mm}
{\footnotesize\raggedright \(\Delta\) is the change from OLS to the last stage available for that trajectory, negative meaning better. The run-to-run variation of Sec.~VII-D is \(1.25\%\) for AG02; every entry except AP01 is of that size or smaller.\par}
\end{table}

%% file: tables/tab_audit_sampling.tex
\begin{table}[t]
\centering
\scriptsize
\setlength{\tabcolsep}{3pt}
\renewcommand{\arraystretch}{1.05}
\caption{What the audit's pose sampling decides. Each row repeats the same audit five times with different random seeds; the deterministic grid component is identical across seeds. Violations are given as counts, not fractions, because the two audits differ in size and the lower one deliberately oversamples. The submitted version of this paper accepted its model on the upper audit, whose verdict is seed-dependent.}
\label{tab:auditsampling}
\begin{tabular*}{\columnwidth}{@{\extracolsep{\fill}}l l c c c@{}}
\toprule
Audit & Model & \(\lambda_{\min}\) \([10^{-3}]\) & Violating & PD \\
& & min\,--\,max & poses & \\
\midrule
\multicolumn{5}{@{}l}{\emph{Submitted: \(3^7\) grid \(+\) \(5{,}000\) uniform \(=7{,}187\) poses}}\\
& submitted & \(-0.66\) -- \(+0.81\) & \(0\) -- \(2\) & 3 of 5 \\
& accepted  & \(+1.80\) -- \(+3.09\) & \(0\)          & 5 of 5 \\
\addlinespace[1mm]
\multicolumn{5}{@{}l}{\emph{This paper: grid \(+\) \(150{,}000\) uniform \(+\) \(50{,}000\) focused \(=221{,}875\)}}\\
& submitted & \(-2.22\) -- \(-1.80\) & \(248\) -- \(279\) & 0 of 5 \\
& accepted  & \(+0.51\) -- \(+0.91\) & \(0\)              & 5 of 5 \\
\bottomrule
\end{tabular*}
\vspace{0.6mm}
{\footnotesize\raggedright ``Submitted'' is the model accepted in the earlier version of this study, arXiv:2605.15949v1, v2; ``accepted'' is the model reported here. The \(3^7\) grid on its own accepts the submitted model at \(+3.65\times10^{-3}\) with no violating pose, because the violating region is thinner than the lattice step.\par}
\end{table}

%% file: sections/07_discussion.tex
\Needspace*{5\baselineskip}
\section{Discussion}

\subsection{Reduced Modeling and Identifiability in the Low-Cost Setting}

The reduction from 65 to 39 parameters is motivated by the sensing, communication, and drivetrain constraints of CRANE-X7: retaining every nominal rigid-body parameter amplifies numerical sensitivity and buries physical variation in noise-driven fluctuation. Carrying the products of inertia degrades the regressor condition number on every trajectory tested, by \(2\) to \(93\%\) (Table~\ref{tab:reduction}), and made CLIE markedly slower; even at 39 parameters \(\kappa\) ranges from \(31.6\) to \(182.8\) across the 160 conditions of Table~\ref{tab:appc} and exceeds 100 on 31 of them, so the margin the reduction buys is worth having.
Feasibility-constrained identification of industrial arms~\cite{GazEtAl2019Panda,XuEtAl2020KUKAiiwa,HuangKeZhangOta2023TRO} is the closest prior work. It imposes a per-link inertia LMI inside every iteration of the estimator, in the line opened by Sousa and Cortes\~ao~\cite{SousaCortesao2014LMI} and completed by the pseudo-inertia condition of Wensing \textit{et al.}~\cite{WensingPhysicalConsistency}, as in the hybrid least-squares method validated on the UR5 and the Panda~\cite{HuangKeZhangOta2023TRO}. That condition is built from standard parameters, which a base-parameter model carries only through the groupings of Appendix~A; this paper tests the weaker condition its two intended uses need, positive definiteness of \(\bM(\btheta)\) by the audit of \eqref{eq:pd_check_lambda}, and leaves the link-level reconstruction to Sec.~VII-E.

Friction is kept viscous for the same reason. The hybrid approach replaces Coulomb-viscous friction by a Stribeck model and fits the remaining torque with a neural network~\cite{HuangKeZhangOta2023TRO}, which suits an industrial arm whose unmodelled torque is friction. Here the residual floor is \(27\%\) of torque RMS with a spectrum consistent with encoder-quantization ripple above the closed-loop bandwidth, and Coulomb terms improve held-out prediction by at most \(0.9\%\), which supports retaining the simpler viscous model for the present data.

\subsection{Structured Excitation Design}

That the restriction to at most two moving joints costs nothing in identifiability is verified rather than assumed. Under the reduced parameterization the geometric condition of Mayeda \textit{et al.}~\cite{Mayeda1984NewIdentification} holds for CRANE-X7 (Appendix~\ref{app:traj}-A), and AP03, the pair block \(\Pi_\beta\) used standalone, no segment of which moves more than two arm joints, already gives a 39-column regressor of full rank 39, with \(\kappa=42.7\) at the 40~ms setting against 33.3 for AG02, which also contains the single-joint blocks; AP01 and AP02 give 46.2 and 48.9. The single-joint blocks therefore improve conditioning rather than supply missing rank, and the 24 three-joint segments in the set are posture changes between primitives, not excitation (Appendix~\ref{app:traj}-E).

Rank is a weak test, and held-out prediction answers the question worth asking, how far adjacent-pair excitation gets on its own. Scored exactly as in Table~\ref{tab:ladder}, AG01, the three pair blocks concatenated with no single-joint block, reaches \(0.17308~\mathrm{N\cdot m}\) against \(0.16890\) for AG02, a gap of \(2.5\%\), about twice the run-to-run variation of Sec.~VII-D; one pair block alone reaches \(0.17478\) for \(\Pi_\alpha\), \(0.17740\) to \(0.18051\) for \(\Pi_\gamma\) at its three speed factors, and \(0.19792\) for \(\Pi_\beta\), \(3.5\) to \(17.2\%\) behind AG02. Adjacent-pair excitation is therefore not a fallback on this arm, and only \(\Pi_\beta\) is markedly worse. AP01, weakly excited in Sec.~VI-A yet the best of the five here, shows that OLS prediction is carried by the directions the excitation constrains, whereas its position in the cloud and the \(20.4\%\) that CLIE costs it (Table~\ref{tab:ladder}) are set by the directions it does not.

Conditioning does not order any of this. Over the 40 candidates at \(40~\mathrm{ms}\), \(\kappa\) runs from \(31.6\) to \(172.2\) and held-out RMSE from \(0.15737\) to \(0.19792\), yet the two rank the candidates at a Spearman \(\rho\) of \(-0.216\) (\(p=0.18\)), and among the thirty with two-digit conditioning at \(+0.044\) (\(p=0.82\)); AP03 has the lowest \(\kappa\) of the AP family, \(42.7\), and by a wide margin its worst held-out error. Within this candidate set, once a regressor is not ill-conditioned, its conditioning alone did not reliably rank held-out prediction, which supports the use of \(\kappa\) in Sec.~VI-B as an acceptance check rather than a selector.

Ranked on held-out prediction, AG02 is fifth of the forty, \(7.3\%\) behind PB07, the worst-conditioned candidate in the set; that ranking is not used for selection, since choosing on V01--V03 would spend the held-out set and the ordering is not stable across intervals.

Classical Fourier-series excitation remains the reference design~\cite{SweversEtAl1996MSSP,SweversEtAl1997TRA,Park2006Robotica}. On this platform a simultaneous multi-axis waveform must also respect finite torque margins, unevenly distributed across the arm, and the workspace of a hanging desk mount, so it has to be slowed until the most loaded axis is safe; the structured primitives assign a speed factor per block instead, and were chosen as the reproducible alternative. A quantitative comparison against optimized excitation has not been run.

\subsection{What the Positive-Definiteness Audit Buys}

Fitting a torque record well and being physically feasible are different criteria: accuracy and centrality are properties of the observed data, whereas feasibility must hold over a domain the data never visited. The reason to insist on it is that the two things this model is for on this platform both invert \(\bM(\btheta)\), and both fail silently when the inversion is ill-posed, while the metric normally used to judge an identified model does not notice. Table~\ref{tab:pdnecessity} measures both failures on four infeasible models and on the accepted one.

\input{tables/tab_pd_necessity}

The first use is acceleration-resolved control. Shaping the closed loop with an identified inertia matrix \(\hat\bM(\btheta)\) --- the output of Sec.~IV, not the constant diagonal realization matrix of \eqref{eq:fffb_command} --- gives error dynamics governed by \(\bA=\bM^{-1}\hat\bM\), where the accepted model stands in for the true \(\bM\), which is not measured; along an eigendirection with eigenvalue \(\mu\) the characteristic polynomial of this fixed-configuration, linearized loop is \(s^{2}+\mu K_d s+\mu K_p\), so a negative \(\mu\) puts a root in the right half-plane whatever the gains are. The growth rates quoted below are for \(K_p=1000~\mathrm{s^{-2}}\) and \(K_d=48~\mathrm{s^{-1}}\), a representative acceleration-resolved choice of the same order as the realization gains of Table~\ref{tab:realization}, which are a different layer of the system; only the rates depend on that choice, not the instability. The submitted model does so over \(0.025\%\) of the domain, with the resulting divergence growing on a \(10~\mathrm{ms}\) time constant, and the unrescued OLS solution over \(0.73\%\) at \(3~\mathrm{ms}\).

The second is the reaction-force observer used for bilateral control on this arm~\cite{Yamane2025FastBilateral}, whose acceleration-domain form inverts the same \(\hat\bM(\btheta)\) to convert applied torque into an expected acceleration and reports the mismatch as external torque. Its scalar gain is \(1/\lambda_i\) along the \(i\)th eigendirection of \(\hat\bM\), \(\lambda_i\) being the corresponding eigenvalue. For the accepted model that gain stays bounded over the audit set; for the infeasible models an eigenvalue crosses zero, so the gain passes through a pole and changes sign. The damaging consequence is directional rather than merely large: at each model's worst configuration, between \(43\) and \(68\%\) of \(2{,}000\) random unit torque directions are mapped by \(\hat\bM^{-1}\) to an acceleration more than \(90^\circ\) from the one the accepted model gives, so the inertia inversion at the core of the observer points against the motion it is meant to explain. This fixed-configuration analysis models the inversion, not the observer's filter or its subtraction of the commanded torque.

The last two columns show why this needs a separate acceptance test. For random accelerations at the worst configuration the infeasible models differ from the accepted one by only \(2.3\) to \(6.6\%\) in predicted torque; along the offending eigendirection the same models are wrong by \(244\) to \(637\%\), with reversed sign in every case. A model can therefore pass every validation table in Sec.~VI and still be unusable for the purpose it was identified for.

\subsection{What Reproduces, and to What Precision}

What is reproducible, and at what precision, differs at three levels: rerunning the pipeline on the same recording, which is exact by construction; rerunning the same trajectory on the robot; and changing a processing choice. The last two are measured here.

\input{tables/tab_repeatability}

Table~\ref{tab:repeatability} reports the second level. Five trajectories were executed more than once from a byte-identical command file on the same unit, with separations from 19 minutes to 26 days. Fitting each execution independently and predicting the held-out set changes the RMSE by \(0.45\) to \(3.55\%\). Elapsed time does not order this: the three PB01 executions spanning five days agree to \(0.45\%\), while the three AP03 executions, two of them 34 minutes apart, differ by \(3.55\%\). Sample count does. The three long trajectories supply between \(3124\) and \(4241\) samples per axis and stay within \(1.25\%\); the two short adjacent-pair trajectories supply about \(500\) and are the two worst. What is being measured is therefore consistent with estimation noise rather than drift of the hardware, and it is the variation against which the comparisons of Sec.~VI are read. For AG02, the trajectory behind the accepted model, that precision is \(1.25\%\).

The two AG02 executions, one month apart, separate two things that are easy to conflate. At the level of individual parameters, agreement is poor and unevenly so: the median relative difference over the 39 parameters and four sampling intervals is \(4.1\%\), but the ninetieth percentile is \(72.8\%\) and the worst case is \(ZZR_3\) at \(689\%\). At the level of what the model predicts, agreement is good: a model identified on one execution and evaluated on the other loses only \(0.37\) to \(0.79\%\) of RMSE relative to the model fitted on that execution itself. The discrepancy is not a contradiction. Parameter reproducibility tracks identifiability, and the parameters that move are the ones the excitation barely constrains: \(ZZR_2\), whose OLS estimate is \(98\) times its standard error, reproduces to \(0.3\)--\(0.9\%\), whereas \(ZZR_3\), at \(1.7\) standard errors, does not reproduce at all. A base-parameter value should therefore be quoted with that ratio, and the model should be compared through its predictions.

Feasibility reproduces more sharply than either. Repeating the all-pose audit of Sec.~VI-E, on its \(221{,}875\)-configuration set, on the OLS estimates of the two executions returns the same verdict at every sampling interval: not PD at \(10\), \(20\), and \(40~\mathrm{ms}\), PD at \(80~\mathrm{ms}\), with the positive margins at \(80~\mathrm{ms}\) agreeing to \(2\%\). The audit verdict is thus a stable property of the trajectory and the interval, not an accident of one recording, which is what allows it to serve as an acceptance criterion.

The selection of \(40~\mathrm{ms}\) rests on the excitation: the information about the inertial parameters lives on the acceleration ramps of the 4-1-4 profile, whose effective duration across the set has a median of \(0.200~\mathrm{s}\) and a minimum of \(0.139~\mathrm{s}\), so that at \(10\), \(20\), and \(40~\mathrm{ms}\) every ramp carries at least three samples while at \(80~\mathrm{ms}\) between \(59\) and \(68\%\) of them carry fewer. The sampling-interval ladder (Fig.~\ref{fig:ts_ladder}, Appendix~\ref{app:validation}-B) separates two effects: the in-sample residual falls by \(24\) to \(55\%\) from \(10\) to \(80~\mathrm{ms}\) as the anti-alias corner descends, whereas held-out RMSE improves by less than \(1\%\) from \(40\) to \(80~\mathrm{ms}\) on all six trajectories, less than the run-to-run variation, which supports the \(40~\mathrm{ms}\) interval selected from the excitation ramps.

Appendix~\ref{app:impl}-I itemizes the cost: producing one accepted model is a few hours of offline computation on a laptop, and the \(307~\mathrm{h}\) of Sec.~IV-C is the cost of the \emph{study}, of running every trajectory at every interval, not of an identification. How often it has to be repeated is a separate question: on the evidence of the two AG02 executions, which agree to \(1.25\%\) on held-out prediction and return the same audit verdict at every interval, identification in an unchanged configuration looks like a one-time calibration on a timescale of at least a month rather than a periodic maintenance task; what would invalidate it is listed in Sec.~VII-E.

One input is not measured at all: the torque record is a commanded current times the constants of Sec.~III-A, catalogue values times an empirical \(1.3\) that no torque measurement calibrates. A uniform error scales the least-squares solution by the same factor and nothing else, so the inertias and first moments of Appendix~A are no more accurate in absolute terms than a datasheet coefficient, however well they reproduce. Independent per-joint errors of \(\pm10\%\), the realistic case since \(K_t\) varies from unit to unit, move the median parameter by \(6.0\%\) over \(200\) draws, essentially the input's own spread, held-out MNE by at most \(1.26\%\), and the feasibility verdict of the OLS estimate in \(11\) draws (Appendix~\ref{app:validation}-C). Such an error can change the estimate and its route through Algorithm~\ref{alg:pipeline}, but not the acceptance criterion: whichever route is taken, a model is accepted only if it passes the same audit.

The same scale fixes what the modeling choices of Sec.~III-C are worth: on the seven trajectories of Table~\ref{tab:reduction}, removing the Coulomb coefficients costs up to \(0.9\%\) of held-out RMSE and removing the products of inertia gains \(0.5\) to \(5.9\%\), the latter exceeding the \(1.25\%\) run-to-run variation on three of the seven; the first is a robustness choice with a descriptive cost, the second an improvement observed on every trajectory tested.

\subsection{Implications and Limitations}

Useful dynamic modeling for a low-cost arm does not require a maximally detailed model or an optimized excitation; it requires a staged framework matched to the platform, in which the PD audit decides acceptance over a pose domain wider than the data.

One entry of the accepted model in Table~\ref{tab:appc} is negative, \(ZZ_7=-3.7\times10^{-5}~\mathrm{kg\cdot m^2}\). \(ZZ_7\) is not a grouped quantity but the axial inertia component of the last link itself, and a negative value is not physically consistent; it is an estimate two orders of magnitude below the rotor inertia \(IA_7=4.85\times10^{-3}\) with which it forms the \((7,7)\) entry of \(\bM(\btheta)\), which is \(4.81\times10^{-3}\) at every configuration, and of the size that does not reproduce between executions (Sec.~VII-D). The assembled matrix passes the finite audit; the link does not satisfy link-level consistency. That is exactly the gap named below between positive definiteness of \(\bM(\btheta)\) and consistency of the links, and the sign of a base entry, grouped or not, is not what the audit tests.

Throughout this paper, physical feasibility means \eqref{eq:pd_check_lambda}, positive definiteness of the inertia matrix over the pose domain, which is what the two uses in Sec.~VII-C require. It is weaker than full physical consistency in the sense of Wensing \textit{et al.}~\cite{WensingPhysicalConsistency}, positive definiteness of each link's \(4\times4\) pseudo-inertia, which forces the triangle inequality on the principal moments and is what link-level simulators such as MuJoCo~\cite{TodorovErezTassa2012MuJoCo} enforce; reinforcement learning with domain randomization over link parameters therefore needs a fully consistent link-level set, which the model identified here is not.

The gap is narrower on this arm than it might appear. Base-parameter identification is a non-invertible projection of the link parameters~\cite{KhalilGautierIDIM,MayedaYoshidaOsukaBaseParam}, but because every joint offset of Table~\ref{tab:dh} vanishes except \(RL_3\) and \(RL_5\), the grouping of Appendix~A mixes only the diagonal of the inertia tensors and the axial first moments and leaves the rest ungrouped. A link-level reconstruction therefore has to supply only that mixed part, by searching the non-identifiable subspace for a standard-parameter set whose pseudo-inertias are all positive definite or by seeding it from a CAD or URDF prior and refining under that constraint~\cite{JanotWensing2021CEP,GazEtAl2019Panda,XuEtAl2020KUKAiiwa}, the two positions on such a prior being those of Sec.~II-C. Not every base vector admits such a realization: the negative ungrouped \(ZZ_7\) of the accepted model rules one out as it stands, so a reconstruction would have to move that entry and re-audit. CRANE-X7 has a published URDF, so that route is open; it is a different problem from the one solved here, and the present result gives it a target, a measured \(\bM(\btheta)\) whose predictions are characterized to the precision of Sec.~VII-D.

\subsection{What the evidence does not cover}

Everything reported here comes from one unit in one configuration: a single CRANE-X7 with the cross-structure hand of \cite{Yamane2024CrossStructureHand}, no payload, and gravity along \(+z_0\) of frame~0 of Table~\ref{tab:dh}; unit-to-unit variation was not measured. Three changes would invalidate the model rather than degrade it: a payload changes the inertial parameters of the last link and the first moments distal to it; a change of mounting orientation changes the gravity rows of the regressor and hence what slow motion can observe, the Joint-1 gravity row being identically zero under the present mounting; and actuator replacement changes both the torque constant and the rotor inertia of that axis. Temperature was neither logged nor controlled, so the current-to-torque relation and the viscous coefficients should be expected to drift over a long session. Friction is viscous only, no backlash dynamics are modelled, feasibility is tested on a finite pose sample, and the whole procedure is offline.

%% file: tables/tab_pd_necessity.tex
\begin{table}[t]
\centering
\footnotesize
\setlength{\tabcolsep}{2.4pt}
\renewcommand{\arraystretch}{1.1}
\caption{What an infeasible model does, evaluated over the audit domain. ``Non-PD'' is the fraction of the domain on which \(\bM(\btheta)\) loses positive definiteness; \(\tau_{\mathrm{grow}}\) is the time constant of the resulting unstable closed-loop root under the acceleration-resolved gains of Sec.~VII-C; \(1/\lambda\) is the observer's acceleration-domain gain. The last three columns are taken at each model's worst pose, the configuration with the smallest \(\lambda_{\min}(\hat{\bM})\): the fraction of \(2{,}000\) random unit torque directions whose acceleration under \(\hat{\bM}^{-1}\) lies more than \(90^\circ\) from the accepted model's, the median over \(200\) random unit accelerations \(\ddot{\btheta}\) of \(\lVert(\hat{\bM}-\bM_{\mathrm{acc}})\ddot{\btheta}\rVert/\lVert\bM_{\mathrm{acc}}\ddot{\btheta}\rVert\), and the same ratio along the eigenvector of \(\lambda_{\min}(\hat{\bM})\).}
\label{tab:pdnecessity}
\begin{tabular*}{\columnwidth}{@{\extracolsep{\fill}}l c c c c c c@{}}
\toprule
& & & \(1/\lambda\) & \multicolumn{3}{c}{At the worst pose} \\
\cmidrule(l){5-7}
Model & Non-PD & \(\tau_{\mathrm{grow}}\) & range & \(>\!90^\circ\) & \(\Delta\tau\) & \(\Delta\tau\) worst \\
\midrule
OLS, 40 ms        & 0.73\%  & 3\,ms  & \(\pm10^{6}\) & 43.4\% & 4.6\% & 637\% \\
SDP, \(\epsPD\!=\!10^{-3}\) & 0.04\%  & 7\,ms  & \(\pm10^{6}\) & 65.4\% & 2.3\% & 345\% \\
CLIE, pre-rescue  & 0.06\%  & 6\,ms  & \(\pm10^{5}\) & 62.8\% & 2.6\% & 365\% \\
Submitted model   & 0.025\% & 10\,ms & \(\pm10^{5}\) & 67.5\% & 6.6\% & 244\% \\
\midrule
Accepted model    & \textbf{0} & --- & \([208,1101]\) & \textbf{0} & --- & --- \\
\bottomrule
\end{tabular*}
\vspace{0.6mm}
{\footnotesize\raggedright All four infeasible models reverse the sign of the predicted torque along the offending direction. The domain is \(\Theta\) of Sec.~V-A. ``Non-PD'' is a fraction of the domain, so it is estimated from the \(150{,}000\) uniform draws alone; the audit set of Sec.~VI-E deliberately oversamples the folded-arm region and a fraction taken over it would be inflated by construction, by a factor of three to six on these models. The accepted model stands in for the true inertia matrix, which is not measured; \(\lambda\) is \(\lambda_{\min}(\hat{\bM})\) at a configuration, \(\mu\) an eigenvalue of \(\bM_{\mathrm{acc}}^{-1}\hat{\bM}\). The \(1/\lambda\) range is the range of \(1/\lambda_{\min}(\hat{\bM})\), in \(\mathrm{(kg\cdot m^2)^{-1}}\), and it and the worst pose are taken over a \(171{,}875\)-configuration set (the grid, \(100{,}000\) uniform and \(50{,}000\) focused draws, seed 1), so the range differs slightly from the audit draws of Sec.~VI-E; the directions are Gaussian vectors normalized to unit length (seed 11). The worst pose depends on the draw; its evaluation does not.\par}
\end{table}

%% file: tables/tab_repeatability.tex
\begin{table}[t]
\centering
\footnotesize
\setlength{\tabcolsep}{3pt}
\renewcommand{\arraystretch}{1.05}
\caption{Run-to-run repeatability. Every group is repeated executions of one byte-identical command file on the same unit. Each execution is fitted independently at 40~ms and used to predict the held-out set; the last column is the spread of that RMSE across the executions of the group, largest minus smallest, relative to their mean. \(n\) is samples per axis.}
\label{tab:repeatability}
\begin{tabular*}{\columnwidth}{@{\extracolsep{\fill}}l c l r c@{}}
\toprule
Trajectory & Runs & Separations & \(n\) & RMSE spread \\
\midrule
PC01 & 2 & 19 min                & 4241 & 0.56\% \\
PB01 & 3 & 4 h, then 5 days      & 4007 & 0.45\% \\
AG02 & 2 & 26 days               & 3124 & 1.25\% \\
AP02 & 3 & 42 min, then 11 days  & \ 474 & 1.43\% \\
AP03 & 3 & 34 min, then 11 days  & \ 532 & 3.55\% \\
\bottomrule
\end{tabular*}
\end{table}

%% file: sections/08_conclusion.tex
\section{Conclusion}

This paper presented a reproducible dynamic parameter identification pipeline for CRANE-X7, a low-cost 7-DOF arm with 12-bit encoders, communication delay, and backlash-affected drivetrains: hand-designed single-joint and adjacent-pair excitation with a complete numerical specification (Appendix~\ref{app:traj}), a 39-parameter OpenSYMORO model without CAD data, OLS, a conditional SDP projection, CLIE refinement, an all-pose positive-definiteness audit of the inertia matrix as the acceptance test, and a localized SDP rescue when the audit fails.

Three results follow. First, the reduction from 65 to 39 parameters is justified by held-out prediction: the products of inertia carry \(1.7\%\) of the commanded torque against a \(27\%\) unmodelled residual floor, and retaining them degrades prediction on every trajectory tested. Second, physical feasibility is an acceptance criterion independent of prediction accuracy: an infeasible model within a few per cent in predicted torque produces, in a fixed-configuration analysis, an unstable closed-loop root and misdirects the inertia inversion of the reaction-force observer for up to two thirds of torque directions, and the audit is itself a measurement, its margin moving by a factor of \(1.8\) between random seeds and its verdict becoming seed-dependent on a sparse pose set. Third, reproducibility and sensitivity are quantified: two executions of the accepted trajectory 26 days apart put the run-to-run spread of held-out RMSE at \(1.25\%\), the scale against which every comparison is read, and independent per-axis perturbations of the uncalibrated torque constants within \(\pm10\%\) change held-out mean normalized error by at most \(1.26\%\) over \(200\) trials while changing the PD verdict of the OLS estimate in \(11\) of them.

The resulting model supports inertia-matrix-based evaluation within the tested configuration and pose sample; full link-level physical consistency is a separate reconstruction task (Sec.~VII-E). Future work will investigate refinement based on acceleration prediction alongside torque prediction, with the aim of improving the inverse-inertia behavior the intended applications require, together with richer drivetrain and payload models. Further validation will address optimized excitation, certified feasibility, parameter uncertainty, and transfer to other units and platforms.

%% file: sections/appendix_a_model.tex

\section{Model Specification}\label{app:model}

This appendix specifies the symbolic model completely enough to be regenerated independently of our implementation: with the modified DH parameters of Table~\ref{tab:dh}, the relations below define the base-parameter set that OpenSYMORO produces for CRANE-X7, and hence the regressor \(\bW\) and the direct dynamic model of the CLIE stage.
Each relation states which standard inertial parameters of neighbouring links are observed only through their sum, so an entry of the identified vector is either such a grouped quantity or a link parameter proper, a distinction Sec.~VII-E relies on.

Two offsets of the manufacturer's published URDF~\cite{RTCraneX7Description}, a \(0.105~\mathrm{m}\) rise along the Joint-1 axis and a \(0.019~\mathrm{m}\) offset along \(z_7\) to the gripper base, are absent from Table~\ref{tab:dh}. The first is a rigid translation along the gravity axis and leaves \(\bM(\btheta)\) and the gravity vector unchanged; the second is absorbed into the inertial parameters of link~7, identified in the frame of Table~\ref{tab:dh}, so that \(MZ_7\), \(YY_7\), and the groupings \(XXR_6\), \(ZZR_6\), and \(MYR_6\) below would take other values in the displaced frame, and a comparison with CAD values requires the parallel-axis transfer.

\subsection{Base-parameter grouping}

Write \(MR_j=\sum_{k\ge j}M_k\) for the accumulated distal mass, so that \(MR_7=M_7\).
The grouped (base) inertial parameters are
\begingroup\allowdisplaybreaks
\begin{align*}
ZZR_1 &= ZZ_1 + YY_2 + IA_1,\\
XXR_2 &= XX_2 - YY_2 + YY_3 + RL_3^{2}MR_3 + 2RL_3MZ_3,\\
ZZR_2 &= ZZ_2 + YY_3 + RL_3^{2}MR_3 + 2RL_3MZ_3 + IA_2,\\
MYR_2 &= MY_2 - MZ_3 - RL_3MR_3,\\
XXR_3 &= XX_3 - YY_3 + YY_4, \quad ZZR_3 = ZZ_3 + YY_4,\\
MYR_3 &= MY_3 + MZ_4,\\
XXR_4 &= XX_4 - YY_4 + YY_5 + RL_5^{2}MR_5 + 2RL_5MZ_5,\\
ZZR_4 &= ZZ_4 + YY_5 + RL_5^{2}MR_5 + 2RL_5MZ_5,\\
MYR_4 &= MY_4 - MZ_5 - RL_5MR_5,\\
XXR_5 &= XX_5 - YY_5 + YY_6, \quad ZZR_5 = ZZ_5 + YY_6,\\
MYR_5 &= MY_5 + MZ_6,\\
XXR_6 &= XX_6 - YY_6 + YY_7, \quad ZZR_6 = ZZ_6 + YY_7,\\
MYR_6 &= MY_6 - MZ_7,\\
XXR_7 &= XX_7 - YY_7 .
\end{align*}
\endgroup
The remaining base parameters (\(MX_j\), \(XY_j\), \(XZ_j\), \(YZ_j\), \(ZZ_7\), \(MY_7\), \(IA_3\)--\(IA_7\)) are not regrouped and appear unchanged.
The rotor inertias of Axes~1 and 2 are absorbed into \(ZZR_1\) and \(ZZR_2\), which is why only \(IA_3\)--\(IA_8\) appear separately, and entries such as \(ZZR_6=ZZ_6+YY_7\) are sums or differences across links, so the sign of an individual base entry carries no direct physical meaning (Sec.~VII-E).

\subsection{From 65 to 39 parameters}

The full grouped set contains 65 parameters: 48 inertial terms for the seven arm links, \(FS_1\)--\(FS_7\) and \(FV_1\)--\(FV_7\), and \(IA_8\), \(FS_8\), \(FV_8\) for the hand axis.
The reduced model of Sec.~III-C removes the 18 products of inertia \(XY_j\), \(XZ_j\), \(YZ_j\) for \(j=2,\dots,7\) and the 8 Coulomb friction coefficients \(FS_1\)--\(FS_8\) and retains the rest, which leaves \(65-18-8=39\) parameters, listed with their final values in Table~\ref{tab:appc}.

\subsection{Note on the symbol \(MY_7\)}

In the OpenSYMORO input the \(y\) first moment of link~7 was entered under the symbol \(MZ_7\); its regressor column, \(-\dot v_{x,7}\), is that of \(MY_j\) in \(\tau_j = MX_j\dot v_y - MY_j\dot v_x\), and \(MZ_7\) proper is not identifiable for a revolute joint.
We write it \(MY_7\) throughout; the \(MZ_7\) absorbed by the relation for \(MYR_6\) above is the axial first moment proper, not a parameter of the model, and the numerical values are unaffected.

\subsection{Numerical record}

Table~\ref{tab:appc} is the complete numerical record behind Secs.~VI and VII: all \(160\) trajectory--sampling conditions with regressor condition number, routing test, route, SDP margin, and CLIE iteration count (the cost measure of Sec.~IV-C), and the 39 parameters of the accepted model, the grouped quantities defined above, at the three significant digits at which the model was audited; the earlier version's I/S/C and A/G labels are PA/PB/PC and AP/AG here.

\input{tables/tab_numerical_record}

%% file: tables/tab_numerical_record.tex
\begin{table*}[p]
\centering
\caption{Complete numerical record of the \(160\) identification conditions and the accepted model. First four panels, per trajectory and sampling interval \(T_s\). \(\kappa=\sigma_{\max}(\bW_{\mathrm{all}})/\sigma_{\min}(\bW_{\mathrm{all}})\): regressor condition number. PD: positive definiteness of the OLS estimate at the configurations the trajectory visits, the routing test of Sec.~VI-E, not the acceptance audit. Rt: route, O-C = OLS\(\rightarrow\)CLIE, O-S-C = OLS\(\rightarrow\)SDP\(\rightarrow\)CLIE. \(\varepsilon_{\mathrm{PD}}\): SDP margin, a dash meaning no SDP step. \(n_{\mathrm{it}}\): CLIE iterations reported by \texttt{lsqnonlin} at exit, a dagger marking the nine stopped by \texttt{MaxIterations}\,\(=400\) after their 401st iteration. Last two panels: the 39 parameters of the accepted model, the PD-rescued AG02 solution at \(40~\mathrm{ms}\), to the three significant digits at which it was audited.}
\label{tab:appc}
\scriptsize
\renewcommand{\arraystretch}{0.94}
\begin{minipage}[t]{0.318\textwidth}
\centering
\scriptsize
{\normalsize\bfseries PA01--PA11 (Posture A)}\\[0.2mm]
\setlength{\tabcolsep}{2.4pt}
\begin{tabular}{lrrrrcr}
\toprule
Tr & $T_s$ [ms] & $\kappa$ & PD & Rt & $\varepsilon_{\mathrm{PD}}$ & $n_{\mathrm{it}}$ \\
\midrule
PA01 & 10 & 37.8 & N & O-S-C & 0.005 & 5 \\
 & 20 & 37.5 & N & O-S-C & 0.001 & 5 \\
 & 40 & 38.0 & Y & O-C & -- & 5 \\
 & 80 & 41.4 & Y & O-C & -- & 4 \\
PA02 & 10 & 34.0 & N & O-S-C & 0.005 & 30 \\
 & 20 & 33.2 & N & O-S-C & 0.005 & 47 \\
 & 40 & 33.2 & Y & O-C & -- & 37 \\
 & 80 & 33.7 & Y & O-C & -- & 23 \\
PA03 & 10 & 36.3 & N & O-S-C & 0.005 & 6 \\
 & 20 & 39.3 & Y & O-C & -- & 6 \\
 & 40 & 41.2 & Y & O-C & -- & 5 \\
 & 80 & 42.6 & Y & O-C & -- & 6 \\
PA04 & 10 & 34.6 & Y & O-C & -- & 6 \\
 & 20 & 37.9 & Y & O-C & -- & 5 \\
 & 40 & 39.0 & Y & O-C & -- & 5 \\
 & 80 & 41.4 & Y & O-C & -- & 4 \\
PA05 & 10 & 33.8 & N & O-S-C & 0.001 & 31 \\
 & 20 & 31.7 & N & O-S-C & 0.001 & 179 \\
 & 40 & 31.6 & N & O-S-C & 0.001 & 205 \\
 & 80 & 35.8 & Y & O-C & -- & 40 \\
PA06 & 10 & 42.0 & N & O-S-C & 0.001 & 6 \\
 & 20 & 50.1 & Y & O-C & -- & 6 \\
 & 40 & 52.8 & Y & O-C & -- & 6 \\
 & 80 & 55.1 & Y & O-C & -- & 5 \\
PA07 & 10 & 107.0 & N & O-S-C & 0.005 & 6 \\
 & 20 & 133.3 & N & O-S-C & 0.001 & 5 \\
 & 40 & 144.8 & N & O-S-C & 0.010 & 7 \\
 & 80 & 154.8 & N & O-S-C & 0.010 & 7 \\
PA08 & 10 & 39.4 & N & O-S-C & 0.001 & 6 \\
 & 20 & 47.9 & Y & O-C & -- & 6 \\
 & 40 & 51.3 & Y & O-C & -- & 5 \\
 & 80 & 54.4 & Y & O-C & -- & 4 \\
PA09 & 10 & 84.1 & N & O-S-C & 0.001 & 23 \\
 & 20 & 124.6 & N & O-S-C & 0.001 & 68 \\
 & 40 & 143.1 & N & O-S-C & 0.005 & 50 \\
 & 80 & 155.5 & N & O-S-C & 0.005 & 24 \\
PA10 & 10 & 37.3 & N & O-S-C & 0.005 & 60 \\
 & 20 & 37.1 & N & O-S-C & 0.001 & 30 \\
 & 40 & 39.8 & N & O-S-C & 0.001 & 51 \\
 & 80 & 44.2 & N & O-S-C & 0.001 & 156 \\
PA11 & 10 & 85.8 & N & O-S-C & 0.010 & 40 \\
 & 20 & 120.1 & N & O-S-C & 0.010 & 18 \\
 & 40 & 148.1 & N & O-S-C & 0.005 & 31 \\
 & 80 & 173.3 & N & O-S-C & 0.005 & 43 \\
\bottomrule
\end{tabular}
\end{minipage}
\hfill
\begin{minipage}[t]{0.318\textwidth}
\centering
\scriptsize
{\normalsize\bfseries PB01--PB11 (Posture B)}\\[0.2mm]
\setlength{\tabcolsep}{2.4pt}
\begin{tabular}{lrrrrcr}
\toprule
Tr & $T_s$ [ms] & $\kappa$ & PD & Rt & $\varepsilon_{\mathrm{PD}}$ & $n_{\mathrm{it}}$ \\
\midrule
PB01 & 10 & 45.8 & N & O-S-C & 0.005 & 15 \\
 & 20 & 45.7 & N & O-S-C & 0.005 & 104 \\
 & 40 & 45.9 & Y & O-C & -- & 110 \\
 & 80 & 49.4 & Y & O-C & -- & 21 \\
PB02 & 10 & 36.8 & N & O-S-C & 0.001 & 32 \\
 & 20 & 36.6 & N & O-S-C & 0.002 & 88 \\
 & 40 & 36.7 & N & O-S-C & 0.002 & 259 \\
 & 80 & 37.3 & Y & O-C & -- & 154 \\
PB03 & 10 & 39.7 & N & O-S-C & 0.001 & 6 \\
 & 20 & 44.3 & N & O-S-C & 0.001 & 6 \\
 & 40 & 45.9 & Y & O-C & -- & 5 \\
 & 80 & 47.6 & Y & O-C & -- & 6 \\
PB04 & 10 & 41.3 & Y & O-C & -- & 6 \\
 & 20 & 47.9 & Y & O-C & -- & 5 \\
 & 40 & 49.9 & Y & O-C & -- & 4 \\
 & 80 & 53.2 & Y & O-C & -- & 4 \\
PB05 & 10 & 40.7 & N & O-S-C & 0.005 & 32 \\
 & 20 & 40.4 & N & O-S-C & 0.005 & 401$^{\dagger}$ \\
 & 40 & 40.5 & N & O-S-C & 0.005 & 401$^{\dagger}$ \\
 & 80 & 44.6 & N & O-S-C & 0.005 & 401$^{\dagger}$ \\
PB06 & 10 & 49.2 & N & O-S-C & 0.001 & 6 \\
 & 20 & 56.9 & Y & O-C & -- & 6 \\
 & 40 & 59.9 & Y & O-C & -- & 5 \\
 & 80 & 62.3 & Y & O-C & -- & 5 \\
PB07 & 10 & 121.1 & N & O-S-C & 0.005 & 158 \\
 & 20 & 157.5 & N & O-S-C & 0.005 & 33 \\
 & 40 & 172.2 & N & O-S-C & 0.005 & 31 \\
 & 80 & 182.5 & N & O-S-C & 0.005 & 137 \\
PB08 & 10 & 47.7 & Y & O-C & -- & 6 \\
 & 20 & 58.9 & Y & O-C & -- & 9 \\
 & 40 & 62.4 & Y & O-C & -- & 4 \\
 & 80 & 65.9 & Y & O-C & -- & 3 \\
PB09 & 10 & 102.8 & N & O-S-C & 0.001 & 50 \\
 & 20 & 152.3 & Y & O-C & -- & 22 \\
 & 40 & 169.9 & Y & O-C & -- & 4 \\
 & 80 & 182.8 & N & O-S-C & 0.010 & 30 \\
PB10 & 10 & 46.4 & N & O-S-C & 0.005 & 13 \\
 & 20 & 46.3 & N & O-S-C & 0.005 & 4 \\
 & 40 & 48.1 & Y & O-C & -- & 3 \\
 & 80 & 53.4 & Y & O-C & -- & 3 \\
PB11 & 10 & 80.6 & N & O-S-C & 0.001 & 69 \\
 & 20 & 113.6 & N & O-S-C & 0.005 & 116 \\
 & 40 & 138.6 & N & O-S-C & 0.005 & 14 \\
 & 80 & 159.7 & N & O-S-C & 0.005 & 401$^{\dagger}$ \\
\bottomrule
\end{tabular}
\end{minipage}
\hfill
\begin{minipage}[t]{0.318\textwidth}
\centering
\scriptsize
{\normalsize\bfseries PC01--PC11 (Posture C)}\\[0.2mm]
\setlength{\tabcolsep}{2.4pt}
\begin{tabular}{lrrrrcr}
\toprule
Tr & $T_s$ [ms] & $\kappa$ & PD & Rt & $\varepsilon_{\mathrm{PD}}$ & $n_{\mathrm{it}}$ \\
\midrule
PC01 & 10 & 39.1 & N & O-S-C & 0.001 & 244 \\
 & 20 & 39.0 & N & O-S-C & 0.001 & 132 \\
 & 40 & 39.1 & N & O-S-C & 0.001 & 109 \\
 & 80 & 39.7 & Y & O-C & -- & 79 \\
PC02 & 10 & 45.5 & N & O-S-C & 0.010 & 10 \\
 & 20 & 45.4 & N & O-S-C & 0.005 & 86 \\
 & 40 & 45.6 & Y & O-C & -- & 71 \\
 & 80 & 49.2 & Y & O-C & -- & 63 \\
PC03 & 10 & 38.3 & N & O-S-C & 0.001 & 9 \\
 & 20 & 37.9 & Y & O-C & -- & 93 \\
 & 40 & 38.0 & Y & O-C & -- & 54 \\
 & 80 & 38.7 & Y & O-C & -- & 30 \\
PC04 & 10 & 35.2 & Y & O-C & -- & 6 \\
 & 20 & 35.1 & Y & O-C & -- & 5 \\
 & 40 & 35.2 & Y & O-C & -- & 4 \\
 & 80 & 35.8 & Y & O-C & -- & 4 \\
PC05 & 10 & 34.7 & N & O-S-C & 0.001 & 401$^{\dagger}$ \\
 & 20 & 34.6 & N & O-S-C & 0.001 & 24 \\
 & 40 & 34.7 & N & O-S-C & 0.001 & 112 \\
 & 80 & 35.2 & N & O-S-C & 0.001 & 47 \\
PC06 & 10 & 42.9 & N & O-S-C & 0.005 & 7 \\
 & 20 & 42.8 & Y & O-C & -- & 6 \\
 & 40 & 44.7 & Y & O-C & -- & 6 \\
 & 80 & 48.0 & Y & O-C & -- & 5 \\
PC07 & 10 & 75.8 & N & O-S-C & 0.001 & 5 \\
 & 20 & 107.1 & N & O-S-C & 0.005 & 5 \\
 & 40 & 125.1 & N & O-S-C & 0.001 & 7 \\
 & 80 & 135.5 & N & O-S-C & 0.010 & 6 \\
PC08 & 10 & 41.4 & Y & O-C & -- & 5 \\
 & 20 & 41.4 & Y & O-C & -- & 4 \\
 & 40 & 42.7 & Y & O-C & -- & 4 \\
 & 80 & 45.9 & Y & O-C & -- & 3 \\
PC09 & 10 & 72.0 & N & O-S-C & 0.001 & 175 \\
 & 20 & 103.1 & N & O-S-C & 0.001 & 64 \\
 & 40 & 123.7 & N & O-S-C & 0.001 & 48 \\
 & 80 & 134.7 & N & O-S-C & 0.001 & 55 \\
PC10 & 10 & 40.1 & N & O-S-C & 0.001 & 295 \\
 & 20 & 40.0 & N & O-S-C & 0.001 & 297 \\
 & 40 & 40.1 & N & O-S-C & 0.001 & 56 \\
 & 80 & 42.8 & N & O-S-C & 0.001 & 48 \\
PC11 & 10 & 70.1 & N & O-S-C & 0.001 & 246 \\
 & 20 & 96.9 & N & O-S-C & 0.001 & 15 \\
 & 40 & 120.1 & N & O-S-C & 0.001 & 20 \\
 & 80 & 130.8 & N & O-S-C & 0.001 & 180 \\
\bottomrule
\end{tabular}
\end{minipage}
\\[1.6mm]
\begin{minipage}[t]{0.318\textwidth}
\centering
\scriptsize
{\normalsize\bfseries AP01--AP05, AG01--AG02}\\[0.2mm]
\setlength{\tabcolsep}{2.4pt}
\begin{tabular}{lrrrrcr}
\toprule
Tr & $T_s$ [ms] & $\kappa$ & PD & Rt & $\varepsilon_{\mathrm{PD}}$ & $n_{\mathrm{it}}$ \\
\midrule
AP01 & 10 & 47.1 & N & O-S-C & 0.005 & 22 \\
 & 20 & 46.1 & Y & O-C & -- & 21 \\
 & 40 & 46.2 & Y & O-C & -- & 26 \\
 & 80 & 47.2 & Y & O-C & -- & 40 \\
AP02 & 10 & 49.2 & N & O-S-C & 0.005 & 401$^{\dagger}$ \\
 & 20 & 48.8 & N & O-S-C & 0.005 & 401$^{\dagger}$ \\
 & 40 & 48.9 & N & O-S-C & 0.005 & 401$^{\dagger}$ \\
 & 80 & 50.0 & N & O-S-C & 0.005 & 401$^{\dagger}$ \\
AP03 & 10 & 42.7 & Y & O-C & -- & 10 \\
 & 20 & 42.6 & Y & O-C & -- & 8 \\
 & 40 & 42.7 & Y & O-C & -- & 7 \\
 & 80 & 43.5 & Y & O-C & -- & 6 \\
AP04 & 10 & 88.3 & N & O-S-C & 0.001 & 48 \\
 & 20 & 88.2 & N & O-S-C & 0.001 & 16 \\
 & 40 & 102.0 & N & O-S-C & 0.001 & 14 \\
 & 80 & 113.3 & N & O-S-C & 0.010 & 17 \\
AP05 & 10 & 49.3 & N & O-S-C & 0.001 & 5 \\
 & 20 & 49.0 & N & O-S-C & 0.001 & 4 \\
 & 40 & 49.2 & Y & O-C & -- & 4 \\
 & 80 & 50.0 & Y & O-C & -- & 3 \\
AG01 & 10 & 38.5 & N & O-S-C & 0.001 & 14 \\
 & 20 & 38.0 & N & O-S-C & 0.001 & 128 \\
 & 40 & 38.1 & Y & O-C & -- & 326 \\
 & 80 & 38.8 & Y & O-C & -- & 24 \\
AG02 & 10 & 33.4 & N & O-S-C & 0.001 & 118 \\
 & 20 & 33.2 & N & O-S-C & 0.001 & 74 \\
 & 40 & 33.3 & N & O-S-C & 0.001 & 86 \\
 & 80 & 34.1 & Y & O-C & -- & 41 \\
\bottomrule
\end{tabular}
\end{minipage}
\hfill
\begin{minipage}[t]{0.318\textwidth}
\centering
\footnotesize
\renewcommand{\arraystretch}{1.00}
{\small\bfseries Final parameters (1/2)}\\[0.9mm]
\setlength{\tabcolsep}{3.4pt}
\begin{tabular}{llr}
\toprule
Param & Unit & Value \\
\midrule
ZZR1 & \(\mathrm{kg\cdot m^2}\) & \(1.23\times10^{-2}\) \\
XXR2 & \(\mathrm{kg\cdot m^2}\) & \(5.93\times10^{-2}\) \\
ZZR2 & \(\mathrm{kg\cdot m^2}\) & \(9.62\times10^{-2}\) \\
MX2 & \(\mathrm{kg\cdot m}\) & \(9.48\times10^{-3}\) \\
MYR2 & \(\mathrm{kg\cdot m}\) & \(-2.38\times10^{-1}\) \\
XXR3 & \(\mathrm{kg\cdot m^2}\) & \(2.54\times10^{-3}\) \\
ZZR3 & \(\mathrm{kg\cdot m^2}\) & \(1.60\times10^{-3}\) \\
MX3 & \(\mathrm{kg\cdot m}\) & \(7.22\times10^{-3}\) \\
MYR3 & \(\mathrm{kg\cdot m}\) & \(8.72\times10^{-4}\) \\
XXR4 & \(\mathrm{kg\cdot m^2}\) & \(1.04\times10^{-2}\) \\
ZZR4 & \(\mathrm{kg\cdot m^2}\) & \(2.35\times10^{-2}\) \\
MX4 & \(\mathrm{kg\cdot m}\) & \(2.63\times10^{-3}\) \\
MYR4 & \(\mathrm{kg\cdot m}\) & \(-1.43\times10^{-1}\) \\
IA3 & \(\mathrm{kg\cdot m^2}\) & \(9.38\times10^{-3}\) \\
IA4 & \(\mathrm{kg\cdot m^2}\) & \(1.79\times10^{-2}\) \\
IA5 & \(\mathrm{kg\cdot m^2}\) & \(4.73\times10^{-3}\) \\
IA6 & \(\mathrm{kg\cdot m^2}\) & \(4.81\times10^{-3}\) \\
IA7 & \(\mathrm{kg\cdot m^2}\) & \(4.85\times10^{-3}\) \\
IA8 & \(\mathrm{kg\cdot m^2}\) & \(5.33\times10^{-3}\) \\
\bottomrule
\end{tabular}
\end{minipage}
\hfill
\begin{minipage}[t]{0.318\textwidth}
\centering
\footnotesize
\renewcommand{\arraystretch}{1.00}
{\small\bfseries Final parameters (2/2)}\\[0.9mm]
\setlength{\tabcolsep}{3.4pt}
\begin{tabular}{llr}
\toprule
Param & Unit & Value \\
\midrule
XXR5 & \(\mathrm{kg\cdot m^2}\) & \(1.69\times10^{-2}\) \\
ZZR5 & \(\mathrm{kg\cdot m^2}\) & \(4.11\times10^{-4}\) \\
MX5 & \(\mathrm{kg\cdot m}\) & \(-1.20\times10^{-3}\) \\
MYR5 & \(\mathrm{kg\cdot m}\) & \(1.12\times10^{-3}\) \\
XXR6 & \(\mathrm{kg\cdot m^2}\) & \(3.31\times10^{-4}\) \\
ZZR6 & \(\mathrm{kg\cdot m^2}\) & \(7.36\times10^{-4}\) \\
MX6 & \(\mathrm{kg\cdot m}\) & \(1.18\times10^{-3}\) \\
MYR6 & \(\mathrm{kg\cdot m}\) & \(-5.24\times10^{-3}\) \\
XXR7 & \(\mathrm{kg\cdot m^2}\) & \(2.65\times10^{-4}\) \\
ZZ7 & \(\mathrm{kg\cdot m^2}\) & \(-3.68\times10^{-5}\) \\
MX7 & \(\mathrm{kg\cdot m}\) & \(-1.09\times10^{-4}\) \\
MY7 & \(\mathrm{kg\cdot m}\) & \(-1.05\times10^{-4}\) \\
FV1 & \(\mathrm{N\cdot m\cdot s/rad}\) & \(5.40\times10^{-2}\) \\
FV2 & \(\mathrm{N\cdot m\cdot s/rad}\) & \(3.08\times10^{-1}\) \\
FV3 & \(\mathrm{N\cdot m\cdot s/rad}\) & \(5.66\times10^{-2}\) \\
FV4 & \(\mathrm{N\cdot m\cdot s/rad}\) & \(2.35\times10^{-1}\) \\
FV5 & \(\mathrm{N\cdot m\cdot s/rad}\) & \(2.85\times10^{-2}\) \\
FV6 & \(\mathrm{N\cdot m\cdot s/rad}\) & \(4.56\times10^{-2}\) \\
FV7 & \(\mathrm{N\cdot m\cdot s/rad}\) & \(3.31\times10^{-2}\) \\
FV8 & \(\mathrm{N\cdot m\cdot s/rad}\) & \(3.12\times10^{-2}\) \\
\bottomrule
\end{tabular}
\end{minipage}
\end{table*}

%% file: sections/appendices.tex

\section{Excitation Trajectory Specification}\label{app:traj}

This section describes the excitation design behind all 40 identification trajectories.
Every trajectory is a concatenation of two kinds of block, and the tables below give the blocks, their numerical content, and the block sequence of each trajectory: this section describes the excitation primitives and the block composition.
The accompanying trajectory files are the authoritative numerical specification of every commanded waypoint and speed, including the posture transitions between primitives and the terminal rows, together with the per-axis velocity ceilings and ramp durations.

\subsection{Why the trajectories are hand-designed}

The excitation motions are joint-space waypoint sequences rather than numerically optimized waveforms.
Three properties of the platform drive that choice.
First, the actuators' continuous-torque margin is small and unevenly distributed (Axis~2 carries the whole distal arm, Axis~7 only the hand), so a single global waveform must be slowed until the most loaded axis is safe; with blocks, the speed factor is assigned per block (Table~\ref{tab:pairblock}).
Second, a block can be repeated at another speed factor without changing a waypoint, which produces the speed ladder of Sec.~VI.
Third, a waypoint list is reproducible on other hardware as it stands; an optimized waveform is reproducible once its coefficients are published, but its derivation is not without the optimizer, its constraints, and its seed.

The reason such restricted motions are sufficient is classical.
Mayeda \textit{et al.}~\cite{Mayeda1984NewIdentification} observed that when the link inertia tensors carry no products of inertia and consecutive joint axes are mutually perpendicular or parallel, the identifiable parameters can be excited by motions in which at most two joints move at a time.
Two facts underlie this.
In the Lagrangian form (3) the dependence on joint velocity is exactly quadratic, so every entry of the regressor is a configuration-dependent function multiplied by one acceleration \(\ddot\theta_k\), by one velocity product \(\dot\theta_k\dot\theta_l\), or by nothing at all; no regressor entry requires three joints to move simultaneously.
The configuration dependence, in turn, requires the remaining joints only to be \emph{placed} at suitable angles, not to move.
Restricting the moving set to at most two joints therefore removes no regressor monomial; it only fixes the configuration at which each monomial is sampled, and the geometric condition, which CRANE-X7 satisfies with \(\alpha_1=0\) and \(\alpha_2,\dots,\alpha_7=\pm\pi/2\) in Table~1, is what keeps the resulting samples independent.

This is verified directly on the present regressor rather than assumed.
The AP family (Table~\ref{tab:trajcomp}) contains adjacent-pair primitives only.
AP03 is \(\Pi_\beta\) used standalone, and counting every segment of its command file shows that it never moves more than two arm joints at any instant; Appendix~\ref{app:traj}-E identifies the three segments in the whole set where a third joint moves and states how often each occurs.
The 39-column regressor built from a single AP trajectory has full column rank~39, with condition numbers of 46.2, 48.9, and 42.7 for AP01, AP02, and AP03 at the 40~ms setting, comparable to the 33.3 obtained from AG02, which contains the single-joint blocks as well.
AP03 is thus the cleanest case: two joints at a time throughout, rank 39 on its own, and \(\kappa=42.7\).
Adjacent-pair excitation alone is therefore sufficient for the reduced model on this arm, the single-joint blocks improve conditioning rather than supply missing rank, and Sec.~VII-B measures how far that excitation gets on held-out prediction.

\subsection{Command format and interpolation}

Each trajectory is stored as a command file whose header fixes the generator constants and whose body lists absolute joint-space waypoints.
The header is identical in every file: the per-axis velocity ceiling is \(274.800^\circ/\mathrm{s}\) for Axes~1 and 3--8 and \(175.872^\circ/\mathrm{s}\) for Axis~2, and the nominal acceleration time is \(0.40\), \(0.20\), \(0.40\), \(0.40\), \(0.20\), \(0.20\), \(0.20\), \(0.20~\mathrm{s}\) for Axes~1--8, respectively.
The dwell appended to each segment is zero throughout, except after the final row, which is followed by \(1~\mathrm{s}\).
Each waypoint line carries the eight target angles in degrees and a speed factor in percent, and the peak velocity of that segment is that percentage of the per-axis ceiling above, so a factor of \(0.90\) means \(247.3^\circ/\mathrm{s}\) on Axes~1 and 3--8 and \(158.3^\circ/\mathrm{s}\) on Axis~2.
Every file assumes the initial posture \(\bar{\bq}_{\mathrm A}\), which is not commanded, and ends at \(\bar{\bq}_0\); the final row repeats \(\bar{\bq}_0\) with a speed field of \(20\%\) and, because the preceding row already reaches \(\bar{\bq}_0\), commands no displacement, and the \(1~\mathrm{s}\) dwell follows it.

Segments are interpolated by a synchronized point-to-point 4-1-4 profile, that is, a cubic-polynomial velocity ramp, a constant-velocity phase, and the mirrored ramp.
On the ramp of duration \(T_a\) with peak velocity \(V\), and with \(\sigma=t/T_a\) the normalized ramp time, the reference velocity and acceleration of the axis are
\begin{equation}
\dot\theta_{\mathrm{ref}} = V\left(3\sigma^{2}-2\sigma^{3}\right),\quad
\ddot\theta_{\mathrm{ref}} = \frac{6V}{T_a}\,\sigma(1-\sigma),
\label{eq:414}
\end{equation}
so the position is quartic in \(t\) and the acceleration is a parabola that vanishes at both ends of the ramp.
The reference acceleration is therefore continuous across every segment boundary, which matters on this platform because a step in \(\ddot\theta_{\mathrm{ref}}\) would appear directly in the feedforward term of (2) and be tracked as a torque discontinuity.
A ramp covers \(\tfrac12 V T_a\), so a segment of displacement \(\Delta\theta\) has a constant-velocity phase whenever \(|\Delta\theta| \ge |V T_a|\); shorter segments keep the profile shape and shrink the ramp to \(T_a'=\sqrt{|\Delta\theta|\,T_a/|V|}\), with peak velocity \(V'=|\Delta\theta|/T_a'\) so that \(V/T_a\), and hence the peak acceleration, is preserved, and no constant-velocity phase.
Each axis is first planned independently from its own speed factor and acceleration time, then all axes are replanned to the longest of the resulting segment times, so that every axis of a segment starts and finishes together with \(V=\Delta\theta/T_c\), \(T_c\) being the segment time less one ramp; each axis keeps its ramp time unless the segment is shorter than two ramps, when half the segment becomes the ramp.
The arm comes to rest at every waypoint; no path blending is used.

The nominal postures are written in the eight-axis command space, the first seven components being the arm joints and the eighth the hand axis:
\[
\begin{aligned}
\bar{\bq}_{0} &= [0,\;0,\;0,\;0,\;0,\;0,\;0,\;0]^\circ,\\
\bar{\bq}_{\mathrm A} &= [0,\;0,\;0,\;-90,\;0,\;0,\;0,\;0]^\circ,\\
\bar{\bq}_{\mathrm B} &= [0,\;0,\;0,\;-150,\;0,\;0,\;0,\;0]^\circ,\\
\bar{\bq}_{\mathrm C} &= [0,\;60,\;0,\;-150,\;0,\;0,\;0,\;0]^\circ ,
\end{aligned}
\]
and \(\bw = -89^\circ\be_6\) denotes the wrist offset used to reach a more severe Axis-6 configuration, \(\be_j\) being the unit vector of joint \(j\).

\subsection{The two primitives}

The \emph{single-joint} primitive is two-sided about a home posture \(\bh\),
\[
\mathcal S_j(a_j;\bh):\quad
\bh \rightarrow \bh-a_j\be_j \rightarrow \bh+a_j\be_j \rightarrow \bh ,
\]
so joint \(j\) visits the three angles \(-a_j\), \(+a_j\), and its home value while all other joints are held.

The \emph{adjacent-pair} primitive is one-sided and in-phase,
\[
\mathcal P_{j,j+1}(s;\bh):\quad
\bh \rightarrow \bh + s\,\delta\,(\be_j+\be_{j+1}) \rightarrow \bh ,
\; s\in\{-1,+1\},
\]
with a common amplitude \(\delta=60^\circ\) on both joints.
Both joints of a pair therefore always move in the same direction and by the same amount; no opposite-phase pair is used anywhere in the set, and no pair excursion is two-sided.
The sign \(s\) is a property of the pattern and is listed in Table~\ref{tab:pairblock}.

\input{tables/tab_single_block}

\subsection{The single-joint block}

The single-joint block \(\mathcal S(\bar{\bq})\) is the same ordered list of ten primitives in every trajectory that contains one, and is listed in Table~\ref{tab:sjblock}.
Eight of the ten sweep one axis each; the remaining two repeat the Axis-5 and Axis-6 sweeps at the shifted partner angle, which is the only place in the identification set where the wrist reaches \(\theta_6=-89^\circ\) with \(\theta_5\) moving, or \(\theta_5=-90^\circ\) with \(\theta_6\) moving.
The block is executed at a single speed factor throughout.
Two departures from the table occur in the files: the hand primitive (\#10) is executed at \(\bar{\bq}_{\mathrm A}\) in the Posture-A block but at \(\bar{\bq}_0\), the arm straightened, in the Posture-B and Posture-C blocks; and at Posture~C the return of Axis~2 to \(+60^\circ\) is merged with the outbound leg of the Axis-3 sweep, so that one transit segment moves two joints, as the transits of the pair blocks do (Appendix~\ref{app:traj}-E).

Two entries are posture-dependent.
The Axis-4 primitive is two-sided at Posture~A but one-sided at Postures~B and~C, because \(-150^\circ-60^\circ\) lies outside the joint range.
At Posture~C the nominal value \(\theta_2=+60^\circ\) coincides with one endpoint of the Axis-2 sweep, so that primitive is executed as \(+60^\circ\rightarrow0^\circ\rightarrow-60^\circ\rightarrow+60^\circ\); it visits the same three angles as at Postures~A and~B, in a rotated order.

\input{tables/tab_pair_block}

\subsection{The three pair blocks}

Each pair block applies the six adjacent-pair primitives \((1,2),(2,3),\dots,(6,7)\) in that order, followed by one hand-axis primitive.
The blocks differ only in the home posture and sign assigned to each pair, and those assignments are given in Table~\ref{tab:pairblock}.
Pattern~\(\alpha\) keeps every pair at the fully extended posture \(\bar{\bq}_0\); pattern~\(\beta\) alternates between \(\bar{\bq}_{\mathrm B}\) and the wrist-offset extended posture; pattern~\(\gamma\) additionally offsets Axis~2 and applies the wrist offset to the proximal pairs.
Within a block at the normal level the three proximal pairs run at \(0.75\) of the per-axis velocity ceiling and the three distal pairs and the hand primitive at \(0.90\), absolute fractions of the ceiling rather than multiples of the trajectory's level; when a block is repeated at the reduced levels the whole block, proximal pairs included, runs at the single stated fraction, \(0.30\) or \(0.60\).

Every excitation primitive in the set moves one joint or two adjacent joints, and no command file departs from that.
What does depart is the repositioning between primitives: a posture change is sometimes merged into the outbound leg of the next primitive rather than commanded as a segment of its own.
The entry into the \((3,4)\) primitive of \(\Pi_\beta\) and \(\Pi_\gamma\) retracts Axis~4 from \(-150^\circ\) to \(0^\circ\) on the way, which folds into that primitive's own Axis-4 motion and so still moves two joints; the entry into the \((4,5)\) primitive of \(\Pi_\gamma\) carries Axis~2 from \(-60^\circ\) to \(+60^\circ\), and that one does move a third.
Counting all \(2543\) commanded segments of the 40 files, three arm joints move together in \(24\) and never more than three: twenty are that \(\Pi_\gamma\) entry, one per trajectory containing \(\Pi_\gamma\); three are the entry into \(\Pi_\gamma\)'s first pair in PB05, PB10 and PB11, where the wrist offset is applied in the same segment; and one is the opening segment of AP01, which enters \(\Pi_\alpha\) straight from \(\bar{\bq}_{\mathrm A}\) and so carries Axis~4 while Axes~1 and 2 make the first excursion.
Nineteen trajectories, AP03 among them, contain none.
All \(24\) are posture changes rather than excitation, and all are present in the recorded data and modeled as commanded.
The at-most-two-joint property is therefore a property of the primitives, which is where the identifiability argument uses it; the exceptions are the transits between them.

\input{tables/tab_traj_composition}

\begin{figure*}[t]
\centering
\includegraphics[width=0.46\textwidth]{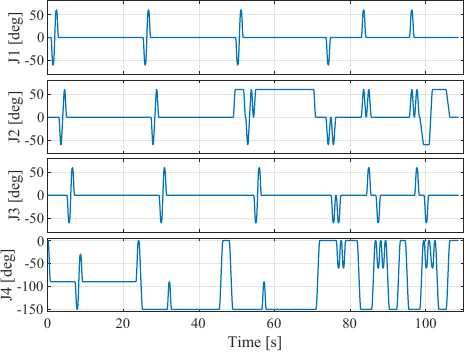}\hfill
\includegraphics[width=0.46\textwidth]{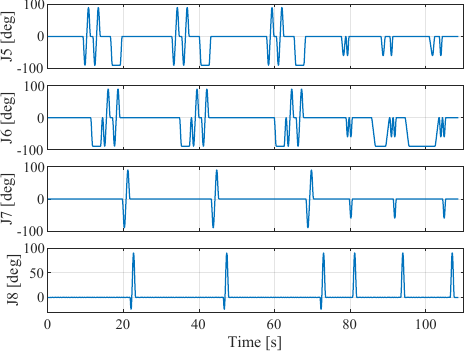}
\caption{Reference commands of AG02, the trajectory whose solution is finally accepted: Axes~1--4 (left) and Axes~5--8 (right). The three single-joint blocks occupy the first two thirds of the record and the three pair blocks the last third, where several axes move together.}
\label{fig:g02_commands}
\end{figure*}

\subsection{Composition of the 40 trajectories}

Table~\ref{tab:trajcomp} lists every trajectory as a block sequence.
The posture families PA, PB, and PC differ only in the nominal posture supplied to the single-joint block: their pair blocks are the same three patterns, executed at the same home postures, so the pattern-to-index assignment in the table is identical for the three families and is written once with \(\mathrm X\in\{\mathrm A,\mathrm B,\mathrm C\}\).
The AP family consists of the three pair blocks used standalone, with AP04 and AP05 repeating the waypoints of AP02 at the two lower speed factors.
AG01 concatenates the three pair blocks, and AG02 concatenates the three single-joint blocks and then the three pair blocks; AG02 is thus the longest trajectory in the set at 163 waypoints, and it is the one whose CLIE solution is finally accepted in Sec.~VI.

\input{tables/tab_validation_severity}

\subsection{Validation trajectories}

\begin{figure}[t]
\centering
\includegraphics[width=0.85\columnwidth]{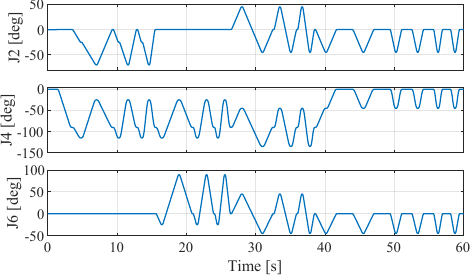}
\caption{Held-out validation trajectory V03. Representative commands of J2, J4, and J6 are shown.}
\label{fig:validation_traj}
\end{figure}

The three validation trajectories V01--V03 use neither the single-joint nor the adjacent-pair structure.
V01 drives all seven arm joints simultaneously between \(\pm45^\circ\) and back at one speed factor; V02 repeats the same waypoints at three factors; V03 adds four-joint arm groups, wrist groups, and shoulder--wrist alternations, again at three factors.
No validation trajectory was used at any stage of parameter estimation, selection, or rescue.

The intent was simultaneous excitation of every axis, which the block-structured identification set deliberately avoids.
The available amplitude was then set by two hardware constraints rather than by choice: the actuators do not have the torque margin to accelerate all seven axes together over a wide arc, and the arm is mounted on a desk, so a wide simultaneous arc collides with the mounting.
Hence \(\pm45^\circ\).
The narrow throw works in the opposite direction to the apparent simplicity: with a short travel there is little constant-velocity phase, so each axis accelerates and almost immediately decelerates, and the share carried by \(\bM(\btheta)\ddot\btheta\), with all of its off-diagonal coupling active, is larger on V01 and V02 than on AG02. The comparison does not hold against every identification trajectory: AP01 has a higher inertia-term share, and V03 is close to AG02 (Table~\ref{tab:valseverity}).

Table~\ref{tab:valseverity} measures this.
The share of predicted-torque RMS carried by \(\bM(\btheta)\ddot\btheta\) alone is \(76.9\%\) on V01 against \(32.9\%\) on the accepted identification trajectory AG02, and \(10.2\%\) on the slowest identification trajectory PA07.
The validation set therefore stresses the inertia matrix harder than the trajectory the model was fitted to, though not harder than every identification trajectory; the shares are not orthogonal, so they rank the trajectories against each other without saying which term dominates in absolute terms.
The higher residual reflects that, at \(41.5\%\) of torque RMS on V01 against \(27.5\%\) in sample on AG02.

Stressing the inertia matrix covers more of the model than it might appear, because the terms are not parameterized independently.
The Coriolis and centrifugal term is built from \(\bM(\btheta)\) through the Christoffel symbols
\begin{equation}
c_{abc}=\frac12\left(\frac{\partial M_{ab}}{\partial\theta_c}+\frac{\partial M_{ac}}{\partial\theta_b}-\frac{\partial M_{bc}}{\partial\theta_a}\right)
\label{eq:christoffel}
\end{equation}
and so introduces no parameter of its own; checking the regressor columns over \(400\) random configurations confirms that all \(21\) parameters entering \(\bC(\btheta,\dot\btheta)\dot\btheta\) are among the \(29\) entering the inertia terms of the eight torque equations, \(28\) in the arm's \(7\times7\) matrix \(\bM(\btheta)\) and \(IA_8\) on the decoupled hand axis.
The same check finds that all first moments from Link~3 outward enter \(\bM(\btheta)\) as well.
Exactly two parameters appear in gravity but not in the inertia matrix, namely the shoulder first moments \(MX_2\) and \(MYR_2\): joint axes 1 and 2 intersect, so the first moment of Link~2 produces no inertial coupling.
Both are excited by gravity in every trajectory at any speed, and one of them, \(MYR_2=-0.238~\mathrm{kg\cdot m}\), is the largest first moment in the model.
Apart from them, the only parameters outside \(\bM(\btheta)\) are the eight viscous friction coefficients.

One limitation remains.
The validation trajectories are slower: peak joint velocity reaches \(2.66\) to \(3.54~\mathrm{rad/s}\) against \(4.41~\mathrm{rad/s}\) for AG02, so the velocity-dependent part of the model, viscous friction in particular, is validated over a narrower range than the acceleration-dependent part.
No held-out record covers the highest-speed regime; there the velocity-dependent terms are supported only by the in-sample fit.

%

\section{Routing Tables and Audit Re-Evaluation}\label{app:routing}

This section collects two small tables from Sec.~VI-B and two notes on the audit re-evaluation of its Sec.~VI-E; the complete numerical record of the 160 identification conditions is Table~\ref{tab:appc}.
Table~\ref{tab:route_examples} compares AP01 with AG02 by condition number, route, and SDP margin, and Table~\ref{tab:g02_rescue_delta} lists the four entries that the post-CLIE rescue of the accepted model moves.

The re-evaluation of all 160 OLS estimates on the routing test and on the audit reported in Sec.~VI-E uses the stored three-significant-digit estimates and counts \(\lambda_{\min}\le0\) of the symmetrized matrix as a violation, the trajectory test at the record's resampled configurations; the routing itself was decided at run time on the full-precision estimate, which Table~\ref{tab:routing_summary} and Table~\ref{tab:appc} report, and the re-audit disagrees with it on two marginal conditions, AP01 at \(10~\mathrm{ms}\) and PA07 at \(40~\mathrm{ms}\), both with \(\lambda_{\min}\) below \(10^{-3}\), where rounding is enough to move the sign.
With the same construction and sample count as the acceptance audit, the accepted model returns a minimum eigenvalue of \(8.40\times10^{-4}\) over \(\Theta\), \(8.67\times10^{-4}\) over the manufacturer's joint limits, and \(7.24\times10^{-4}\) over the narrower box the earlier version described; no conclusion is drawn from the ordering of those three numbers.

\input{tables/tab4_route_examples}
\input{tables/tab6_g02_rescue_delta}


%% file: tables/tab_single_block.tex
\begin{table}[t]
\centering
\footnotesize
\setlength{\tabcolsep}{4pt}
\renewcommand{\arraystretch}{1.05}
\caption{The single-joint block: ten primitives in the order executed. Angles are absolute commanded values in degrees. ``Held'' lists the joints displaced from the nominal posture while the primitive runs; all unlisted joints stay at their nominal values. Every primitive in the block uses the same speed factor.}
\label{tab:sjblock}
\begin{tabular*}{\columnwidth}{@{\extracolsep{\fill}}c c l l@{}}
\toprule
\# & Axis & Commanded sweep & Held \\
\midrule
1  & 1 & \(0 \rightarrow -60 \rightarrow +60 \rightarrow 0\)      & --- \\
2  & 2 & \(0 \rightarrow -60 \rightarrow +60 \rightarrow 0\)      & --- \\
3  & 3 & \(0 \rightarrow -60 \rightarrow +60 \rightarrow 0\)      & --- \\
4  & 4 & \(\bar q_4 \rightarrow \bar q_4-60 \rightarrow \bar q_4+60 \rightarrow \bar q_4\) & --- \\
5  & 5 & \(0 \rightarrow -90 \rightarrow +90 \rightarrow 0\)      & --- \\
6  & 5 & \(0 \rightarrow -90 \rightarrow +90 \rightarrow 0\)      & \(\theta_6=-89\) \\
7  & 6 & \(0 \rightarrow -89 \rightarrow +89 \rightarrow 0\)      & --- \\
8  & 6 & \(0 \rightarrow -89 \rightarrow +89 \rightarrow 0\)      & \(\theta_5=-90\) \\
9  & 7 & \(0 \rightarrow -90 \rightarrow +90 \rightarrow 0\)      & --- \\
10 & 8 & \(0 \rightarrow -24 \rightarrow +90 \rightarrow 0\)      & --- \\
\midrule
\multicolumn{4}{@{}l@{}}{Posture-dependent entries:} \\
\multicolumn{4}{@{}l@{}}{\ \ \#4 at Postures B and C: \(\bar q_4 \rightarrow \bar q_4+60 \rightarrow \bar q_4\) (one-sided)} \\
\multicolumn{4}{@{}l@{}}{\ \ \#2 at Posture C: \(+60 \rightarrow 0 \rightarrow -60 \rightarrow +60\)} \\
\multicolumn{4}{@{}l@{}}{\ \ \#10 at Postures B and C: executed at \(\bar{\bq}_0\) (arm joints at \(0^\circ\)), at Posture A at \(\bar{\bq}_{\mathrm A}\)} \\
\bottomrule
\end{tabular*}
\end{table}

%% file: tables/tab_pair_block.tex
\begin{table}[t]
\centering
\footnotesize
\setlength{\tabcolsep}{3pt}
\renewcommand{\arraystretch}{1.15}
\caption{The three pair blocks. Each row is one adjacent-pair primitive \(\mathcal P_{j,j+1}(s;\bh)\) with \(\delta=60^\circ\): the entry gives the home posture \(\bh\) and, as a sign, the direction \(s\) applied to both joints. \(\bw=-89^\circ\be_6\) is the wrist offset and \(\be_2\) the Axis-2 unit vector. Speed is the commanded velocity ceiling as a fraction of the per-axis \(v_{\max}\) at the normal level, \(0.75\) for the proximal pairs and \(0.90\) for the distal pairs and the hand, not multiplied by the level listed in Table~\ref{tab:trajcomp}; at the reduced levels every primitive of the block runs at \(0.30\) or \(0.60\).}
\label{tab:pairblock}
\begin{tabular*}{\columnwidth}{@{\extracolsep{\fill}}c c c c c@{}}
\toprule
Pair & \(\Pi_\alpha\) & \(\Pi_\beta\) & \(\Pi_\gamma\) & Speed \\
\midrule
\((1,2)\) & \(\bar{\bq}_0\), \(-\) & \(\bar{\bq}_{\mathrm B}\), \(+\)          & \(\bar{\bq}_{\mathrm B}{+}\bw\), \(+\)           & \(0.75\) \\
\((2,3)\) & \(\bar{\bq}_0\), \(-\) & \(\bar{\bq}_{\mathrm B}\), \(+\)          & \(\bar{\bq}_{\mathrm B}{+}\bw\), \(+\)           & \(0.75\) \\
\((3,4)\) & \(\bar{\bq}_0\), \(-\) & \(\bar{\bq}_{0}{+}\bw\), \(-\)            & \(\bar{\bq}_{0}{+}\bw{-}60\be_2\), \(-\)         & \(0.75\) \\
\((4,5)\) & \(\bar{\bq}_0\), \(-\) & \(\bar{\bq}_{0}{+}\bw\), \(-\)            & \(\bar{\bq}_{0}{+}\bw{+}60\be_2\), \(-\)         & \(0.90\) \\
\((5,6)\) & \(\bar{\bq}_0\), \(-\) & \(\bar{\bq}_{\mathrm B}\), \(-\)          & \(\bar{\bq}_{\mathrm C}\), \(-\)                 & \(0.90\) \\
\((6,7)\) & \(\bar{\bq}_0\), \(-\) & \(\bar{\bq}_{\mathrm B}\), \(-\)          & \(\bar{\bq}_{\mathrm C}\), \(-\)                 & \(0.90\) \\
\midrule
hand & \multicolumn{3}{c}{\(\theta_8: 0 \rightarrow +90 \rightarrow 0\) at \(\bar{\bq}_0\)} & \(0.90\) \\
\bottomrule
\end{tabular*}
\end{table}

%% file: tables/tab_traj_composition.tex
\begin{table}[t]
\centering
\footnotesize
\setlength{\tabcolsep}{4pt}
\renewcommand{\arraystretch}{1.05}
\caption{Composition of the 40 identification trajectories. The single-joint block is that of Table~\ref{tab:sjblock} and the three pair blocks are those of Table~\ref{tab:pairblock}; \(+\) denotes concatenation. The first eleven rows apply verbatim to \(\mathrm X\in\{\mathrm A,\mathrm B,\mathrm C\}\) with the corresponding nominal posture, so they define 33 trajectories.}
\label{tab:trajcomp}
\begin{tabular*}{\columnwidth}{@{\extracolsep{\fill}}l l c@{}}
\toprule
Label & Block sequence & Speed \\
\midrule
PX01 & \(\mathcal S+\mathcal S+\mathcal S+\Pi_\alpha+\Pi_\beta+\Pi_\gamma\) & \(0.30/0.60/0.90\) \\
PX02 & \(\mathcal S+\Pi_\alpha+\Pi_\beta+\Pi_\gamma\) & \(0.90\) \\
PX03 & \(\mathcal S+\Pi_\alpha\) & \(0.90\) \\
PX04 & \(\mathcal S+\Pi_\beta\)  & \(0.90\) \\
PX05 & \(\mathcal S+\Pi_\gamma\) & \(0.90\) \\
PX06 & \(\mathcal S+\Pi_\alpha\) & \(0.60\) \\
PX07 & \(\mathcal S+\Pi_\alpha\) & \(0.30\) \\
PX08 & \(\mathcal S+\Pi_\beta\)  & \(0.60\) \\
PX09 & \(\mathcal S+\Pi_\beta\)  & \(0.30\) \\
PX10 & \(\mathcal S+\Pi_\gamma\) & \(0.60\) \\
PX11 & \(\mathcal S+\Pi_\gamma\) & \(0.30\) \\
\midrule
AP01 & \(\Pi_\alpha\) & \(0.90\) \\
AP02 & \(\Pi_\gamma\)  & \(0.90\) \\
AP03 & \(\Pi_\beta\)   & \(0.90\) \\
AP04 & \(\Pi_\gamma\)  & \(0.30\) \\
AP05 & \(\Pi_\gamma\)  & \(0.60\) \\
\midrule
AG01 & \(\Pi_\alpha+\Pi_\beta+\Pi_\gamma\) & \(0.90\) \\
AG02 & \(\mathcal S(\bar{\bq}_{\mathrm A})+\mathcal S(\bar{\bq}_{\mathrm B})+\mathcal S(\bar{\bq}_{\mathrm C})\) & \(0.90\) \\
     & \(\quad{}+\Pi_\alpha+\Pi_\beta+\Pi_\gamma\) & \\
\midrule
\multicolumn{3}{@{}l@{}}{Speed: fraction of the per-axis velocity ceiling. At \(0.90\), the normal level,} \\
\multicolumn{3}{@{}l@{}}{\ \ the proximal pairs of each \(\Pi\) block run at \(0.75\) (Table~\ref{tab:pairblock}).} \\
\multicolumn{3}{@{}l@{}}{PX01: the three \(\mathcal S\) blocks repeat the same sweeps at \(0.30\), \(0.60\), \(0.90\);} \\
\multicolumn{3}{@{}l@{}}{\ \ transit waypoints can differ (trajectory files); \(\Pi\) blocks at the normal level.} \\
\multicolumn{3}{@{}l@{}}{AP04, AP05: identical waypoints to AP02.} \\
\bottomrule
\end{tabular*}
\end{table}

%% file: tables/tab_validation_severity.tex
\begin{table}[t]
\centering
\footnotesize
\setlength{\tabcolsep}{3pt}
\renewcommand{\arraystretch}{1.05}
\caption{How demanding the held-out set is, measured on the 40~ms bundles with the accepted model. Shares are of predicted-torque RMS, evaluated by zeroing velocity, acceleration, or both at the same configurations; they are not orthogonal and need not sum to 100. ``\(k\!\ge\!6\)'' is the fraction of samples at which at least six of the seven arm joints exceed \(0.05~\mathrm{rad/s}\).}
\label{tab:valseverity}
\begin{tabular*}{\columnwidth}{@{\extracolsep{\fill}}l c c c c c@{}}
\toprule
& \multicolumn{3}{c}{Torque share [\%]} & \multicolumn{1}{c}{Peak} & \\
\cmidrule(lr){2-4}\cmidrule(lr){5-5}
Trajectory & \(\bM\ddot{\btheta}\) & \(\bC\dot{\btheta}\) & gravity & \(|\dot\theta|\) & \(k\!\ge\!6\) \\
\midrule
V01 & \textbf{76.9} & 33.0 & 112.9 & 2.66 & 30.6\% \\
V02 & \textbf{53.8} & 23.0 & 109.0 & 3.19 & 46.8\% \\
V03 & \textbf{33.5} & 12.9 & 101.6 & 3.54 & 18.0\% \\
\midrule
AG02 (accepted) & 32.9 & 10.5 & \ \,98.8 & 4.41 & 0\% \\
AP01 & 84.9 & 25.7 & 113.9 & 4.03 & 0\% \\
PA03 (\(0.90\)) & 38.6 & 11.7 & 100.9 & 4.38 & 0\% \\
PC08 (\(0.60\)) & 18.7 & \ \,6.6 & \ \,99.2 & 2.75 & 0\% \\
PA07 (\(0.30\)) & 10.2 & \ \,2.1 & 101.2 & 1.51 & 0\% \\
\midrule
\multicolumn{6}{@{}l@{}}{Peak \(|\dot\theta|\) in rad/s.} \\
\multicolumn{6}{@{}l@{}}{No identification segment commands more than three arm joints at} \\
\multicolumn{6}{@{}l@{}}{once, and at most five are above the threshold in these records;} \\
\multicolumn{6}{@{}l@{}}{every validation trajectory moves all seven.} \\
\bottomrule
\end{tabular*}
\end{table}

%% file: tables/tab4_route_examples.tex
\begin{table}[t]
\centering
\footnotesize
\renewcommand{\arraystretch}{0.95}
\caption{Comparison of the adjacent-pair trajectory AP01 and the representative candidate AG02. A dash in \(\varepsilon_{\mathrm{PD}}\) indicates that the OLS solution was already PD and the SDP projection was skipped.}
\label{tab:route_examples}
\begin{tabular*}{\columnwidth}{@{\extracolsep{\fill}}c c c c c c c@{}}
\toprule
\multirow{2}{*}{\(T_s\) [ms]} & \multicolumn{3}{c}{AP01} & \multicolumn{3}{c}{AG02} \\
\cmidrule(lr){2-4}\cmidrule(l){5-7}
& \(\kappa\) & Route & \(\varepsilon_{\mathrm{PD}}\) & \(\kappa\) & Route & \(\varepsilon_{\mathrm{PD}}\) \\
\midrule
10 & 47.1 & O-S-C & 0.005 & 33.4 & O-S-C & 0.001 \\
20 & 46.1 & O-C   & -- & 33.2 & O-S-C & 0.001 \\
40 & 46.2 & O-C   & -- & 33.3 & O-S-C & 0.001 \\
80 & 47.2 & O-C   & -- & 34.1 & O-C   & -- \\
\bottomrule
\end{tabular*}
\end{table}

%% file: tables/tab6_g02_rescue_delta.tex
\begin{table}[t]
\centering
\footnotesize
\renewcommand{\arraystretch}{0.95}
\caption{Largest parameter changes from CLIE to PD-rescue for AG02 at the 40~ms setting. The last column is the rescued value minus the CLIE value; no other entry moves by more than \(2.1\times10^{-4}\), and the viscous coefficients and \(IA_8\), which do not enter \(\bM(\btheta)\), do not move.}
\label{tab:g02_rescue_delta}
\begin{tabular*}{\columnwidth}{@{\extracolsep{\fill}}l c c c c@{}}
\toprule
Parameter & Unit & CLIE & PD-rescue & \(\Delta\) \\
\midrule
XXR4 & \(\mathrm{kg\cdot m^2}\) & 0.00957 & 0.01040 & 0.00083 \\
ZZR3 & \(\mathrm{kg\cdot m^2}\) & 0.00097 & 0.00160 & 0.00063 \\
ZZR6 & \(\mathrm{kg\cdot m^2}\) & 0.00014 & 0.00074 & 0.00059 \\
ZZR1 & \(\mathrm{kg\cdot m^2}\) & 0.01190 & 0.01226 & 0.00036 \\
\bottomrule
\end{tabular*}
\end{table}

%% file: sections/appendix_d_implementation.tex

\section{Implementation Settings}\label{app:impl}

This section records the settings needed to repeat the computations of Sec.~IV to Sec.~VI: solver choices and tolerances, parameter bounds, stopping criteria, filter settings, and software versions.
The source code is not released, so this section and the pseudocode of Algorithm~1 are the specification.
Where a setting was changed between the submitted and the present version, both are given.

\subsection{Platforms and realization layer}

Two machines are involved and they should not be confused.
The robot was driven from an Intel Core i7-7700 under Ubuntu 24.04, at a \(1~\mathrm{ms}\) control period over an RS485 daisy chain, with the delays quantified in Sec.~III-A.
All offline identification and analysis ran on a mobile workstation with an Intel Core Ultra~7 155H (16 physical, 22 logical cores) and \(32~\mathrm{GB}\) of memory under Windows~11 Pro.
The \(160\) CLIE refinements were executed as \(15\) concurrent MATLAB processes on that machine, which is why Sec.~IV-C reports iterations rather than elapsed time.
Table~\ref{tab:realization} gives the constants of the realization controller (2) for the eight axes; they were set from the joint-level modeling of Sec.~III-B and are controller settings, not identified parameters.

\input{tables/tab_realization_constants}

\subsection{Software}

MATLAB R2024b (24.2.0.3212159, Update~9) throughout, with the Optimization Toolbox for \texttt{lsqnonlin}, the Signal Processing Toolbox for \texttt{designfilt}, \texttt{filtfilt} and \texttt{resample}, the Statistics and Machine Learning Toolbox for \texttt{pca} and \texttt{prctile}, the Parallel Computing Toolbox for the batch, the System Identification Toolbox for the joint-level models of Sec.~III-B, and the Control System Toolbox for the frequency responses.
The semidefinite programs were written in CVX 2.2 (Build 1148) and solved by the SeDuMi 1.3.4 it bundles, selected with \texttt{cvx\_solver sedumi}, at \texttt{cvx\_precision best}.
The symbolic model and regressor were generated by OpenSYMORO, run as its Python~3 port \texttt{symoro-python3} at commit \texttt{bb86bca} (2022).

\subsection{Preprocessing}

The raw log is a nominally \(1~\mathrm{ms}\), unevenly spaced, \(106\)-column record; joint angles are columns 2--9 with a \(\pi\) offset correction, and torque commands are columns 50--57, converted from current by the constants of Sec.~III-A. Recordings are longer than the commanded trajectories, \(128~\mathrm{s}\) against \(107.8~\mathrm{s}\) commanded and about \(115~\mathrm{s}\) realized for AG02, because logging was started and stopped by hand; the idle samples at both ends are retained and inform only the gravity terms.
Angles and torque commands are first placed on an even grid by linear interpolation over the uneven time stamps; resampling to \(1\), \(10\), \(20\), \(40\), and \(80~\mathrm{ms}\) is then preceded by a zero-phase fourth-order Butterworth anti-alias filter (\texttt{filtfilt}) with a corner at \(0.45\) times the Nyquist frequency of the target interval, applied identically to angles and torques, and joint velocities and accelerations are central differences of the filtered angles on the target grid.
The same routine produces every bundle used in the paper; it was re-executed for the present version and reproduced the stored bundles byte for byte.

\subsection{Ordinary least squares}

The regressor has \(39\) columns, taken from the OpenSYMORO inverse-dynamics export, and the eight axes are stacked over all samples.
The solution is obtained by MATLAB's \texttt{\textbackslash} operator, that is by QR factorization of the stacked regressor, and the normal equations are never formed.

\subsection{SDP projection and rescue}

Both programs are the nearest-point projections (8) and (12), and both are solved by CVX/SeDuMi at \texttt{cvx\_precision best}.
They differ in their constraint sets.
The pre-CLIE projection's constraint set \(\Theta_{\mathrm c}\) holds up to \(200\) configurations drawn from the identification trajectory itself, \(120\) at even intervals along the record and the \(80\) with the largest Euclidean norm of the seven joint velocities, duplicates merged, and the projection takes \(\varepsilon_{\mathrm{PD}}\) from the ascending ladder \(\{0.001, 0.002, 0.005, 0.01\}\), adopting the smallest value whose projected solution is positive definite at every configuration of the record, not only at the constrained ones.
Running CLIE from each of the four projections of AG02 at \(40~\mathrm{ms}\) gave final objectives between \(555.0\) and \(555.9\), a spread of \(0.16\%\), with \(86\), \(38\), \(55\), and \(23\) iterations for \(\varepsilon_{\mathrm{PD}}=0.001\), \(0.002\), \(0.005\), and \(0.01\): the margin changed the cost far more than the objective reached, which does not by itself establish agreement of the parameter vectors.
The post-CLIE rescue's set \(\Theta_{\mathrm r}\) is drawn from the whole domain \(\Theta\) of paragraph~G below, at a fixed \(\varepsilon_{\mathrm{PD}} = 10^{-3}\); the earlier version used \(\varepsilon_{\mathrm{PD}} = 0\), which asks only for semidefiniteness, and Sec.~IV-C explains why the change matters.
The rescue is iterative: after each solve the current solution is checked over \(\Theta\), the worst violating configurations are appended to \(\Theta_{\mathrm r}\), up to \(500\) per pass and at most five passes, and a configuration counts as violating when \(\lambda_{\min} \le \varepsilon_{\mathrm{PD}}/2\), the halved threshold serving as a margin against the gaps between constraint samples.
The one rescue needed in this study started \(\Theta_{\mathrm r}\) from \(2{,}000\) random configurations, the \(128\) corners of \(\Theta\), and the \(937\) violating configurations the audit had found for the CLIE solution, and converged in a single pass; re-run without the seeded configurations, for the timing reported in Sec.~VII-D, it needed two passes and appended \(16\). The five-pass cap was never approached; had it been, the solution would not have been accepted, because the audit rather than the solver's exit is the criterion.

\subsection{CLIE refinement}

The refinement calls \texttt{lsqnonlin} with the trust-region-reflective algorithm and \texttt{FunctionTolerance} \(=10^{-6}\); \texttt{UseParallel} is false and every other option is at its default, \texttt{MaxIterations} \(=400\) and \texttt{MaxFunctionEvaluations} \(=100\) times the number of variables among them.
Each iteration cost \(40\) function evaluations, a finite-difference Jacobian over the \(39\) parameters plus the reference point, the solver's count advancing by exactly \(40\) per iteration in every log; within a trajectory the count is therefore proportional to the work done, whereas across trajectories the cost of one evaluation grows with the record length, so counts compare different trajectories only roughly.
The \(160\) refinements ran concurrently on one workstation, so elapsed time reflects the machine's load as much as the problem: sixteen of them were solved more than once after worker restarts, and every repeat reproduced its iterates and residual to all printed digits while elapsed times differed by \(9\%\) on average and \(61\%\) at worst, which is why the paper reports cost as a count. The mean count is \(66\), \(64\), \(65\), and \(63\) iterations at \(10\), \(20\), \(40\), and \(80~\mathrm{ms}\), and three-quarters of the conditions finish within \(65\).
The nine capped runs exited at the solver's iteration \(401\), the limit being tested after the iteration that exceeds it, with \(16{,}080\) evaluations and an exit message naming \texttt{MaxIterations}; the documented evaluation limit of \(3{,}900\) did not stop them, and a 39-variable test problem in the same release, supplied with its output in the accompanying solver-check files, reproduces \(401\) iterations (\texttt{output.iterations}) and \(16{,}080\) evaluations (\texttt{output.funcCount}) with the same algorithm and default limit settings, its three tolerances set to zero solely to isolate limit-based termination; with the evaluation limit entered numerically as \(3{,}900\) the same run stops after \(97\) iterations, so it is the default setting as left that does not bind, and the counts are reported as observed.
Parameter bounds are \([-100\,|x_0|,\,+100\,|x_0|]\) about the initial value \(x_0\), the lower bound raised to zero for \(ZZR_1\) and \(IA_3\)--\(IA_8\) and left unchanged for the viscous coefficients, which are not sign-constrained and all came out positive (Table~\ref{tab:appc}); a diverging simulation returns an infinite residual, which the trust-region step rejects, as happened in the first iterations of several O-S-C runs.
The forward model is the OpenSYMORO direct-dynamics export, integrated by the trapezoidal rule at \(0.25~\mathrm{ms}\); the simulation starts from rest at the first measured positions of the record and is driven by the reference sequence the controller executed, on the record's own time base, and the delay of Sec.~III-B is not modelled.
The objective is the same at all four identification intervals: the raw \(1~\mathrm{ms}\) record is simulated, and the residual is filtered and decimated as described below; the interval enters CLIE only through the OLS or SDP starting point it supplies.
The velocity feedforward is \(\hat{\bM}\hat{\bD}\dot{\btheta}_{\mathrm{ref}}\), matching (2) as the firmware implements it; the submitted simulator applied \(\hat{\bD}\dot{\btheta}_{\mathrm{ref}}\) outside the \(\hat{\bM}\) bracket, and all \(160\) conditions were recomputed for the present version.
The residual is band-limited before it enters the cost by a zero-phase fourth-order IIR low-pass at \(5.63~\mathrm{Hz}\) and decimated to \(40~\mathrm{ms}\).
Samples at which any axis reached its torque clamp are excluded, the test being \(|u_j| \ge 0.98\) of the axis limit --- \(7.4385~\mathrm{N\cdot m}\) for the XM430-W350 axes and \(17.2445~\mathrm{N\cdot m}\) for the XM540-W270 axis --- with the mask dilated by one decimation window to reduce, not eliminate, the leakage of a clamped sample into its neighbours through the residual filter.
In this mounting the clamp is reached on \(0.14\%\) of the samples of AG02; other mountings would reach it more often.

\subsection{All-pose PD audit}

The domain is \(\Theta:\theta_1,\theta_3,\theta_5,\theta_7\in[-180^\circ,180^\circ]\), \(\theta_2,\theta_6\in[-90^\circ,90^\circ]\), \(\theta_4\in[-180^\circ,9^\circ]\), and the pose set is the \(221{,}875\) configurations described in Sec.~VI-E: a \(1\times5\times5\times7\times5\times5\times5\) structured grid, \(150{,}000\) uniform draws, and \(50{,}000\) draws from the folded-arm region \(\theta_2\in[30^\circ,90^\circ]\), \(\theta_4\in[-160^\circ,-75^\circ]\), \(\theta_6\in[55^\circ,90^\circ]\).
Joint~1 is held at one value because \(\bM(\btheta)\) does not depend on \(\theta_1\).
The criterion is \(\lambda_{\min}(\bM(\btheta))>0\) at every configuration, \(\bM(\btheta)\) being the \(7\times7\) arm inertia matrix of Sec.~III-C, and the verdict is unchanged across five random seeds.
Evaluating one model over the set takes about three minutes.
The earlier version used a \(3^7 = 2{,}187\) grid plus \(5{,}000\) uniform draws over the narrower domain \(\theta_{1\text{--}3},\theta_{5\text{--}7}\in[-90^\circ,90^\circ]\), \(\theta_4\in[-180^\circ,0^\circ]\); Sec.~VI-E reports what that sampling missed.

\subsection{Acceptance}

The candidate is named structurally, by the two conditions of Sec.~VI-B, and is then subject to three checks: feasibility, \(\lambda_{\min}>0\) over \(\Theta\) after rescue with \(\varepsilon_{\mathrm{PD}}\le10^{-3}\); conditioning, \(\kappa(\bW_{\mathrm{all}})\) within the lowest decile of the \(160\) conditions; and centrality, within the central quintile of the parameter cloud under a Euclidean metric on the \(39\) standardized dimensions, \(d_i=\lVert(\bphi_i-\bar{\bphi})/\bm{\sigma}\rVert_2\).
The quintile and the decile are conventional cut-offs; a candidate failing either check would be reported as failing rather than replaced, since the checks do not select. Both characterize the candidate against the 160-condition study; a model for a new record needs only the audit and, if needed, the rescue.

\subsection{Cost budget}

Recording is cheap: the 40 identification trajectories total \(46\) minutes of robot motion and the three validation trajectories \(105\) seconds; preprocessing, regressor construction and the OLS solve take seconds per condition. Two stages dominate, by very different amounts. Re-running the post-CLIE rescue of the accepted candidate from a fresh constraint set of \(2{,}128\) configurations (Appendix~\ref{app:impl}-E), including its enrichment loop and the postcheck, takes \(255~\mathrm{s}\) and converges in two passes, with \(16\) configurations appended. One CLIE refinement of the accepted condition takes \(86\) iterations. One audit takes about three minutes. Producing one accepted model therefore costs a few hours of offline computation on a laptop, the population checks of Algorithm~1, centrality and the condition-number decile, being properties of the 160-condition study rather than steps in producing one model; the \(307~\mathrm{h}\) of Sec.~IV-C is the cost of the study, of running every trajectory at every interval, not the cost of an identification. Sec.~VI-B reports the measurements that led us to demote centrality from selector to check.

%% file: tables/tab_realization_constants.tex
\begin{table}[t]
\centering
\footnotesize
\setlength{\tabcolsep}{4pt}
\renewcommand{\arraystretch}{1.0}
\caption{Realization-layer constants of the fixed FF+FB controller of (2). These are motion-realization parameters, not identified dynamic parameters.}
\label{tab:realization}
\begin{tabular*}{\columnwidth}{@{\extracolsep{\fill}}c c c@{}}
\toprule
Axis & \(\hat{M}_j\) [kg$\cdot$m$^2$] & \(\hat{D}_j\) [s$^{-1}$] \\
\midrule
1 & 0.012258 & 0.207 \\
2 & 0.112990 & 0.332 \\
3 & 0.012028 & 0.147 \\
4 & 0.040000 & 0.224 \\
5 & 0.005676 & 0.034 \\
6 & 0.006600 & 0.113 \\
7 & 0.006281 & 0.034 \\
8 & 0.006891 & 0.039 \\
\midrule
\multicolumn{3}{@{}l@{}}{\(K_P = 1000~\mathrm{s^{-2}}\), \(K_V = 50~\mathrm{s^{-1}}\), common to all axes} \\
\multicolumn{3}{@{}l@{}}{(implemented as position gain \(K_PK_V^{-1}=20~\mathrm{s^{-1}}\) and velocity gain \(K_V\hat{M}_j\))} \\
\multicolumn{3}{@{}l@{}}{\(T_f = 0.002~\mathrm{s}\), control period \(1~\mathrm{ms}\), integral action disabled} \\
\bottomrule
\end{tabular*}
\end{table}

%% file: sections/appendix_e_validation.tex

\section{Held-Out Validation in Detail}\label{app:validation}

\subsection{Per-axis errors}

Fig.~\ref{fig:validation_traj} in Appendix~\ref{app:traj}-G shows the representative commands of V03, the most comprehensive validation trajectory and the only one shown qualitatively, and Fig.~\ref{fig:validation_torque} compares measured and predicted torques for Axes~2, 4, and 6. Those three are the pitch axes of the arm: with \(\alpha_1=0\) and \(\alpha_i=\mp\pi/2\) alternating thereafter (Table~1), Axes~2, 4, and 6 rotate about axes perpendicular to the preceding link and therefore carry the gravity load, whereas Axes~3, 5, and 7 rotate about the link itself and Axis~1 about the base vertical; they also span the loading range. That is a defensible choice but it is still a choice, so Table~\ref{tab:perjoint} reports the error on all eight axes rather than letting three of them stand for the model.

Two orderings in that table run opposite to each other. The absolute error follows the torque: Axis~2 has the largest RMSE at about \(0.40~\mathrm{N\cdot m}\) and the hand the smallest at \(0.02\). The relative error runs the other way, from \(3.5\%\) of the peak-to-peak range on Axis~1 to \(13.2\%\) on Axis~7. Axis~6, one of the three plotted in Fig.~\ref{fig:validation_torque}, sits near that end: its RMSE of \(0.042\) to \(0.065~\mathrm{N\cdot m}\) is small in absolute terms but amounts to \(11\) to \(14\%\) of a peak-to-peak range of only \(0.39\) to \(0.54~\mathrm{N\cdot m}\), the second-largest relative error of the eight. The two orderings are one fact seen twice. Going from Axis~2 to Axis~6 the torque range falls by a factor of \(14\) while the error falls by a factor of only \(7\), so part of the residual does not scale with the signal. That is what one expects of encoder quantization, current resolution, and unmodelled drivetrain effects, and it is the light distal axes that pay for it. Fig.~\ref{fig:validation_torque} is thus representative of the model but not of the worst case, and Table~\ref{tab:perjoint} supplies the worst case.

\input{tables/tab_perjoint}

\subsection{The sampling-interval ladder}

\begin{figure}[t]
\centering
\includegraphics[width=0.85\columnwidth]{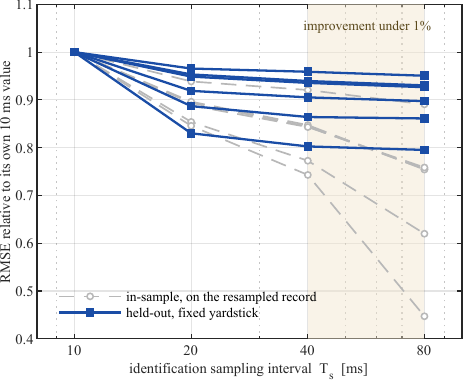}
\caption{The identification sampling interval affects two different things. The in-sample residual, computed on the same record the estimate was fitted to, falls steeply because the anti-alias corner descends with the interval. Measured against a fixed held-out yardstick, the same estimates improve far less, by under \(1\%\) beyond \(40~\mathrm{ms}\). Each curve is one of the six trajectories of Sec.~VII-D, normalized to its own \(10~\mathrm{ms}\) value.}
\label{fig:ts_ladder}
\end{figure}

Fig.~\ref{fig:ts_ladder} separates the two quantities that the sampling-interval ladder of Sec.~VII-D makes easy to conflate. The OLS residual computed on the resampled record itself falls steeply and monotonically with \(T_s\), by \(24\%\) from \(10\) to \(80~\mathrm{ms}\) on AG02 and by \(55\%\) on AP01, almost all of it the anti-alias corner descending with the interval. Measured instead against a fixed yardstick, the same three validation records for every interval, the improvement is still monotone but far smaller and is concentrated at the fine end: from \(10\) to \(20~\mathrm{ms}\) held-out RMSE improves by \(3.4\) to \(17\%\) across trajectories, whereas from \(40\) to \(80~\mathrm{ms}\) it improves by less than \(1\%\) on all six, less than the run-to-run variation of Sec.~VII-D.

\subsection{Torque-constant sensitivity}

Drawing \(\delta_j\sim\mathcal{U}(-10\%,+10\%)\) independently on the eight axes, re-solving OLS on AG02 at \(40~\mathrm{ms}\), and re-evaluating on V01--V03 with the same correction applied there, over \(200\) draws the median parameter moves by \(6.0\%\), essentially the \(5.8\%\) standard deviation of the input. The handful that move much further in relative terms are the ones that are nearly zero --- \(ZZR_3\) at \(2.7\times10^{-4}\) and \(XXR_3\) at \(2.2\times10^{-3}\), the same parameters that fail to reproduce between executions --- and measured against the largest second moment no parameter moves by more than \(4.5\%\). Held-out MNE changes by at most \(1.26\%\).

What does move is the feasibility verdict. The OLS estimate for AG02 at \(40~\mathrm{ms}\) has \(\lambda_{\min}=-4.0\times10^{-3}\); across the \(200\) draws it ranges from \(-9.6\times10^{-3}\) to \(+1.5\times10^{-3}\), and in \(11\) of them the estimate becomes PD. A calibration error of this size can therefore change the estimate and the route it takes through Algorithm~1. What it cannot change is the acceptance criterion: whichever route is taken, a model is accepted only if it passes the same finite-pose audit, the rescue being attempted when it fails, which neither makes the identified parameters invariant nor guarantees that the rescue succeeds.

%% file: tables/tab_perjoint.tex
\begin{table}[t]
\centering
\scriptsize
\setlength{\tabcolsep}{3.2pt}
\renewcommand{\arraystretch}{1.02}
\caption{Joint-wise held-out prediction error of the accepted model at the \(40~\mathrm{ms}\) setting. \(\bar\tau_{\mathrm{pp}}\) is the peak-to-peak measured torque of that axis, averaged over V01--V03; the RMSE columns are per axis on each record; the last column is the error as a fraction of the range, averaged over the three records. Axis~8 is the hand.}
\label{tab:perjoint}
\begin{tabular*}{\columnwidth}{@{\extracolsep{\fill}}l c c c c c@{}}
\toprule
& \(\bar\tau_{\mathrm{pp}}\) & \multicolumn{3}{c}{RMSE [N$\cdot$m]} & \(\overline{e/\tau_{\mathrm{pp}}}\) \\
\cmidrule(lr){3-5}
Axis & [N$\cdot$m] & V01 & V02 & V03 & [\%] \\
\midrule
1 & 3.72 & 0.1260 & 0.1212 & 0.1178 & 3.5 \\
2 & 6.47 & 0.3952 & 0.4105 & 0.3953 & 6.3 \\
3 & 2.05 & 0.0635 & 0.0709 & 0.0764 & 3.7 \\
4 & 3.08 & 0.1979 & 0.1825 & 0.1976 & 6.4 \\
5 & 0.33 & 0.0251 & 0.0383 & 0.0360 & 10.0 \\
6 & 0.46 & 0.0415 & 0.0610 & 0.0651 & 12.2 \\
7 & 0.36 & 0.0407 & 0.0544 & 0.0463 & 13.2 \\
8 & 0.34 & 0.0195 & 0.0295 & 0.0198 & 6.8 \\
\midrule
\multicolumn{5}{l}{Mean over the eight axes (\(=\) MNE of Table~7)} & 7.8 \\
\bottomrule
\end{tabular*}
\vspace{0.6mm}
{\footnotesize\raggedright The eight entries of the last column average to the MNE of Table~7 record by record: \(7.70\), \(8.82\), and \(6.75\%\) for V01, V02, and V03. The V03 column is what Fig.~\ref{fig:validation_torque} plots for Axes~2, 4, and 6.\par}
\end{table}

%% file: sections/acknowledgment.tex

\section*{Acknowledgment}

The authors used Claude (Anthropic)~\cite{AnthropicClaude2026} under their direction to draft and edit the English text of every section, to write the scripts that produced the derived statistics, Figs.~5, 6, and 7 and Figs.~\ref{fig:g02_commands}--\ref{fig:ts_ladder} from the authors' data, and to check the manuscript's numbers, and \mbox{GPT-6 Astra} to check the package for consistency. The authors designed and carried out all experiments, decided every point of method and interpretation, verified every passage and number, and take full responsibility for the content.

\section*{Data Availability}

The numerical specification of the 40 identification trajectories and the 3 validation trajectories described in Appendix~\ref{app:traj} accompanies this preprint as ancillary files: for each, the waypoints and per-segment speeds as executed on the robot, together with the per-axis velocity ceiling and ramp duration that the interpolation needs, and a minimal MATLAB test problem with its output that reproduces the solver stopping behavior recorded in Appendix~\ref{app:impl}-F. The implementation source code is not released.

%% file: sections/author.tex
\begin{IEEEbiography}[{\includegraphics[width=1in,height=1.25in,clip,keepaspectratio]{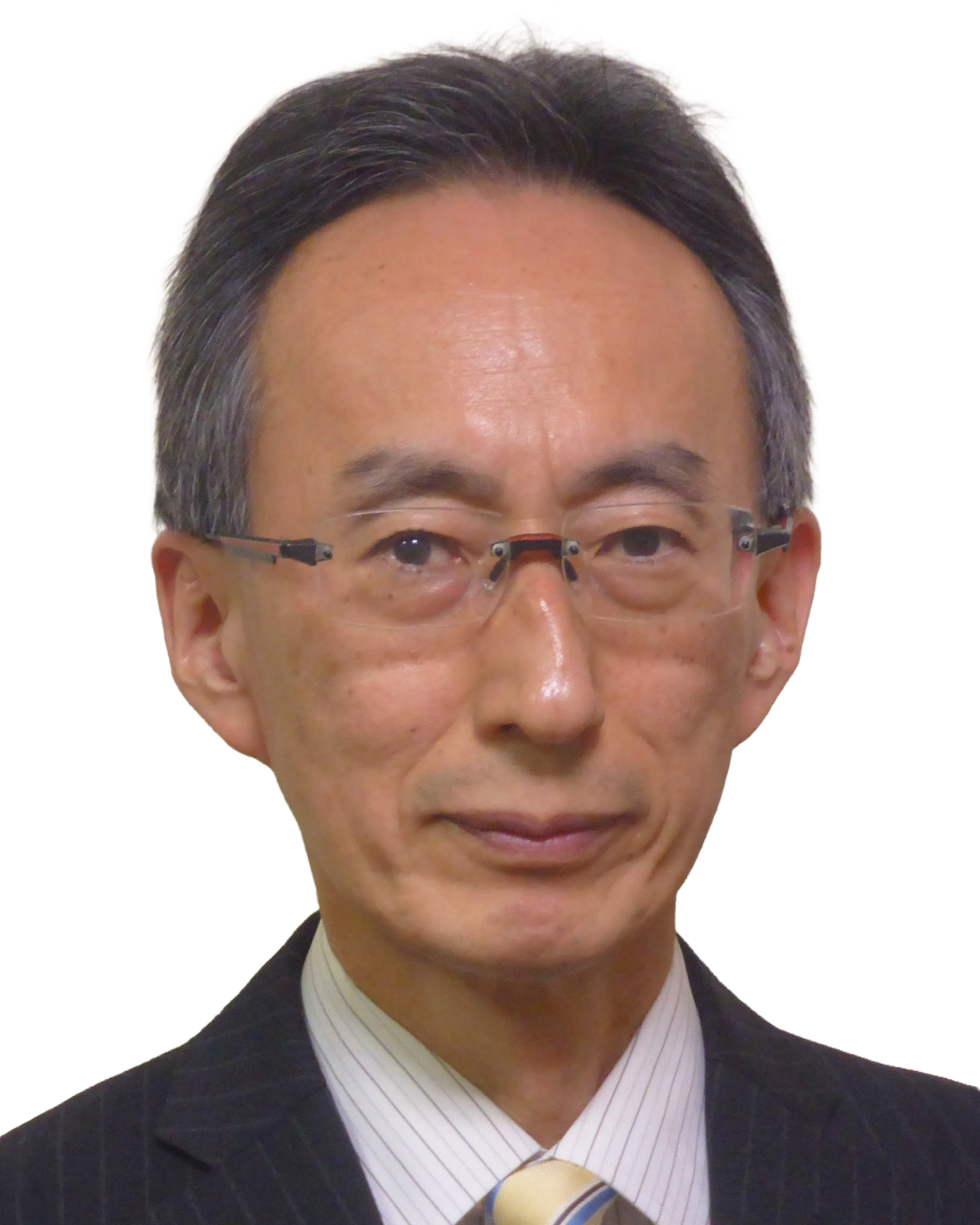}}]{Junji Oaki}
received the B.E. and M.E. degrees in electrical engineering from Chiba University, Chiba, Japan, in 1983 and 1985, respectively, and the Ph.D. degree in fundamental science and technology from Keio University, Yokohama, Japan, in 2010.
From 1985 to 2024, he was with Corporate Research and Development Center, Toshiba Corporation, Kawasaki, Japan.
Since 2024, he has been a full-time Researcher with the University of Tsukuba, Tsukuba, Japan.
His research interests include system identification, motion control, and real-time control systems.
He is a fellow of the Robotics Society of Japan (RSJ).
He received the SICE Control Division Technology Award, in 2012, the IEEJ Industry Applications Society Distinguished Transaction Paper Award, in 2016, and the METI Minister's Award for Excellence at the World Robot Summit, in 2018.
\end{IEEEbiography}

\begin{IEEEbiography}[{\includegraphics[width=1in,height=1.25in,clip,keepaspectratio]{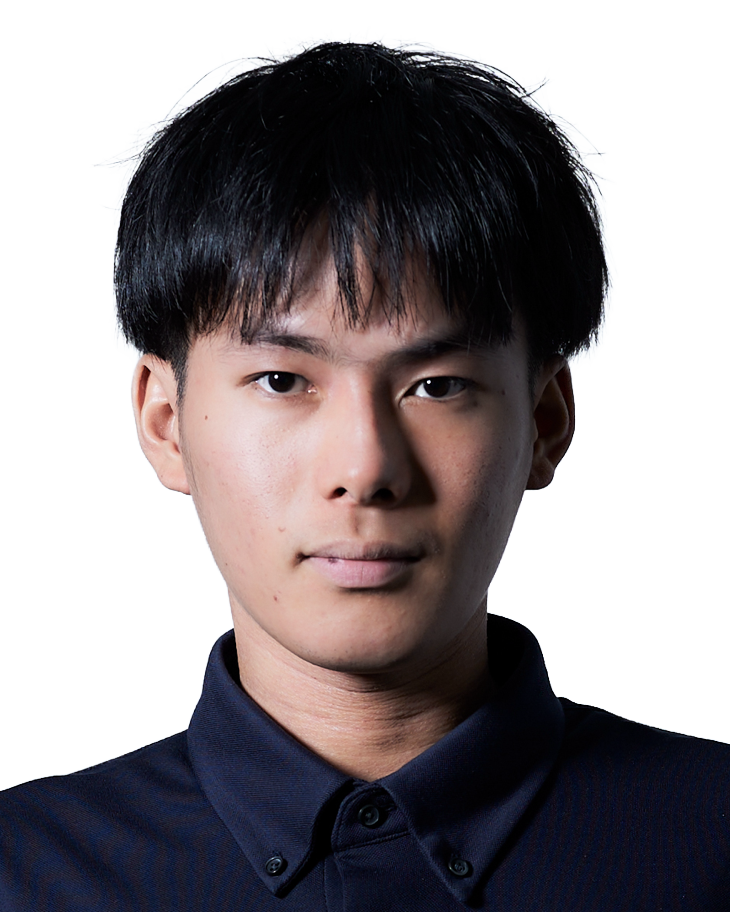}}]{Koki Yamane}
received the B.E. degree in engineering systems and the M.E. degree in intelligent and mechanical interaction systems from the University of Tsukuba, Tsukuba, Japan, in 2022 and 2024, respectively, where he is currently pursuing the Ph.D. degree in intelligent and mechanical interaction systems with the Graduate School of Science and Technology.
From October 2025 to March 2026, he was a Visiting Researcher with the Institute of Cognitive Systems, Technical University of Munich, Munich, Germany.
He has also gained industry research experience with Preferred Networks Inc. and OMRON SINIC X Corporation.
His research interests include force control, bilateral teleoperation, and imitation learning.
He received the Dean's Award and the Outstanding Master's Thesis Award from the University of Tsukuba, in 2024.
He was awarded a JSPS Research Fellowship for Young Scientists (DC1).
\end{IEEEbiography}

\begin{IEEEbiography}[{\includegraphics[width=1in,height=1.25in,clip,keepaspectratio]{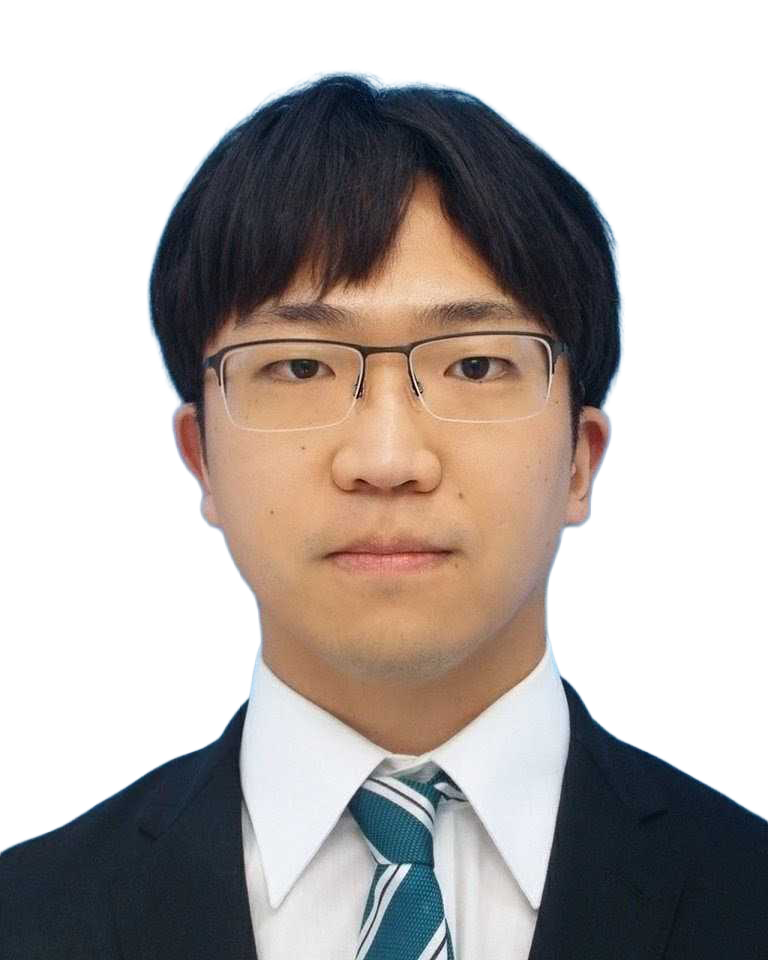}}]{Koki Inami}
received the B.E. degree in policy and planning sciences and the M.E. degree in intelligent and mechanical interaction systems from the University of Tsukuba, Tsukuba, Japan, in 2024 and 2026, respectively, where he is currently pursuing the Ph.D. degree in intelligent and mechanical interaction systems with the Graduate School of Science and Technology.
He has also been involved in industry research on reinforcement learning for humanoid robots.
His research interests include motion control, force control, and imitation learning.
He received the Dean's Award and the Outstanding Master's Thesis Award from the University of Tsukuba, in 2026.
He also received Japan Society of Mechanical Engineers Miura Award, in 2026.
He was awarded a JSPS Research Fellowship for Young Scientists (DC1).
\end{IEEEbiography}

\begin{IEEEbiography}[{\includegraphics[width=1in,height=1.25in,clip,keepaspectratio]{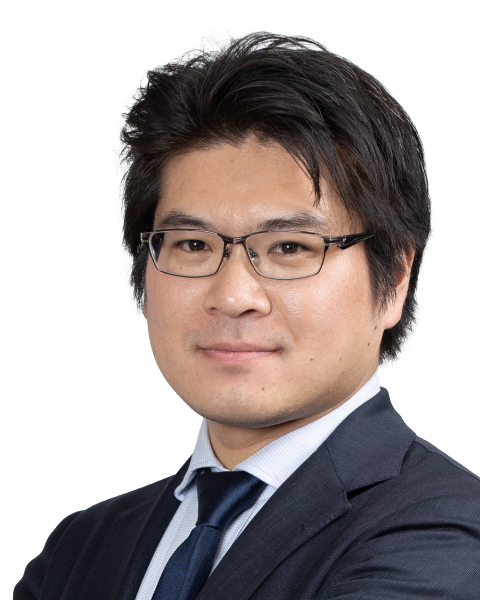}}]{Sho Sakaino}
received the B.E. degree in system design engineering, and the M.E. and Ph.D. degrees in integrated design engineering from Keio University, Yokohama, Japan, in 2006, 2008, and 2011, respectively.
He was an Assistant Professor with Saitama University, from 2011 to 2019.
Since 2019, he has been an Associate Professor with the University of Tsukuba.
His research interests include mechatronics, motion control, robotics, and haptics.
He received the IEEE IES Best Conference Paper Award, in 2022.
He also received the IEEJ Industry Application Society Distinguished Transaction Paper Award, in 2011 and 2020, and the RSJ Advanced Robotics Excellent Paper Award, in 2020.
\end{IEEEbiography}

%% file: refs.bib
@misc{CRANEX7Website,
  author       = {{RT Corporation}},
  title        = {{CRANE-X7}},
  howpublished = {\url{https://rt-net.jp/products/crane-x7/}},
  note         = {Accessed: Sep. 13, 2026}
}

@misc{robotis_dynamixel_x,
  author       = {{ROBOTIS}},
  title        = {{DYNAMIXEL X Series Model Reference}},
  howpublished = {\url{https://docs.robotis.com/docs/dxl/model_reference/x_series/}},
  note         = {Accessed: Sep. 13, 2026}
}

@misc{RTCraneX7Description,
  author       = {{RT Corporation}},
  title        = {{crane\_x7\_description: ROS / ROS 2 package with URDF description macro for CRANE-X7}},
  howpublished = {\url{https://github.com/rt-net/crane_x7_description}},
  note         = {Branch \texttt{ros2}, commit \texttt{d50a70c} (Jun. 12, 2026), accessed: Sep. 13, 2026}
}

@inproceedings{CarpentierEtAl2019Pinocchio,
  author    = {Justin Carpentier and Guilhem Saurel and Gabriele Buondonno and Joseph Mirabel and Florent Lamiraux and Olivier Stasse and Nicolas Mansard},
  title     = {The {Pinocchio} {C++} library: A fast and flexible implementation of rigid body dynamics algorithms and their analytical derivatives},
  booktitle = {Proceedings of the 2019 IEEE/SICE International Symposium on System Integration (SII)},
  address   = {Paris, France},
  month     = jan,
  year      = {2019},
  pages     = {614--619}
}

@misc{AnthropicClaude2026,
  author       = {{Anthropic}},
  title        = {{Claude (Opus 5 and Fable 5.1), used through Claude Code}},
  howpublished = {\url{https://claude.ai}},
  year         = {2026}
}

@article{Yamane2024CrossStructureHand,
  author  = {Koki Yamane and Yuki Saigusa and Sho Sakaino and Toshiaki Tsuji},
  title   = {Soft and Rigid Object Grasping With Cross-Structure Hand Using Bilateral Control-Based Imitation Learning},
  journal = {IEEE Robotics and Automation Letters},
  volume  = {9},
  number  = {2},
  pages   = {1198--1205},
  year    = {2024},
  month   = feb,
  doi     = {10.1109/LRA.2023.3335768}
}

@inproceedings{Inami2025MotionReTouch,
  author    = {Koki Inami and Sho Sakaino and Toshiaki Tsuji},
  title     = {Motion ReTouch: Motion Modification Using Four-Channel Bilateral Control},
  booktitle = {Proceedings of the 2025 IEEE International Conference on Mechatronics},
  address   = {Wollongong, Australia},
  pages     = {1--6},
  year      = {2025},
  month = feb,
  doi ={10.1109/ICM62621.2025.10934911}
}

@article{Yamane2025FastBilateral,
  author  = {Koki Yamane and Yunhan Li and Masashi Konosu and Koki Inami and Junji Oaki and Toshiaki Tsuji and Sho Sakaino},
  title   = {Design and Experimental Validation of Sensorless 4-Channel Bilateral Teleoperation for Low-Cost Manipulators},
  journal = {IEEE Access},
  volume  = {14},
  pages   = {109407--109428},
  year    = {2026},
  doi     = {10.1109/ACCESS.2026.3713604}
}

@inproceedings{Mayeda1984NewIdentification,
  author    = {H. Mayeda and K. Osuka and A. Kangawa},
  title     = {A New Identification Method for Serial Manipulator Arms},
  booktitle = {Proceedings of the 9th IFAC World Congress},
  address   = {Budapest, Hungary},
  pages     = {2429--2434},
  year      = {1984},
  month     = jul,
  note      = {Also in IFAC Proceedings Volumes, vol. 17, no. 2},
  doi       = {10.1016/S1474-6670(17)61346-6}
}

@article{MayedaYoshidaOsukaBaseParam,
  author  = {H. Mayeda and K. Yoshida and K. Osuka},
  title   = {Base Parameters of Manipulator Dynamic Models},
  journal = {IEEE Transactions on Robotics and Automation},
  volume  = {6},
  number  = {3},
  pages   = {312--321},
  year    = {1990},
  month = jun,
  doi = {10.1109/70.56663}
}

@article{KhalilGautierIDIM,
  author  = {Maxime Gautier and Wisama Khalil},
  title   = {Direct Calculation of Minimum Set of Inertial Parameters of Serial Robots},
  journal = {IEEE Transactions on Robotics and Automation},
  volume  = {6},
  number  = {3},
  pages   = {368--373},
  year    = {1990},
  month   = jun,
  doi = {10.1109/70.56655}
}

@article{YoshidaKhalilPhysicalConsistency,
  author  = {Koji Yoshida and Wisama Khalil},
  title   = {Verification of the Positive Definiteness of the Inertial Matrix of Manipulators Using Base Inertial Parameters},
  journal = {The International Journal of Robotics Research},
  volume  = {19},
  number  = {5},
  pages   = {498--510},
  year    = {2000},
  month   = may,
  doi = {10.1177/02783640022066996}
}

@article{SousaCortesao2014LMI,
  author  = {C. D. Sousa and Rui Cortes{\~a}o},
  title   = {Physical Feasibility of Robot Base Inertial Parameter Identification: A Linear Matrix Inequality Approach},
  journal = {The International Journal of Robotics Research},
  volume  = {33},
  number  = {6},
  pages   = {931--944},
  year    = {2014},
  month   = feb,
  doi = {10.1177/0278364913514870}
}

@article{WensingPhysicalConsistency,
  author  = {Patrick M. Wensing and Sangbae Kim and Jean-Jacques E. Slotine},
  title   = {Linear Matrix Inequalities for Physically Consistent Inertial Parameter Identification: A Statistical Perspective on the Mass Distribution},
  journal = {IEEE Robotics and Automation Letters},
  volume  = {3},
  number  = {1},
  pages   = {60--67},
  year    = {2018},
  month   = jan,
  doi = {10.1109/LRA.2017.2729659}
}

@article{BIRDy2021,
  author  = {Quentin Leboutet and Julien Roux and Alexandre Janot and Julio Rogelio Guadarrama-Olvera and Gordon Cheng},
  title   = {Inertial Parameter Identification in Robotics: A Survey},
  journal = {Applied Sciences},
  volume  = {11},
  number  = {9, 4303},
  year    = {2021},
  month = may,
  doi     = {10.3390/app11094303}
}

@inproceedings{OpenSYMORO,
  author    = {W. Khalil and A. Vijayalingam and B. Khomutenko and I. Mukhanov and P. Lemoine and G. Ecorchard},
  title     = {{OpenSYMORO}: An Open-Source Software Package for Symbolic Modelling of Robots},
  booktitle = {Proceedings of the 2014 IEEE/ASME International Conference on Advanced Intelligent Mechatronics},
  address = {Besancon, France},
  pages     = {1206--1211},
  year      = {2014},
  month = jul,
  doi = {10.1109/AIM.2014.6878246}
}

@article{CLOEReference,
  author  = {Gautier, Maxime and Janot, Alexandre and Vandanjon, Pierre-Olivier},
  title   = {A New Closed-Loop Output Error Method for Parameter Identification of Robot Dynamics},
  journal = {IEEE Transactions on Control Systems Technology},
  volume  = {21},
  number  = {2},
  pages   = {428--444},
  year    = {2013},
  month   = mar,
  doi = {10.1109/TCST.2012.2185697}
}

@article{CLIEReference,
  author  = {Adolfo Perrusquía and Ruben Garrido and Wen Yu},
  title   = {Stable robot manipulator parameter identification: A closed-loop input error approach},
  journal = {Automatica},
  volume  = {141, 110294},
  year    = {2022},
  month   = jul,
  doi     = {10.1016/j.automatica.2022.110294}
}

@article{GazEtAl2019Panda,
  author  = {Claudio Gaz and Marco Cognetti and Alexander Oliva and Paolo Robuffo Giordano and Alessandro De Luca},
  title   = {Dynamic Identification of the {Franka Emika Panda} Robot With Retrieval of Feasible Parameters Using Penalty-Based Optimization},
  journal = {IEEE Robotics and Automation Letters},
  volume  = {4},
  number  = {4},
  pages   = {4147--4154},
  year    = {2019},
  month   = oct,
  doi     = {10.1109/LRA.2019.2931248}
}

@article{XuEtAl2020KUKAiiwa,
  author  = {Tian Xu and Jizhuang Fan and Yiwen Chen and Xianyao Ng and Ang, Jr., Marcelo H. and Qianqian Fang and Yanhe Zhu and Jie Zhao},
  title   = {Dynamic Identification of the {KUKA LBR} iiwa Robot With Retrieval of Physical Parameters Using Global Optimization},
  journal = {IEEE Access},
  volume  = {8},
  pages   = {108018--108031},
  year    = {2020},
  month = jun,
  doi     = {10.1109/ACCESS.2020.3000997}
}

@article{HuangKeZhangOta2023TRO,
  author  = {Yanjiang Huang and Jianhong Ke and Xianmin Zhang and Jun Ota},
  title   = {Dynamic Parameter Identification of Serial Robots Using a Hybrid Approach},
  journal = {IEEE Transactions on Robotics},
 volume  = {39},
  number  = {2},
  pages   = {1607--1621},
  year    = {2023},
  month   = apr,
  doi     = {10.1109/TRO.2022.3211194}
}

@article{AtkesonAnHollerbach1986IJRR,
  author  = {Christopher G. Atkeson and Chae H. An and John M. Hollerbach},
  title   = {Estimation of Inertial Parameters of Manipulator Loads and Links},
  journal = {The International Journal of Robotics Research},
  volume  = {5},
  number  = {3},
  pages   = {101--119},
  year    = {1986},
  month   = sep,
  doi     = {10.1177/027836498600500306}
}

@article{SweversEtAl1996MSSP,
  author  = {Jan Swevers and Chris Ganseman and Joris {De Schutter} and Hendrik {Van Brussel}},
  title   = {Experimental Robot Identification Using Optimised Periodic Trajectories},
  journal = {Mechanical Systems and Signal Processing},
  volume  = {10},
  number  = {5},
  pages   = {561--577},
  year    = {1996},
  month   = sep,
  doi     = {10.1006/mssp.1996.0039}
}

@article{SweversEtAl1997TRA,
  author  = {Jan Swevers and Chris Ganseman and Dilek Bilgin Tukel and Joris {De Schutter} and Hendrik {Van Brussel}},
  title   = {Optimal Robot Excitation and Identification},
  journal = {IEEE Transactions on Robotics and Automation},
  volume  = {13},
  number  = {5},
  pages   = {730--740},
  year    = {1997},
  month   = oct,
  doi     = {10.1109/70.631234}
}

@inproceedings{PresseGautier1993ICRA,
  author    = {C. Presse and Maxime Gautier},
  title     = {New Criteria of Exciting Trajectories for Robot Identification},
  booktitle = {Proceedings of the 1993 IEEE International Conference on Robotics and Automation},
  address = {Atlanta, Georgia},
  pages     = {907--912},
  year      = {1993},
  month   = may,
  doi       = {10.1109/ROBOT.1993.292259}
}

@article{Park2006Robotica,
  author  = {Kyung-Jo Park},
  title   = {Fourier-Based Optimal Excitation Trajectories for the Dynamic Identification of Robots},
  journal = {Robotica},
  volume  = {24},
  number  = {5},
  pages   = {625--633},
  year    = {2006},
  month   = mar,
  doi     = {10.1017/S0263574706002712}
}

@article{WuWangYou2010RCIM,
  author  = {Jun Wu and Jinsong Wang and Zheng You},
  title   = {An Overview of Dynamic Parameter Identification of Robots},
  journal = {Robotics and Computer-Integrated Manufacturing},
  volume  = {26},
  number  = {5},
  pages   = {414--419},
  year    = {2010},
  month   = oct,
  doi     = {10.1016/j.rcim.2010.03.013}
}

@article{MataEtAl2005AdvancedRobotics,
  author  = {Vicente Mata and Francesc Benimeli and Nidal Farhat and Angel Valera},
  title   = {Dynamic Parameter Identification in Industrial Robots Considering Physical Feasibility},
  journal = {Advanced Robotics},
  volume  = {19},
  number  = {1},
  pages   = {101--119},
  year    = {2005},
  month   = apr,
  doi     = {10.1163/1568553053020269}
}

@article{JinGans2015RCIM,
  author  = {Jingfu Jin and Nicholas Gans},
  title   = {Parameter Identification for Industrial Robots With a Fast and Robust Trajectory Design Approach},
  journal = {Robotics and Computer-Integrated Manufacturing},
  volume  = {31},
  pages   = {21--29},
  year    = {2015},
  month   = feb,
  doi     = {10.1016/j.rcim.2014.06.004}
}

@article{SweversVerdonckDeSchutter2007CSM,
  author  = {Jan Swevers and Walter Verdonck and Joris {De Schutter}},
  title   = {Dynamic Model Identification for Industrial Robots},
  journal = {IEEE Control Systems Magazine},
  volume  = {27},
  number  = {5},
  pages   = {58--71},
  year    = {2007},
  month   = oct,
  doi     = {10.1109/MCS.2007.904659}
}

@article{SousaCortesao2019TMECH,
  author  = {Crist{\'o}v{\~a}o D. Sousa and Rui Cortes{\~a}o},
  title   = {Inertia Tensor Properties in Robot Dynamics Identification: A Linear Matrix Inequality Approach},
  journal = {IEEE/ASME Transactions on Mechatronics},
  volume  = {24},
  number  = {1},
  pages   = {406--411},
  year    = {2019},
  month   = feb,
  doi     = {10.1109/TMECH.2019.2891177}
}

@article{FuEtAl2021TMECH,
  author  = {Zhongtao Fu and Jiabin Pan and Emmanouil Spyrakos-Papastavridis and Yen-Hua Lin and Xiaodong Zhou and Xubing Chen and Jian S. Dai},
  title   = {A Lie-Theory-Based Dynamic Parameter Identification Methodology for Serial Manipulators},
  journal = {IEEE/ASME Transactions on Mechatronics},
  volume  = {26},
  number  = {5},
  pages   = {2688--2699},
  year    = {2021},
  month   = oct,
  doi     = {10.1109/TMECH.2020.3044758}
}

@article{SujanDubowsky2003TMECH,
  author  = {Vivek A. Sujan and Steven Dubowsky},
  title   = {An Optimal Information Method for Mobile Manipulator Dynamic Parameter Identification},
  journal = {IEEE/ASME Transactions on Mechatronics},
  volume  = {8},
  number  = {2},
  pages   = {215--225},
  month   = jun,
  year    = {2003},
  doi     = {10.1109/TMECH.2003.812830}
}

@article{DiazRodriguezEtAl2013TMECH,
  author  = {Miguel D{\'i}az-Rodr{\'i}guez and {\'A}ngel Valera and Vicente Mata and Marina Vall{\'e}s},
  title   = {Model-Based Control of a 3-DOF Parallel Robot Based on Identified Relevant Parameters},
  journal = {IEEE/ASME Transactions on Mechatronics},
  volume  = {18},
  number  = {6},
  pages   = {1737--1744},
  year    = {2013},
  month   = dec,
  doi     = {10.1109/TMECH.2012.2212716}
}

@article{JanotWensing2021CEP,
  author  = {Alexandre Janot and Patrick M. Wensing},
  title   = {Sequential Semidefinite Optimization for Physically and Statistically Consistent Robot Identification},
  journal = {Control Engineering Practice},
  volume  = {107},
  year    = {104699, Feb. 2021},
  doi     = {10.1016/j.conengprac.2020.104699}
}

@article{WangEtAl2019DVRK,
  author  = {Yan Wang and Radian Gondokaryono and Adnan Munawar and Gregory S. Fischer},
  title   = {A Convex Optimization-Based Dynamic Model Identification Package for the da Vinci Research Kit},
  journal = {IEEE Robotics and Automation Letters},
  volume  = {4},
  number  = {4},
  pages   = {3657--3664},
  year    = {2019},
  month   = oct,
  doi     = {10.1109/LRA.2019.2927947}
}

@inproceedings{ZhaoEtAl2023ALOHA,
  author  = {Tony Z. Zhao and Vikash Kumar and Sergey Levine and Chelsea Finn},
  title   = {Learning Fine-Grained Bimanual Manipulation With Low-Cost Hardware},
  booktitle = {Robotics: Science and Systems XIX},
  address = {Daegu, Republic of Korea},
  year    = {2023},
  month = jul,
  doi     = {10.15607/RSS.2023.XIX.016}
}

@article{AldacoEtAl2024ALOHA2,
  author  = {{ALOHA 2 Team} and others},
  title   = {{ALOHA} 2: An Enhanced Low-Cost Hardware for Bimanual Teleoperation},
  journal = {arXiv preprint},
  year    = {arXiv:2405.02292 [cs.RO], Feb. 2024},
  doi     = {10.48550/arXiv.2405.02292}
}

@book{KhalilDombre2002,
  author    = {Wisama Khalil and Etienne Dombre},
  title     = {Modeling, Identification and Control of Robots},
  publisher = {Hermes Penton Science},
  address   = {London, UK},
  year      = {2002}
}

@article{MaoEtAl2022RimInspection,
  author  = {Wei-Lung Mao and Yu-Ying Chiu and Bing-Hong Lin and Chun-Chi Wang and Yi-Ting Wu and Cheng-Yu You and Ying-Ren Chien},
  title   = {Integration of Deep Learning Network and Robot Arm System for Rim Defect Inspection Application},
  journal = {Sensors},
  volume  = {22},
  number  = {10},
  pages   = {3927},
  year    = {2022},
  doi     = {10.3390/s22103927}
}

@inproceedings{TodorovErezTassa2012MuJoCo,
  author    = {Emanuel Todorov and Tom Erez and Yuval Tassa},
  title     = {{MuJoCo}: A Physics Engine for Model-Based Control},
  booktitle = {Proceedings of the 2012 IEEE/RSJ International Conference on Intelligent Robots and Systems},
  address   = {Vilamoura, Portugal},
  pages     = {5026--5033},
  year      = {2012},
  month     = oct,
  doi       = {10.1109/IROS.2012.6386109}
}

@article{ArdianiJanotBenoussaad2025CEP,
  author  = {Fabio Ardiani and Alexandre Janot and Mourad Benoussaad},
  title   = {A Constrained Instrumental Variable Method for Identification of Industrial Robots},
  journal = {Control Engineering Practice},
  volume  = {154},
  year    = {106168, Jan. 2025},
  doi     = {10.1016/j.conengprac.2024.106168}
}

@article{HuangChengHuang2023FeasibilityTest,
  author  = {Hao-Lun Huang and Ming-Yang Cheng and Tzu-Yuan Huang},
  title   = {A Rapid Base Parameter Physical Feasibility Test Algorithm for Industrial Robot Manipulator Identification Using a Recurrent Neural Network},
  journal = {IEEE Access},
  volume  = {11},
  pages   = {145692--145705},
  year    = {2023},
  doi     = {10.1109/ACCESS.2023.3344490}
}

@article{RuckerWensing2022SmoothInertia,
  author  = {Caleb Rucker and Patrick M. Wensing},
  title   = {Smooth Parameterization of Rigid-Body Inertia},
  journal = {IEEE Robotics and Automation Letters},
  volume  = {7},
  number  = {2},
  pages   = {2771--2778},
  year    = {2022},
  month   = apr,
  doi     = {10.1109/LRA.2022.3144517}
}

@article{LeeWensingPark2020Geometric,
  author  = {Taeyoon Lee and Patrick M. Wensing and Frank C. Park},
  title   = {Geometric Robot Dynamic Identification: A Convex Programming Approach},
  journal = {IEEE Transactions on Robotics},
  volume  = {36},
  number  = {2},
  pages   = {348--365},
  year    = {2020},
  month   = apr,
  doi     = {10.1109/TRO.2019.2926491}
}

@article{GiacomuzzoEtAl2024GPLIP,
  author  = {Giulio Giacomuzzo and Ruggero Carli and Diego Romeres and Alberto {Dalla Libera}},
  title   = {A Black-Box Physics-Informed Estimator Based on {Gaussian} Process Regression for Robot Inverse Dynamics Identification},
  journal = {IEEE Transactions on Robotics},
  volume  = {40},
  pages   = {4820--4836},
  year    = {2024},
  doi     = {10.1109/TRO.2024.3474851}
}
